\documentclass[11pt]{article}

\usepackage[margin=1in]{geometry}

\usepackage{amsmath,amssymb,amsthm,mathtools,bm}
\usepackage{mathrsfs}
\usepackage{booktabs,array,tabularx,longtable}
\usepackage{enumitem}
\usepackage{graphicx}
\usepackage{float}
\usepackage{tikz}
\usetikzlibrary{arrows.meta,positioning,fit,calc}
\usepackage{xcolor}
\usepackage{microtype}

\usepackage[round,authoryear]{natbib}
\usepackage{hyperref}
\usepackage[nameinlink,noabbrev]{cleveref}

\setlist[itemize]{leftmargin=1.5em,itemsep=1.6pt,topsep=2pt}
\setlist[enumerate]{leftmargin=1.8em,itemsep=1.6pt,topsep=2pt}

\theoremstyle{plain}
\newtheorem{theorem}{Theorem}
\newtheorem{proposition}[theorem]{Proposition}
\newtheorem{lemma}[theorem]{Lemma}
\newtheorem{corollary}[theorem]{Corollary}

\theoremstyle{definition}
\newtheorem{definition}[theorem]{Definition}
\newtheorem{example}[theorem]{Example}

\theoremstyle{remark}

\crefname{assumption}{assumption}{assumptions}
\Crefname{assumption}{Assumption}{Assumptions}
\crefname{lemma}{lemma}{lemmas}
\Crefname{lemma}{Lemma}{Lemmas}
\crefname{proposition}{proposition}{propositions}
\Crefname{proposition}{Proposition}{Propositions}
\crefname{corollary}{corollary}{corollaries}
\Crefname{corollary}{Corollary}{Corollaries}
\crefname{theorem}{theorem}{theorems}
\Crefname{theorem}{Theorem}{Theorems}
\crefname{definition}{definition}{definitions}
\Crefname{definition}{Definition}{Definitions}
\crefname{example}{example}{examples}
\Crefname{example}{Example}{Examples}
\crefname{remark}{remark}{remarks}
\Crefname{remark}{Remark}{Remarks}

\newcommand{\R}{\mathbb{R}}
\newcommand{\N}{\mathbb{N}}
\newcommand{\E}{\mathbb{E}}
\newcommand{\Id}{\operatorname{Id}}

\newcommand{\norm}[1]{\left\lVert #1\right\rVert}

\hypersetup{
  hidelinks,
  pdftitle={Architecture Before the Formula: Training-Dependent Computational Semantics in Transformers},
  pdfauthor={Luis F. Rosario Freytes},
  pdfsubject={Deep learning externalism, represented computation, and Transformer training dynamics},
  pdfkeywords={deep learning externalism, represented computation, Transformers, predictive state, learning dynamics}
}

\begin{document}

\title{Architecture Before the Formula: Training-Dependent Computational Semantics in Transformers}

\author{
Luis F. Rosario Freytes\\
University of Michigan\\
\texttt{luisrosa@umich.edu}
}
\date{}

\maketitle

\begin{abstract}
An operation's input-output function does not, by itself, determine its computational role within a represented process. That role also depends on the processes producing its inputs and receiving its outputs, a philosophical stance termed deep learning externalism (DLE). We develop a mathematical account of represented computation that distinguishes functional equivalence, retained predecessor distinctions, restricted receiver capability, and context-dependent effective architectures. The resulting equivalences depend on which computational relations a scientific description must preserve. We prove that two injective producer contexts can preserve every predecessor distinction while changing whether fixed local and attention receiver families are equivalent.

We investigate these relations in a dimension-labelled PreNorm Transformer and their evolution under training. Residual bypasses restrict complete-state information identifications to two merges per block, while finer branches can change receiver accessibility. Exact gradient-descent models yield covariance-dependent geometric attenuation under learned producer routing and algebraic $(t^{-1/2})$ attenuation under Q/K-softmax training, with optimal bounded-receiver recovery errors derived in both cases. Coupled producer–attention dynamics establish how learning jointly reorganizes the presented features and receiver allocation. Finally, present functional or receiver-profile equivalence need not determine future behavior under an optimizer. In an input-conditioned attention model, two accessibility measurements form an exact reduced predictive state, whereas one alone does not.
\end{abstract}

\noindent\textbf{Keywords:}
deep learning externalism, represented computation, Transformer architecture,
learning dynamics, predictive information, coarse-graining.

\paragraph{Assistance disclosure.}
Large language models were used as assistive tools for formalization, literature search, code and typesetting support, and language editing. The author supplied the conceptual program and retains responsibility for the arguments, references, and wording.

\section{Introduction}
\label{sec:paper-introduction}

Consider a neural operation selected from a larger computation. Its input arrives through one process; its output is used by another. The operation's isolated function specifies neither how its input was produced nor how its output is subsequently used. Those surrounding processes determine its computational role at the chosen interface.

We call the framework \emph{deep learning externalism} (DLE) because deep learning makes several relationships visible in one system: a computation receives a represented presentation, its output is used by particular continuations, and training changes both sides. The name identifies the domain motivating and developing the theory, not a restriction on its mathematical objects. Part~I retains the producer--receiver relations of represented processes without requiring neural layers or a learning rule. Its results apply to processes satisfying their stated categorical and semantic hypotheses. Part~II returns to deep learning, where those relations are realized by Transformer operations and evolve under an implemented optimizer.

Externality is relative to a selected computational cut, not necessarily to the boundary of a network or physical system. The same layer can belong to the receiver continuation of an earlier cut and the producer history of a later cut. A statement about the semantics of an isolated operation therefore has a context: which represented process supplies it, which continuation receives it, and which relations the analysis retains. For example, consider
\[
\Omega\xrightarrow{P}X\xrightarrow{m}Y\xrightarrow{S}Z.
\]
The producer \(P\) determines the presentation reaching \(m\), and the receiver \(S\) determines what the produced output can do in the continuation. Holding \(m\) fixed while changing either context gives a different represented computation at this cut. Holding the complete network fixed while moving the cut changes which components count as its exterior. These are two distinct ways in which computational semantics depend on context.

We formalize this with a category \(\mathcal A\) of represented processes, a category \(\mathcal S\) of extensional semantics, and a possibly nonfaithful realization functor
\[
R:\mathcal A\to\mathcal S.
\]
The realization functor can identify distinct represented processes. Factorization retains the producer and receiver of a cut even where the composite realized map agrees. Subsequent quotients retain only specified distinctions or observables. Thus a function-level comparison, an information-level comparison, and a comparison of represented computational roles have different equality criteria.

These are different criteria of computational identity.
Equality of realized maps determines the same
boundary-to-boundary function; equality of kernel
pairs determines the same predecessor
identifications; equality of represented cuts
retains the producing process and receiving
continuation.

No single factorization or equivalence criterion
is privileged independently of the scientific
question being investigated. A computational
description is adequate when it retains the
structure needed to determine the quantity
being explained. This does not make computational
identity arbitrary: once the represented
processes and comparison criterion are fixed,
the resulting equivalence relations are
mathematically determined.

Section~\ref{sec:contextual-search} establishes
a concrete consequence: the effective equivalence
of fixed receiver architectures can change
with an injective producer context.

\begin{figure}[H]
\centering
\begin{tikzpicture}[
    block/.style={draw,rounded corners=2pt,align=center,text width=0.39\linewidth,minimum height=9mm,inner sep=3pt,font=\footnotesize},
    arr/.style={-{Latex[length=2mm]},thick},
    node distance=5mm]
\node[font=\small\bfseries] (headL) at (-3.65,0) {Static cut descriptions};
\node[font=\small\bfseries] (headR) at (3.65,0) {Training-predictive descriptions};
\node[block,below=5mm of headL] (cut) {Represented cut\\$(P,m,S)$};
\node[block,below=5mm of headR] (state) {Complete training state\\$(\theta,\omega)$};
\node[block,below=5mm of cut] (fun) {Realized composite\\$R(SmP)$};
\node[block,below=5mm of state] (k) {Transition kernel $K$\\and chosen observable $\Phi$};
\node[block,below=5mm of fun] (shadow)
    {Induced equivalence\\relation};
\node[block,below=5mm of k] (pred) {Predictive-information\\state};
\draw[arr] (cut) -- (fun);
\draw[arr] (fun) -- (shadow);
\draw[arr] (state) -- (k);
\draw[arr] (k) -- (pred);
\draw[arr,dashed] (cut.east) to[out=0,in=180] (k.west);
\end{tikzpicture}
\caption{
The left column passes from a represented cut to
its realized composite and the equivalence relation
induced on predecessor states. The right column
starts from a complete training state, its transition
kernel, and a selected observable, and derives the
information required to predict that observable.
The two constructions retain different information
about the same represented computation.
}
\label{fig:controlled-forgetting}
\end{figure}

\begin{figure}[H]
\centering
\begin{tikzpicture}[
    >=Latex,
    producer/.style={
        draw,
        rounded corners=2pt,
        text width=48mm,
        minimum height=17mm,
        align=center,
        font=\small,
        inner sep=5pt
    },
    outcome/.style={
        draw,
        rounded corners=2pt,
        text width=50mm,
        minimum height=17mm,
        align=center,
        font=\small,
        inner sep=5pt
    },
    arrow/.style={
        -{Latex[length=2mm]},
        thick
    }
]

\node[font=\footnotesize]
    at (0,1.3)
    {Injective producer};

\node[font=\footnotesize]
    at (6.8,1.3)
    {Fixed local / attention receiver families};

\node[producer] (enc) at (0,0) {
    $A_{\mathrm{enc}}(u,v)$\\[4pt]
    token 1: $(u,v)$\\
    token 2: $(0,0)$
};

\node[outcome] (encout) at (6.8,0) {
    Local and attention\\
    same distinction profile
};

\node[producer] (sep) at (0,-2.5) {
    $A_{\mathrm{sep}}(u,v)$\\[4pt]
    token 1: $(u,0)$\\
    token 2: $(0,v)$
};

\node[outcome] (sepout) at (6.8,-2.5) {
    Local: $u$ only\\
    Attention: $u$ and $v$\\[2pt]
    Different distinction profiles
};

\draw[arrow] (enc.east) -- (encout.west);
\draw[arrow] (sep.east) -- (sepout.west);

\end{tikzpicture}

\caption{
Two injective producers present the same predecessor
information differently. With fixed local and attention
receiver families, the encoded presentation gives
identical distinction profiles. The separated
presentation gives the attention family access to
$v$, which is unavailable to the token-1 local
receiver. Thus receiver equivalence changes
without loss of predecessor information.
}
\label{fig:injective-contextual-comparison}
\end{figure}

The construction shows that injectivity of the
producer does not determine the equivalence
of restricted receiver families. Part~%
\ref{part:represented-architecture} establishes
this result within a general theory of
represented processes.

Returning to the motivating system, the Transformer makes it possible to follow the same dependence through learning. Residual branches retain different distinctions from the complete carried state; earlier blocks supply the features seen at a cut and later blocks supply the loss covectors that train it. A fixed architecture thus admits changing computational roles as the producer and receiver parameters evolve. Part~\ref{part:trained-refinement} resolves the PreNorm operations, derives producer- and attention-learning dynamics, and measures changes in receiver accessibility. Some of the resulting distinctions change only at an exact limiting configuration; others become progressively harder for receivers with bounded sensitivity to use. The training kernel then determines what information about a present contextual role is needed to predict its evolution.

Whether a computation is individuated by its intrinsic mechanism,
represented content, or surrounding context is a longstanding
question in the philosophy of computation. Semantic and
content-sensitive accounts make representation relevant to
computational identity
\citep{shagrir2001content,sprevak2010computation};
mechanistic and externalist accounts examine the organization
and circumstances of the realizing process
\citep{dewhurst2018individuation,harbecke2019role,
fresco2021longarm,richmond2024externalism}.
The choice of descriptive grain matters here.
\citet{klein2026isomorphism} examines how
additional intrinsic computational structure
can constrain apparent indeterminacies,
while causal abstraction studies
correspondences between lower-level mechanisms
and higher-level algorithms through their
intervention behavior
\citep{geiger2025causalabstraction}.
These approaches make different structural commitments about
what must be retained before two computations can be compared.

The categorical analysis below uses deterministic
semantics in \(\mathbf{Set}\), with the
exact-category extension developed in
Appendix~\ref{sec:exact-quotients}.
The Transformer realization fixes its architecture,
data, training law, and optimizer explicitly.
The resulting statements concern the represented
processes and learning experiments determined
by those choices.

\part{Represented Architecture and Deep Learning Externalism}
\label{part:represented-architecture}

\section{Deep learning externalism at represented interfaces}
\label{sec:architecture-individuation}

Part~\ref{part:represented-architecture} abstracts the two-sided relations that motivate DLE. A carrier is encountered through a process: a producer reaches the interface from one side and a receiver continues from it on the other. The state at the carrier does not by itself determine that computational role. This question arises in neural networks, but neither a neural implementation nor a training algorithm is needed to define a represented cut. Producer and receiver relations remain explicit until the scientific quantity under study has been shown to descend through a coarser description.

A simple data-pipeline example already contains the distinction. Suppose
\[
\Omega\xrightarrow{C}X_{\mathrm{clean}}\xrightarrow{H}X_{\mathrm{analysis}}\xrightarrow{M}Z
\]
and the handoff \(H\) preserves the table exactly in extensional semantics. The producer side still records how the state was made--which variation was treated as noise, imputed, merged, or calibrated--while the receiver side records what the next computation can ask of that state. A distinction may be destroyed upstream before a downstream analysis receives it, or it may survive perfectly while remaining unavailable to the permitted continuation. Those are different failure modes; neither requires the downstream process to be a neural network.

The handoff's realized function, the distinctions
retained at that handoff, and the analyses admissible
after it therefore answer different questions.
The first describes a transformation, the second
its identifications, and the third what a specified
continuation can accomplish. Treating these as
separate descriptions makes it possible to say
exactly which portion of a represented process
a proposed explanation needs.

Two familiar questions in deep learning meet at the selected
computational cut. The first concerns representations.
Representation-similarity methods depend on which transformations
are treated as preserving the represented information
\citep{kornblith2019similarity,roeder2021linear}.
Yet an invertible change of representation can preserve every
distinction while changing what a restricted downstream
computation can recover. Computationally usable information
makes the receiving predictor family explicit
\citep{xu2020usable,hewitt2021conditional}.
We therefore distinguish the predecessor identifications
induced by a representation from the distinctions usable by
a specified receiver.

The second question concerns computational organization.
Transformer circuit analyses retain heads, paths, and
feed-forward computations as separately identifiable
components \citep{elhage2021framework,geva2021ffn}.
Interchange and replacement tests examine whether a component
continues to play the same role in another represented position
\citep{garcia2026nofreeswap}. Architecture-search descriptions
likewise retain choices of operations and their organization
\citep{ying2019nasbench,white2020encodings,
ericsson2024einspace}. These comparisons require a decision
about which factorizations count as the same computation.
Categorical treatments of parametric learning provide tools
for studying composition \citep{gavranovic2024categorical}.
Here the represented process is supplied first, and the
factorizations and receiver relations selected from it
determine the comparisons under study.

\subsection{Represented cuts and induced equivalence relations}

For a represented path
\[
X_0\xrightarrow{F_1}X_1
\xrightarrow{F_2}\cdots
\xrightarrow{F_n}X_n,
\]
a cut at \(X_k\) separates the producer prefix
\[
P_k:=F_k\cdots F_1
\]
from the receiver continuation
\[
S_k:=F_n\cdots F_{k+1}.
\]
Together they retain the represented factorization
\[
X_0\xrightarrow{P_k}X_k
\xrightarrow{S_k}X_n.
\]

The realized prefix \(R(P_k)\) induces an
equivalence relation on the initial state space:
two predecessor states are equivalent precisely
when they produce the same state at the selected
interface. This relation records which
distinctions have survived the prefix, while
the represented factorization also retains
the producer and continuation.

Consequently, two cuts can induce the same
equivalence relation on predecessor states
while supporting different downstream
computations. Section~\ref{sec:components-grain}
develops the represented factorizations, and
Section~\ref{sec:receiver-types} derives their
kernel-pair relations.

\begin{figure}[t]
\centering
\begin{tikzpicture}[
  node distance=18mm,
  state/.style={draw,rounded corners,minimum width=12mm,minimum height=8mm,font=\small},
  arr/.style={-{Latex[length=2mm]},thick},
  lab/.style={font=\scriptsize,align=center}
]
\node[state] (x0) {$X_0$};
\node[state,right=of x0] (xk) {$X_k$};
\node[state,right=of xk] (xkp) {$X_{k+1}$};
\node[state,right=of xkp] (xn) {$X_n$};
\draw[arr] (x0) -- node[above,lab] {$P_k$} (xk);
\draw[arr] (xk) -- node[above,lab] {$F_{k+1}$} (xkp);
\draw[arr] (xkp) -- node[above,lab] {$S_{k+1}$} (xn);
\node[below=6mm of xk,lab] (prod) {producer history};
\node[below=6mm of xkp,lab] (recv) {receiver continuation};
\draw[dashed] (prod.north) -- (xk.south);
\draw[dashed] (recv.north) -- (xkp.south);
\end{tikzpicture}
\caption{A represented interface has two external relations. The producer prefix and receiver continuation are not determined by the isolated carrier state.}
\label{fig:two-sided-cut}
\end{figure}

Changing the cut also changes grain. In
\[
X_0\xrightarrow{F_1}X_1\xrightarrow{F_2}X_2\xrightarrow{F_3}X_3,
\]
\(F_2\) lies in the receiver continuation at the first cut and in the producer history at the second. Nothing physical has moved; the represented boundary changed. We therefore use \emph{architecture} for the represented organization of computation across such interfaces: carriers, represented process segments, and factorization relations that survive at the declared grain.

We first study represented factorizations
and the equivalence relations induced by
their realized maps. We then restrict the
receiving computations and compare their
equivalence relations under different
producer contexts.

Part~\ref{part:trained-refinement} realizes
these relations inside a trainable Transformer,
where the surrounding computations determine
both the features and the loss covectors
entering a local parameter update.
\section{Represented processes, realization, and factorization}
\label{sec:components-grain}

A represented computation specifies the
processes connecting exposed state interfaces.
Selecting an intermediate interface separates
the process producing its state from the
continuation receiving that state. Retaining
the factorization makes these components
available for comparison; composing the
factors removes the intermediate interface
from the displayed process.

The realized composite gives a different
description. It records the induced
computation on semantic states but need
not determine the represented factorization
that produced it. We formalize these two
descriptions using a category of represented
processes and a realization functor.

\subsection{Architectural processes and their semantic realization}

We therefore use two categories from the beginning.

\begin{definition}[Architectural realization datum]\label{def:architectural-realization}
An \emph{architectural realization datum} consists of a locally small category $\mathcal A$, a category $\mathcal S$, and a functor
\[
\boxed{
R:\mathcal A\to\mathcal S.
}
\]
Objects of $\mathcal A$ are represented states or structures exposed as carriers by selected cuts of a computational process. A morphism
\[
F:X\to Y
\]
in $\mathcal A$ is a represented process segment taking the state at one selected cut to the state at the next. The functor $R$ sends represented interfaces and process segments to their extensional semantic realizations. In the neural specialization, these objects are hidden or exposed model states and the represented neural computations between them.
\end{definition}

Equality in $\mathcal A$ is equality of the supplied represented process at the retained architectural grain. Functional equality is evaluated only after applying $R$; observational and information-level equivalences arise under subsequent quotients. Thus the distinct equality relations are defined on distinct descriptions.

In the deterministic setting, use
\[
\mathcal S=\mathbf{Set}.
\]
Metric, measurable, probabilistic, or linear structure will be declared only when a later result needs it. In particular, expressions such as $\ker R(F)$, conditional expectation, and metric error belong to the semantic realization, not silently to the abstract arrow $F$.

The functor $R$ is \emph{not} assumed faithful. Hence the theory allows
\[
F\ne G
\qquad\text{in }\mathcal A,
\]
while
\[
R(F)=R(G)
\qquad\text{in }\mathcal S.
\]
Equality of the realized function therefore does not force equality of represented architecture. Network morphisms and function-preserving reparameterizations provide familiar settings in which extensional equality can coexist with different represented factorizations \citep{wei2016networkmorphism}; the present formalism places that possibility at the level of the realization functor rather than treating it as an exceptional case.

Composition is respected:
\[
R(GF)=R(G)R(F).
\]
Composition contracts intermediate represented interfaces; realization can then identify distinct composite arrows. These are two separate identifications.

An identity-block insertion later gives a concrete case: a coarse Transformer and its refinement compute the same predictor while exposing different internal branches and parameter directions. Their functional equality is compatible with their different represented factorizations.

\subsection{One arrow has receiver and producer fibers}

Let $\mathcal A^\to$ denote the arrow category. Its objects are represented component arrows
\[
F:X\to Y.
\]
A morphism from $F:X\to Y$ to $F':X'\to Y'$ is a commuting square
\[
\begin{array}{ccc}
X & \xrightarrow{F} & Y\\
\downarrow A && \downarrow B\\
X'&\xrightarrow{F'}&Y'
\end{array}
\qquad
BF=F'A.
\]
The two canonical projections
\[
\operatorname{dom},\operatorname{cod}:
\mathcal A^\to\to\mathcal A
\]
record the input and output interfaces of each represented event.

For a fixed input interface $X$, the outgoing receiver presentations form the coslice
\[
\mathsf{Rec}(X):=X/\mathcal A.
\]
For an upstream component $A:\Omega\to X$, precomposition gives
\[
A^*:
X/\mathcal A\to\Omega/\mathcal A,
\qquad
A^*(F)=FA.
\]
For a fixed output interface $Y$, the incoming producer presentations form the slice
\[
\mathsf{Prod}(Y):=\mathcal A/Y.
\]
For a downstream component $B:Y\to Z$, postcomposition gives
\[
B_*:
\mathcal A/Y\to\mathcal A/Z,
\qquad
B_*(F)=BF.
\]

These two indexed families are the formal version of the producer/receiver reading introduced in Section~\ref{sec:architecture-individuation}.

\begin{theorem}[Two-sided arrow organization]\label{thm:two-sided-arrow}
The assignments
\[
\mathsf{Rec}:\mathcal A^{\mathrm{op}}\to\mathbf{Cat},
\qquad
X\longmapsto X/\mathcal A,
\]
with reindexing by precomposition, and
\[
\mathsf{Prod}:\mathcal A\to\mathbf{Cat},
\qquad
Y\longmapsto\mathcal A/Y,
\]
with reindexing by postcomposition, have the same total category:
\[
\int\mathsf{Rec}\cong\mathcal A^\to
\cong
\int\mathsf{Prod}.
\]
Under these identifications,
\[
\boxed{
\operatorname{dom}:\mathcal A^\to\to\mathcal A
}
\]
is the split Grothendieck fibration of receiver presentations, while
\[
\boxed{
\operatorname{cod}:\mathcal A^\to\to\mathcal A
}
\]
is the split Grothendieck opfibration of producer presentations.
\end{theorem}

\noindent\emph{Proof in Appendix~\ref{app:categorical-proof-details}.}

An arrow \(F:X\to Y\) has two roles:
\[
F\in X/\mathcal A,
\qquad
F\in\mathcal A/Y.
\]
It is a represented receiver of \(X\) and
a represented producer of \(Y\).
For composable arrows
\[
X\xrightarrow{F}Y\xrightarrow{G}Z,
\]
the interface \(Y\) is simultaneously
the output of \(F\) and the input of \(G\).

Fixing an ambient interface \(X\), its
producer contexts form the slice category
\[
\boxed{
\mathsf{Ctx}_X
=
\mathcal A/X.
}
\]
A downstream receiver belongs to
\(X/\mathcal A\). These two categories
provide the contexts compared in
Section~\ref{sec:contextual-search}.
In Part~\ref{part:trained-refinement},
the corresponding neural computations
supply forward features and backward
loss covectors to local parameter updates.

\subsection{Composition contracts interfaces}

A represented factorization contains more displayed architecture than its composite arrow. Consider
\[
X\xrightarrow{Q}Y\xrightarrow{G}Z.
\]
Composition gives
\[
X\xrightarrow{GQ}Z.
\]
The intermediate interface $Y$ is no longer an object of the displayed coarse segment, and the producer presentation $Q$ and receiver presentation $G$ that met at $Y$ are no longer separately visible.

Let $\mathcal A^{[2]}$ denote the category of composable pairs. Composition is a functor
\[
\boxed{
\mu:\mathcal A^{[2]}\to\mathcal A^\to,
\qquad
(Q,G)\longmapsto GQ.
}
\]
This is the precise sense in which formula contraction forgets interfaces.

There is then a second forgetful step:
\[
\mathcal A^{[2]}
\xrightarrow{\mu}
\mathcal A^\to
\xrightarrow{R^\to}
\mathcal S^\to.
\]
The first map contracts the chosen represented factorization. The second passes from represented architecture to extensional semantics. The two operations should not be conflated: even after a factorization has been contracted, $R$ may still identify distinct represented composite arrows.

Contraction is adequate for a scientific quantity
when that quantity can still be determined from
the retained coarse description. A boundary-to-boundary
output may descend through composition while a claim
about the producing features, receiving operations,
or local gradients does not.
Appendix~\ref{app:matched-effective-block} makes
this test explicit for a resolved Transformer block:
an observable is preserved by a matched contraction
when it factors through the retained block data.

\subsection{Cuts form factorization categories}

A cut through a fixed represented branch retains the producer and receiver factorization of that branch.

\begin{definition}[Factorization category]\label{def:factorization-category}
Fix a represented component
\[
B:X\to Z
\]
in $\mathcal A$. The category
\[
\boxed{
\mathsf{Fact}_{\mathcal A}(B)
}
\]
has as objects represented factorizations
\[
X\xrightarrow{Q}Y\xrightarrow{G}Z,
\qquad
GQ=B.
\]
A morphism
\[
T:(Q,G)\to(Q',G')
\]
is a morphism $T:Y\to Y'$ satisfying
\[
\boxed{
TQ=Q',
\qquad
G=G'T.
}
\]
\end{definition}

An object of $\mathsf{Fact}_{\mathcal A}(B)$ is a selected intermediate architectural cut of the represented branch $B$. An isomorphism in this category is a passive represented re-presentation of that cut that preserves the full branch on both sides.

The factorization category distinguishes a cut
in the represented computation from a merely
possible algebraic decomposition of its realized
function. One can factor $R(B)$ through a
mathematically convenient intermediate space
without that space being an exposed state of
the modeled process. An architectural cut instead
requires the intermediate carrier and its two
arrows to occur in $\mathcal A$ at the declared
grain. This is the condition under which a
factorization supports a claim about the
organization of the computation, rather than
only about a decomposition of its function.

For more than one intermediate cut, let $[n]$ denote the chain
\[
0\to1\to\cdots\to n.
\]
An object of $\mathcal A^{[n]}$ is a represented path
\[
X_0\xrightarrow{F_1}X_1
\xrightarrow{F_2}\cdots
\xrightarrow{F_n}X_n.
\]
Total composition defines
\[
\operatorname{comp}_n:
\mathcal A^{[n]}\to\mathcal A^\to,
\qquad
(F_1,\ldots,F_n)\longmapsto F_n\cdots F_1.
\]

The number and placement of retained
interfaces determine the granularity
of the represented description. Refining
that description retains additional
represented factors; coarsening it
composes adjacent factors. Associativity
makes repeated composition independent
of bracketing.

This choice of factorization granularity
is distinct from observational resolution.
Even after a cut has been selected,
different semantic maps or receiver
families can induce different
equivalence relations on its predecessor
states. The next section studies those
relations explicitly.

\subsection{Cuts along a represented process}
\label{subsec:cut-frames}

The factorization category also gives a derived object for moving a cut through one represented path. Let
\[
\Phi:
X_0\xrightarrow{F_1}X_1
\xrightarrow{F_2}\cdots
\xrightarrow{F_n}X_n
\]
be an object of $\mathcal A^{[n]}$, and let
\[
B:=F_n\cdots F_1.
\]
For $0\le k\le n$, define
\[
P_k:=F_k\cdots F_1,
\qquad
P_0:=\operatorname{Id}_{X_0},
\]
and
\[
S_k:=F_n\cdots F_{k+1},
\qquad
S_n:=\operatorname{Id}_{X_n}.
\]
Then
\[
S_kP_k=B.
\]
The cut at $X_k$ therefore determines the factorization object
\[
\boxed{
C_\Phi(k)
:=
\left(
X_0\xrightarrow{P_k}X_k\xrightarrow{S_k}X_n
\right)
\in
\mathsf{Fact}_{\mathcal A}(B).
}
\]

The object \(C_\Phi(k)\) is the represented
cut at \(X_k\). Its producer prefix
\[
P_k\in\mathcal A/X_k
\]
and receiver continuation
\[
S_k\in X_k/\mathcal A
\]
remain explicitly available.

For $k\le\ell$, write
\[
F_{k:\ell}
:=
F_\ell\cdots F_{k+1}:X_k\to X_\ell,
\qquad
F_{k:k}:=\operatorname{Id}_{X_k}.
\]
Then
\[
F_{k:\ell}P_k=P_\ell,
\qquad
S_k=S_\ell F_{k:\ell},
\]
so
\[
F_{k:\ell}:
C_\Phi(k)\to C_\Phi(\ell)
\]
is a morphism in $\mathsf{Fact}_{\mathcal A}(B)$. Hence the represented path induces a functor
\[
\boxed{
C_\Phi:
[n]\to
\mathsf{Fact}_{\mathcal A}(B).
}
\]
The functor \(C_\Phi\) records the
successive factorizations of the same
represented process. Moving from \(X_k\)
to \(X_{k+1}\) transfers \(F_{k+1}\)
from the receiver continuation to
the producer prefix. The composite
process is unchanged, while the
selected operation's surrounding
computation changes.

This construction will also apply to
internal Transformer cuts. Retaining
an intermediate interface makes it
possible to compare the features
arriving there, the continuation
receiving them, and the parameter
directions available at that location.
Composition removes that intermediate
description from the coarser segment.

\subsection{Joint interfaces are products in the receiver fiber}

Branching and residual topologies do not require a different ontology. Assume $\mathcal A$ has the finite products needed by the selected cuts. Given receiver maps
\[
F_i:X\to Y_i,
\qquad
i=1,\ldots,m,
\]
their jointly exposed cut is
\[
\langle F_1,\ldots,F_m\rangle:
X\to
\prod_{i=1}^mY_i.
\]
This is the categorical product of the $F_i$ in the coslice $X/\mathcal A$.

For the deterministic Set realization, assume that the relevant finite products are preserved by $R$. Then product presentation is sent to meet of distinction relations:
\[
\boxed{
\ker R\!\left(\langle F_1,\ldots,F_m\rangle\right)
=
\bigcap_{i=1}^m\ker R(F_i).
}
\]
The identity is the concrete Set coordinate chart for the categorical statement: two predecessor states agree at the joint cut exactly when they agree at every displayed component. No additional notion of joint distinction is needed.

\subsection{Locality is a factorization property}

Some architectural claims require more than the existence of a cut. Suppose a represented interface is a product
\[
X=\prod_{i\in S}X_i
\]
with projections $\pi_N$ onto selected subproducts. A receiver
\[
F_r:X\to Y_r
\]
is local to $N_r\subseteq S$ when it factors as
\[
\boxed{
F_r=f_r\pi_{N_r}.
}
\]
Locality is therefore not an additional type attached to the interface. It is a factorization property of an actual receiver arrow.

If
\[
X\xrightarrow F Y\xrightarrow G Z
\]
and both $F$ and $G$ admit declared projection factorizations, ordinary substitution determines the dependency support of $GF$. Affine block-sparsity is the corresponding linear specialization. These observations will be used only where a concrete neural factorization needs them; they are not part of the categorical foundation.

Realization retains the semantic map and induces the predecessor distinctions it identifies; it need not retain the represented producer/receiver factorization.

\section{Indistinguishability under realized maps}
\label{sec:receiver-types}

A realized receiver
\[
R(F):R(X)\to R(Y)
\]
identifies predecessor states that produce
the same output. Its kernel pair records
those identifications as an equivalence
relation on \(R(X)\).

This is a coarser description than
the represented arrow \(F\). It retains
which predecessor states are identified,
but not the realized output values,
the represented factorization, or
the receiving continuation.

\subsection{Internal equivalence relations as semantic distinction relations}

Assume throughout this section that $\mathcal S$ has finite limits. For $S\in\mathcal S$, let
\[
\mathsf{Eq}(S)
\]
denote the poset of internal equivalence relations on $S$, ordered by inclusion. We regard $\mathbf{Pos}$ as a subcategory of $\mathbf{Cat}$ when convenient.

A semantic map
\[
f:S\to T
\]
has a kernel pair
\[
\operatorname{KP}(f)
=
S\times_T S
\rightrightarrows S.
\]
In $\mathbf{Set}$ this is the ordinary equivalence relation
\[
\{(s,s'):f(s)=f(s')\}.
\]

\begin{definition}[Distinction relation]\label{def:distinction-shadow}
For a represented receiver
\[
F:X\to Y
\]
in $\mathcal A$, define its \emph{distinction relation} by
\[
\boxed{
\mathsf D_X(F)
:=
\operatorname{KP}(R(F))
\in
\mathsf{Eq}(R(X)).
}
\]
\end{definition}

The distinction relation forgets the represented output interface $Y$, the particular arrow $F$, any internal factorization of $F$, and any represented structure that the realization functor does not preserve. It remembers exactly which semantic predecessor states $R(F)$ identifies.

For deterministic Set semantics,
\[
\mathsf D_X(F)
=
\ker R(F)
=
\{(x,x'):R(F)(x)=R(F)(x')\}.
\]
The distinction order used later is therefore inherited from inclusion of equivalence relations. Finer semantic resolution means a smaller equivalence relation.

\subsection{The distinction relation is natural under receiver reindexing}\label{sec:type-fibration}

The receiver fibration from Theorem~\ref{thm:two-sided-arrow} reindexes a downstream component by ordinary precomposition. Internal equivalence relations reindex by pullback. These are the same operation after semantic realization.

For
\[
a:S'\to S
\]
in $\mathcal S$, pullback of relations defines
\[
a^\sharp:
\mathsf{Eq}(S)\to\mathsf{Eq}(S').
\]
Hence
\[
\mathsf{Eq}:\mathcal S^{op}\to\mathbf{Pos},
\]
and composition with $R$ gives
\[
\mathsf{Eq}_R
:=
\mathsf{Eq}\circ R^{op}
:
\mathcal A^{op}\to\mathbf{Pos}.
\]

A morphism
\[
h:F\to F'
\]
in the receiver fiber $X/\mathcal A$ satisfies
\[
F'=hF.
\]
Therefore
\[
\operatorname{KP}(R(F))
\subseteq
\operatorname{KP}(R(F')),
\]
so the distinction relation acts functorially on each receiver fiber.

\begin{theorem}[Natural distinction relation]\label{thm:shadow-naturality}
The kernel-pair assignment defines a natural transformation
\[
\boxed{
\mathsf D:
\mathsf{Rec}
\Longrightarrow
\mathsf{Eq}_R,
}
\]
where $\mathsf{Eq}_R$ is regarded as Cat-valued through the inclusion $\mathbf{Pos}\hookrightarrow\mathbf{Cat}$.

For every represented context
\[
A:\Omega\to X
\]
and receiver
\[
F:X\to Y,
\]
the naturality equation is
\[
\boxed{
\mathsf D_\Omega(FA)
=
R(A)^\sharp\mathsf D_X(F).
}
\]
\end{theorem}

\noindent\emph{Proof in Appendix~\ref{app:categorical-proof-details}.}

In Set notation the theorem reads
\[
\boxed{
\ker\bigl(R(F)R(A)\bigr)
=
(R(A)\times R(A))^{-1}\ker R(F).
}
\]
The equation states that composing a receiver
with an upstream producer pulls the receiver's
equivalence relation back to the predecessor
state space. The resulting relation depends
on both the receiver map and the producer
through which it is reached. It is the equivalence relation obtained by semantically realizing cartesian reindexing in the receiver fibration.

The diagram that controls the deterministic theory is therefore
\[
\begin{array}{ccc}
X/\mathcal A
&\xrightarrow{A^*}&
\Omega/\mathcal A
\\[1mm]
\downarrow\mathsf D_X
&&
\downarrow\mathsf D_\Omega
\\[1mm]
\mathsf{Eq}(R(X))
&\xrightarrow{R(A)^\sharp}&
\mathsf{Eq}(R(\Omega)).
\end{array}
\]
The top row keeps represented receiver roles and the arrows that realize them. The bottom row keeps only the semantic identifications induced on predecessor states.

\subsection{Kernel pairs along a represented path}
\label{subsec:information-frame}

The cut functor \(C_\Phi\) from
Section~\ref{subsec:cut-frames}
retains the producer and receiver
at each selected interface.
A coarser description records only
the equivalence relations induced
on the initial state space by
successive realized prefixes.

Fix
\[
B:X_0\to X_n.
\]
There is a prefix projection
\[
L_B:
\mathsf{Fact}_{\mathcal A}(B)
\to
X_0/\mathcal A,
\qquad
(P,S)\longmapsto P.
\]
A factorization morphism
\[
T:(P,S)\to(P',S')
\]
satisfies $TP=P'$, which is exactly a morphism $P\to P'$ in the coslice $X_0/\mathcal A$. Thus $L_B$ is a functor.

For a represented path $\Phi$, define its \emph{distinction profile}
\[
\boxed{
I_\Phi
:=
\mathsf D_{X_0}
\,L_B\,
C_\Phi
:
[n]\to
\mathsf{Eq}(R(X_0)).
}
\]
At cut $k$,
\[
\boxed{
I_\Phi(k)
=
\mathsf D_{X_0}(P_k)
=
\operatorname{KP}(R(P_k)).
}
\]
The functor \(I_\Phi\) records the
equivalence relations induced on the
initial predecessor states by successive
realized prefixes.

In deterministic Set semantics, write
\[
E_k:=I_\Phi(k)=\ker R(P_k).
\]
Because $I_\Phi$ is a functor from a chain into a poset,
\[
\boxed{
E_k\subseteq E_\ell
\qquad
(k\le\ell).
}
\]
Equivalently, once two initial states have become equal at one deterministic prefix, no later deterministic continuation can separate them again.

The represented cut
\[
C_\Phi(k)=(P_k,S_k)
\]
retains the producer, intermediate
interface, and continuation. Its
prefix kernel
\[
I_\Phi(k)=\operatorname{KP}(R(P_k))
\]
retains only the predecessor
identifications induced up to that cut.

Different represented cuts may therefore
have the same prefix kernel. The
coordinate-mixing example in
Section~\ref{sec:barriers} shows why the
continuation available after the cut
can still distinguish their computational
roles.

\subsection{Products become meets after deterministic realization}

The product structure described in Section~\ref{sec:components-grain} is compatible with the distinction relation. If a cut keeps several receiver states jointly available,
\[
F
=
\langle F_1,\ldots,F_m\rangle,
\]
then two predecessor states collide under $F$ exactly when they collide under every $F_i$.

The Set identity in Section~\ref{sec:components-grain} is the concrete form of
\[
\boxed{
\text{product in }X/\mathcal A
\quad\longmapsto\quad
\text{meet in }\mathsf{Eq}(R(X)).
}
\]
In \(\mathbf{Set}\), this meet
is intersection. The same operation
will define the common kernel
of a parameterized receiver family
in Section~\ref{sec:composition}.

\subsection{What the kernel pair does not determine}

The relation \(\mathsf D_X(F)\) is
computed after realization. It need
not determine \(F\), since the functor
\(R\) is not assumed faithful.
Even equality of realized maps,
\[
R(F)=R(G),
\]
does not imply equality of their
represented arrows.

Kernel-pair equality retains still less:
it determines which predecessor states
are identified, without specifying the
output presentation or the continuation
available there. It therefore answers
a question about retained distinctions,
not by itself a question about computational
role. Section~\ref{sec:barriers} shows
that two presentations with the same
distinction relation can support different
restricted continuations.
Appendix~\ref{sec:exact-quotients}
develops the corresponding exact-category
quotient construction.
\section{Retained distinctions and available continuation}\label{sec:barriers}

The categorical core separates three levels:
\[
\text{represented receiver in }\mathcal A
\longrightarrow
\text{semantic realization in }\mathcal S
\longrightarrow
\text{distinction relation in }\mathsf{Eq}.
\]
Task consequences begin only after additional metric, measurable, or probabilistic structure has been declared.

Two failures matter. The first is created by the induced equivalence relation itself. If an interface identifies two predecessor states that the target separates, no continuation receiving only that interface can repair the loss. The second occurs when the semantic state retains the needed distinctions but the represented architecture does not supply the conversion or continuation required to use them. The first depends only on the equivalence relation. The second is presentation sensitive.

\subsection{Target variation is a function of the distinction relation}

Work first in deterministic Set semantics and suppress the realization symbol as formalized in Appendix~\ref{sec:exact-quotients}. Let $E$ be an equivalence relation on $X$ and let
\[
f:X\to(W,d)
\]
be a target into a metric space.

\begin{definition}[Target variation]\label{def:target-variation}
Define
\[
\boxed{
\Delta_E(f)
:=
\sup_{(x,x')\in E}
d\bigl(f(x),f(x')\bigr).
}
\]
For a deterministic receiver $Q:X\to Z$, write
\[
\Delta_Q(f)
:=
\Delta_{\ker Q}(f)
=
\Delta_{\mathsf D_X(Q)}(f)
\]
when the induced distinction relation is understood.
\end{definition}

The analytic quantity is therefore attached to the induced equivalence relation, not to a chosen quotient carrier. Any two quotient presentations with the same distinction relation have the same target variation.

\paragraph{Target variation along successive prefixes.}
For a represented path \(\Phi\), the
kernel pairs of its prefixes form an
increasing sequence of equivalence
relations on the initial state space.
Write
\[
E_k:=I_\Phi(k),
\]
the target-relative quantity
\[
\boxed{
\Delta_k^\Phi(f)
:=
\Delta_{E_k}(f)
}
\]
tracks how much target variation has become hidden inside the equivalence classes induced by the prefix to cut $k$. Since
\[
E_k\subseteq E_\ell
\qquad
(k\le\ell),
\]
we have
\[
\boxed{
\Delta_k^\Phi(f)
\le
\Delta_\ell^\Phi(f).
}
\]
The sequence therefore identifies the first
represented cut at which the prefix equates
a pair of predecessor states whose target
values differ. For a specific pair of predecessor states one may write
\[
\tau_\Phi(x,x')
:=
\min\{k:(x,x')\in E_k\},
\]
with $\tau_\Phi(x,x')=\infty$ if the pair remains distinguished along the whole path. This is derived notation: the analytic consequence at each cut is supplied by the same deterministic barrier below.

\begin{theorem}[Deterministic distinction barrier]\label{thm:deterministic-barrier}
Let $Q:X\to Z$ be a deterministic realized receiver and let
\[
q_Q:X\twoheadrightarrow\operatorname{im}Q
\]
be its realized-image corestriction. Set
\[
E:=\ker Q=\ker q_Q.
\]
For any continuation
\[
G:\operatorname{im}Q\to W,
\]
\[
\boxed{
\sup_{x\in X}
d\bigl(f(x),Gq_Q(x)\bigr)
\ge
\frac12\Delta_E(f).
}
\]
Moreover, an exact continuation on the realized image exists if and only if
\[
\Delta_E(f)=0.
\]
\end{theorem}

\begin{proof}
Choose $(x,x')\in E$. Then $q_Q(x)=q_Q(x')$, so with $z=q_Q(x)$ the triangle inequality gives
\[
d(f(x),f(x'))
\le
d(f(x),G(z))
+
d(G(z),f(x')).
\]
At least one endpoint error is at least half the target separation. Taking the supremum over $E$ proves the bound.

Exact recovery is possible exactly when $f$ is constant on every $E$-class. By Theorem~\ref{thm:exact-quotient-classification}, this is exactly factorization through the effective quotient $q_Q$.
\end{proof}

The factor $1/2$ is genuinely metric. The categorical theory identifies the equivalence relation against which the target should be tested; the triangle inequality supplies the numerical lower bound.

If a represented factorization
\[
X\xrightarrow{P}\mathcal M\xrightarrow{\operatorname{Agg}}Z
\]
is contracted to
\[
Q=\operatorname{Agg}P,
\]
then
\[
\ker P\subseteq\ker Q.
\]
Hence
\[
\boxed{
\Delta_Q(f)\ge\Delta_P(f).
}
\]
The monotonicity of $\Delta_E(f)$ follows from inclusion of equivalence relations.

\subsection{Semantic transport and architectural availability}

The exact quotient theorem says that equal effective semantic relations determine isomorphic quotient presentations. Unrestricted continuations transport across that semantic isomorphism automatically.

Let $\mathcal S$ be locally small, let
\[
Q:X\twoheadrightarrow Z
\]
be an epimorphism, and let
\[
T:Z\xrightarrow{\cong}Z',
\qquad
Q'=TQ.
\]
For every output object $W$, precomposition with $T$ gives
\[
T^*:
\operatorname{Hom}_{\mathcal S}(Z',W)
\xrightarrow{\cong}
\operatorname{Hom}_{\mathcal S}(Z,W),
\qquad
g'\longmapsto g'T.
\]
Composition with $Q$ gives
\[
c_Q:
\operatorname{Hom}_{\mathcal S}(Z,W)
\to
\operatorname{Hom}_{\mathcal S}(X,W),
\qquad
g\longmapsto gQ.
\]
Because $Q$ is epi, $c_Q$ is injective.

For any restricted continuation subset
\[
\mathcal G\subseteq\operatorname{Hom}_{\mathcal S}(Z,W),
\]
we use
\[
\boxed{
\mathfrak B(Q,\mathcal G)
:=
\operatorname{im}\!\left(
\mathcal G
\hookrightarrow
\operatorname{Hom}_{\mathcal S}(Z,W)
\xrightarrow{c_Q}
\operatorname{Hom}_{\mathcal S}(X,W)
\right)
}
\]
only as shorthand for the complete branches attained by that restricted Hom-subfamily.

\begin{theorem}[Continuation transport]\label{thm:continuation-transport}
Let
\[
\mathcal G'\subseteq\operatorname{Hom}_{\mathcal S}(Z',W).
\]
Then
\[
\boxed{
\mathfrak B(Q',\mathcal G')
=
\mathfrak B(Q,T^*\mathcal G').
}
\]
For any
\[
\mathcal G\subseteq\operatorname{Hom}_{\mathcal S}(Z,W),
\]
epimorphism cancellation gives
\[
\boxed{
\mathfrak B(Q,\mathcal G)
=
\mathfrak B(Q',\mathcal G')
\iff
\mathcal G=T^*\mathcal G'.
}
\]
\end{theorem}

\begin{proof}
For every $g'\in\mathcal G'$,
\[
g'Q'
=
g'TQ,
\]
which proves the first equality. Since $Q$ is epi, $gQ=hQ$ implies $g=h$, so equality of attained branch images is equivalent to equality of the corresponding continuation subsets on $Z$.
\end{proof}

The proof uses functoriality and epi cancellation. Whether the semantic transport $T$ is available within the represented process is a separate question, as is closure of the permitted continuation family under $T^*$. Equal semantic quotients settle neither.

The theorem specifies when an invertible
semantic re-presentation is passive for
the computational comparison at hand.
The isomorphism transports all unrestricted
continuations, but preserves a comparison
of restricted attainable branches only
when the declared continuation families
satisfy $\mathcal G=T^*\mathcal G'$.
Mathematical recoverability therefore
need not amount to a permitted computation:
the two presentations support the same
restricted branches exactly when their
allowed continuation families match
under semantic transport.

\subsection{A dense mixing preserves every distinction and changes restricted accessibility}

In Set, consider
\[
Q(u,v)=(u,v),
\qquad
Q'(u,v)=(u+v,u-v).
\]
The linear map
\[
T(u,v)=(u+v,u-v)
\]
is invertible, so the two interfaces have the same distinction relation.

Restrict the next scalar receiver to one displayed coordinate:
\[
\mathcal G_{\mathrm{coord}}
=
\{\gamma\pi_1:\gamma:\mathbb R\to\mathbb R\}
\cup
\{\gamma\pi_2:\gamma:\mathbb R\to\mathbb R\}.
\]
For the target
\[
f(u,v)=u,
\]
we have
\[
f=\pi_1Q,
\]
so
\[
f\in\mathfrak B(Q,\mathcal G_{\mathrm{coord}}).
\]

The same target is not attainable from $Q'$ by a one-coordinate continuation. The first coordinate $u+v$ agrees at $(0,0)$ and $(1,-1)$ while $u$ differs; the second coordinate $u-v$ agrees at $(0,0)$ and $(1,1)$ while $u$ differs. Exact recovery requires
\[
\boxed{
u=\frac12(Q'_1+Q'_2),
}
\]
which mixes both displayed outputs.

Both maps preserve every predecessor
distinction. The restricted receiver
family nevertheless distinguishes
the presentations: recovering \(u\)
from \(Q'\) requires a coordinate
combination unavailable to that family.

This comparison has a decision-theoretic counterpart.
Blackwell's comparison of statistical experiments orders
observations by the decisions they support
\citep{blackwell1951comparison}. The presentations
$Q(u,v)=(u,v)$ and $Q'(u,v)=(u+v,u-v)$ are equivalent
when arbitrary measurable decision rules are admitted:
either observation can be recovered from the other.
The equivalence changes when the decision rule must
factor through one selected coordinate. In the present
example, the inverse mixing operation is not available
to that receiver. The two presentations therefore retain
the same predecessor information but support different
decisions under the declared computational restriction.

\subsection{Squared loss separates distinction loss from restricted use}

Let
\[
Q:X\to Z
\]
be measurable, let $H$ be $X$-valued, and write
\[
Z_H:=Q(H).
\]
Let $Y$ be a finite-dimensional Euclidean target with finite second moment. For a nonempty family of measurable square-integrable continuations $\mathcal G$, define
\[
m_Q
:=
\mathbb E[Y\mid\sigma(Z_H)].
\]

\begin{theorem}[Squared-loss decomposition]\label{thm:squared-decomposition}
Under the stated integrability assumptions,
\[
\boxed{
\inf_{g\in\mathcal G}
\mathbb E\|Y-g(Z_H)\|^2
=
\mathbb E\|Y-m_Q\|^2
+
\inf_{g\in\mathcal G}
\mathbb E\|m_Q-g(Z_H)\|^2.
}
\]
\end{theorem}

\noindent\emph{Proof in Appendix~\ref{app:categorical-proof-details}.}

The first term is the irreducible
squared prediction error given the
observation \(Z_H=Q(H)\).
For the fixed joint law, it is
determined by the conditional
expectation
\[
m_Q=\mathbb E[Y\mid\sigma(Z_H)]
\]
and the observable sigma-algebra
\(\sigma(Z_H)\), up to almost-sure
equivalence. Equality of
set-theoretic kernel relations alone
does not establish equality of these
measurable information structures
without additional assumptions.
The second term measures the additional error imposed by
the continuation family: the information is present in the
observation, but the specified receiver cannot necessarily
use it. Predictive $\mathcal V$-information makes receiver
restrictions part of the definition of usable information
\citep{xu2020usable}, while probing analyses examine
what information can be extracted by selected predictors
\citep{hewitt2021conditional}. Here the squared-loss
projection identity separates the two effects additively
for the declared target and continuation class. The same
separation will permit training-dependent changes in
receiver accessibility to be studied without attributing
them to a loss of predecessor distinctions.

\subsection{Bayes invariance under semantic re-presentation}

\begin{corollary}[Re-presentation invariance of the Bayes term]\label{cor:bayes-representation}
Suppose
\[
Q'=TQ
\]
where $T$ is a measurable bijection on $\operatorname{im}Q$ with measurable inverse. Then
\[
\sigma(Q(H))
=
\sigma(Q'(H)),
\]
and therefore
\[
\mathbb E\|Y-m_Q\|^2
=
\mathbb E\|Y-m_{Q'}\|^2.
\]
If continuation families also satisfy
\[
\mathcal G=T^*\mathcal G',
\]
their optimal squared losses are equal.
\end{corollary}

\begin{proof}
The measurable maps $T$ and $T^{-1}$ give both sigma-algebra inclusions. Equality of the Bayes terms follows. Under continuation transport, Theorem~\ref{thm:continuation-transport} gives the same attainable predictors, so the approximation terms in Theorem~\ref{thm:squared-decomposition} also agree.
\end{proof}

Mutual semantic recoverability therefore settles the information term. Architectural equivalence of restricted continuations requires the stronger transport condition.

During training, the receiver continuation is the actual downstream Transformer suffix. Its derivative pulls the loss covector to the retained cut, where it pairs with the producer feature in Equation~\eqref{eq:training-producer-receiver-gradient}. A distinction retained in the forward state influences the local parameter gradient through this receiver-dependent pairing.

\section{Static parameter families under composition}\label{sec:composition}

A represented process may supply several receiver arrows at one interface, indexed by a set of admissible configurations. The index may describe parameter values, component choices, or other declared receiver settings; it need not arise from neural training. Each setting selects a receiver arrow in the same receiver fiber. We first compare this family statically, without placing any morphisms or an update law between settings. Neural parameter states supply the concrete family in Part~II.

\subsection{Parameterized receiver families}

Let $P$ be a parameter set and let
\[
\Theta:=\operatorname{Disc}(P)
\]
be the corresponding discrete category.

\begin{definition}[Parameterized receiver family]\label{def:parameter-family}
A parameterized receiver family at interface $X$ is a functor
\[
\boxed{
\mathcal Q:\Theta\to X/\mathcal A,
\qquad
\theta\longmapsto Q_\theta:X\to Z_\theta.
}
\]
\end{definition}

An upstream context
\[
A:\Omega\to X
\]
reindexes the family:
\[
A^*\mathcal Q:
\Theta\to\Omega/\mathcal A.
\]
Applying the natural distinction relation gives the static diagram
\[
\boxed{
\Theta
\xrightarrow{A^*\mathcal Q}
\Omega/\mathcal A
\xrightarrow{\mathsf D_\Omega}
\mathsf{Eq}(R\Omega).
}
\]

After the producer context \(A\) is fixed,
the represented receiver family induces a
family of equivalence relations on the
predecessor state space. Its image records
exactly which distinction profiles are
realized as the receiver parameters vary:
\[
\boxed{
\mathcal C_A(\mathcal Q)
:=
\operatorname{im}\!\left(
\mathsf D_\Omega A^*\mathcal Q
\right).
}
\]
In Set,
\[
\mathcal C_A(\mathcal Q)
=
\{\ker(Q_\theta A):\theta\in P\},
\]
with duplicates removed.

Naturality of $\mathsf D$ transports these images under further context restriction. The Set facts from Section~\ref{sec:exact-quotients} explain the two extreme cases used later: surjective contexts cannot merge distinct ambient relations, while injective contexts may make ambient relations agree on the reachable image without identifying predecessor states.

\subsection{The common kernel of a receiver family}

Assume the required meet exists in
\(\mathsf{Eq}(R\Omega)\).
The common kernel is the meet of
the equivalence relations induced
by the receiver family. Define
\[
\boxed{
E_{A,\mathcal Q}
:=
\bigwedge_{\theta\in P}
\mathsf D_\Omega(Q_\theta A).
}
\]
In $\mathbf{Set}$,
\[
\boxed{
E_{A,\mathcal Q}
=
\bigcap_{\theta\in P}
\ker(Q_\theta A).
}
\]

Only when a quotient representative is useful do we choose the effective quotient
\[
J_{A,\mathcal Q}:
R(\Omega)\twoheadrightarrow
R(\Omega)/E_{A,\mathcal Q}.
\]
The quotient map represents the meet; it is not the fundamental family object.

The universal property is immediate:
\[
\boxed{
E\subseteq
\mathsf D_\Omega(Q_\theta A)
\quad\forall\theta
\iff
E\subseteq E_{A,\mathcal Q}.
}
\]
The image \(\mathcal C_A(\mathcal Q)\) records
the repertoire of predecessor-distinction
relations realized by the receiver family.
The meet \(E_{A,\mathcal Q}\) records only
those predecessor identifications shared
by every family member. The meet therefore
suffices for uniform information barriers,
while the image retains which different
distinction patterns the architecture can
realize as its admissible receiver setting
varies. Families with the same image have
the same meet; families with the same meet
need not have the same image.
Section~\ref{sec:contextual-search}
accordingly compares nominal receiver
architectures through their images,
not merely through their common kernels.

\subsection{Uniform factorization is the meet universal property}

Let
\[
P_0:\Omega\twoheadrightarrow U
\]
be a semantic quotient associated with an earlier represented interface and write
\[
E_{P_0}:=\ker P_0.
\]
By exact quotient classification,
\[
Q_\theta A=R_\theta P_0
\qquad\forall\theta
\]
is equivalent to
\[
E_{P_0}
\subseteq
\mathsf D_\Omega(Q_\theta A)
\qquad\forall\theta.
\]
Hence the entire family factors through $P_0$ exactly when
\[
\boxed{
E_{P_0}
\subseteq
E_{A,\mathcal Q}.
}
\]
Parameter-uniform factorization follows from the meet universal property and exact quotient classification.

The useful consequences below are genuinely analytic.

\begin{corollary}[Uniform deterministic barrier]\label{cor:uniform-deterministic}
Assume every effective family interface factors through $P_0$. For branches
\[
B_\theta=G_\theta Q_\theta A
\]
and target
\[
f:\Omega\to(W,d),
\]
\[
\boxed{
\sup_\omega
d\bigl(
f(\omega),
B_\theta(\omega)
\bigr)
\ge
\frac12\Delta_{E_{P_0}}(f)
\qquad\forall\theta.
}
\]
\end{corollary}

\begin{proof}
Each branch factors through the effective quotient with relation $E_{P_0}$. Apply Theorem~\ref{thm:deterministic-barrier}.
\end{proof}

No parameter choice can restore a predecessor distinction absent before the parameterized family.

\begin{corollary}[Uniform Bayes-risk floor]\label{cor:uniform-bayes}
Let $H$ be $\Omega$-valued and $Y$ finite-dimensional Euclidean-valued with finite second moment. Assume $P_0$ and the effective interfaces are measurable and the family factorizations through $P_0$ are measurable. With
\[
m_{P_0}
=
\mathbb E[Y\mid\sigma(P_0(H))],
\]
every measurable square-integrable continuation after every effective family interface satisfies
\[
\boxed{
\mathbb E\|Y-\widehat Y\|^2
\ge
\mathbb E\|Y-m_{P_0}\|^2.
}
\]
\end{corollary}

\begin{proof}
Every such predictor is a measurable function of $P_0(H)$. Conditional expectation is the $L^2$ projection onto $\sigma(P_0(H))$.
\end{proof}

\subsection{Approximate family bounds require metric structure}

Exact factorization can be relaxed when effective family states vary only slightly inside one $P_0$-class. Give each effective interface codomain a metric $d_\theta$ and define
\[
\omega_{P_0}(Q_\theta A)
:=
\sup_{P_0(\omega)=P_0(\omega')}
d_\theta\bigl(
Q_\theta A(\omega),
Q_\theta A(\omega')
\bigr).
\]
This vanishes in the exact factorization regime.

\begin{theorem}[Approximate contextual barrier]\label{thm:approx-context}
Suppose
\[
G_\theta:
(\operatorname{im}(Q_\theta A),d_\theta)
\to(W,d)
\]
is $L_\theta$-Lipschitz and
\[
L_\theta\,
\omega_{P_0}(Q_\theta A)
\le\eta
\qquad\forall\theta.
\]
Then
\[
\boxed{
\sup_\omega
d\bigl(
f(\omega),
G_\theta Q_\theta A(\omega)
\bigr)
\ge
\frac12
\bigl(
\Delta_{E_{P_0}}(f)-\eta
\bigr)_+
\qquad\forall\theta.
}
\]
\end{theorem}

\noindent\emph{Proof in Appendix~\ref{app:categorical-proof-details}.}

This theorem marks the analytic boundary of the categorical reduction. Exact equivalence relations no longer determine the claim; distances in the presented states and regularity of the actual continuation now matter.

The next subsection gives a concrete
sufficient condition under which
every receiver in a parameterized
family induces the same equivalence
relation on predecessor states.
\subsection{A sufficient condition for equal receiver kernels}
\label{subsec:receiver-sufficient-prefix}

One neural pattern is useful enough to keep a name. An earlier interface can already determine the complete state reaching a downstream family while every downstream receiver retains a displayed component carrying exactly the same predecessor distinctions.

\begin{definition}[Receiver-sufficient prefix]\label{def:receiver-sufficient-prefix}
Let
\[
P:\Omega\twoheadrightarrow U,
\qquad
A:\Omega\to X,
\qquad
L:X\to V
\]
be deterministic Set realizations of represented interfaces. We call $P$ \emph{receiver-sufficient for $A$ relative to $L$} when
\[
A=\widetilde A P
\]
for some $\widetilde A:U\to X$ and
\[
\ker(LA)=\ker P.
\]
\end{definition}

Suppose each family member retains that receiver coordinate:
\[
Q_\theta(x)
=
\bigl(L(x),E_\theta(x)\bigr).
\]
Write
\[
E_P:=\ker P,
\qquad
E_\theta:=\ker(Q_\theta A).
\]
Factorization through $P$ gives
\[
E_P\subseteq E_\theta,
\]
while retention of $LA$ gives
\[
E_\theta\subseteq\ker(LA)=E_P.
\]
The two inclusions therefore imply
\[
\boxed{
E_P
\subseteq
E_\theta
\subseteq
E_P
\quad\Longrightarrow\quad
E_\theta=E_P
\qquad\forall\theta.
}
\]

\begin{proposition}[Receiver-sufficient family collapse]\label{prop:receiver-sufficient-collapse}
Under the preceding hypotheses,
\[
\boxed{
\mathsf D_\Omega(Q_\theta A)=E_P
\qquad\forall\theta.
}
\]
Consequently,
\[
\boxed{
\mathcal C_A(\mathcal Q)=\{E_P\},
\qquad
E_{A,\mathcal Q}=E_P.
}
\]
Under exact semantics, every realized-image quotient $Q_\theta A$ is therefore uniquely isomorphic over $\Omega$ to the quotient presentation of $P$.
\end{proposition}

The proposition records the neural sufficient condition; the mathematics is the displayed absorption law.

If $P$ is injective, then
\[
E_P=\Delta_\Omega.
\]
The same sandwich gives
\[
\mathsf D_\Omega(Q_\theta A)=\Delta_\Omega
\qquad\forall\theta,
\]
so every effective family interface is injective as well. Because
\[
\ker(LA)=\Delta_\Omega,
\]
the retained coordinate $LA$ is injective, and therefore $A$ is injective.

This is the regime used by the main local-versus-attention witness. The prefix preserves every predecessor state, yet the downstream family can still collapse at the distinction grain because contextual restriction changes which ambient receiver relations remain different on the reachable image.

Nothing in receiver sufficiency privileges a neural layer. If a multi-block prefix is contracted to one represented arrow $A$, the same condition is tested at that cut. If the prefix is refined into several factors, the factorization category from Section~\ref{sec:components-grain} records the intermediate cuts at which the condition may first hold.

\section{Context-dependent receiver equivalence}
\label{sec:contextual-search}

A nominal receiver architecture specifies
a family of permitted operations at an
input interface. Its effective distinction
profile also depends on which states an
upstream process presents to that interface.
Two receiver families that differ on the
full input carrier can agree on one
producer's reachable states and disagree
on another's. The classification problem
is therefore to determine when distinct
nominal architectures count as the same
receiver architecture at the declared
distinction grain.

Fix an ambient receiver interface \(X\),
a set \(D\) of nominal downstream
architecture labels, and a producer
\(A:\Omega\to X\). For each label we
retain the family of predecessor
equivalence relations induced after
composition with \(A\). We then derive
contextual equivalence of labels from
equality of these family profiles.
This makes the criterion of effective
architectural identity explicit before
asking how its classes vary across
producer contexts.

\subsection{Contexts are already the producer fiber}

Fix an ambient downstream input interface $X$. Section~\ref{sec:components-grain} already supplied the context category:
\[
\boxed{
\mathsf{Ctx}_X
=
\mathsf{Prod}(X)
=
\mathcal A/X.
}
\]
An object is a represented prefix
\[
A:\Omega\to X.
\]
A context morphism
\[
\kappa:A'\to A
\]
is a represented map
\[
\kappa:\Omega'\to\Omega
\]
with
\[
A'=A\kappa.
\]
Such a context morphism records
further precomposition on the
producer side.
It need not increase informational
resolution or preserve predecessor
distinctions. Receiver families on
\(X\) are correspondingly reindexed
from \(X/\mathcal A\) to
\(\Omega/\mathcal A\), and their
induced equivalence relations are
pulled back along \(R(A)\).

Thus the producer slice and receiver
coslice already supply the two sides
of the contextual comparison.

\subsection{A presheaf of contextual receiver semantics}

Let $D$ be a set of nominal downstream schema labels. For each schema $d\in D$ and each aligned receiver index $j$ used in the comparison, let $P_{d,j}$ be its parameter set, let
\[
\Theta_{d,j}:=\operatorname{Disc}(P_{d,j}),
\]
and let
\[
\mathcal Q_{d,j}:\Theta_{d,j}\to X/\mathcal A
\]
be its static parameterized receiver family.

At a context
\[
A:\Omega\to X,
\]
the family is reindexed and reduced to its distinction relations:
\[
\Theta_{d,j}
\xrightarrow{A^*\mathcal Q_{d,j}}
\Omega/\mathcal A
\xrightarrow{\mathsf D_\Omega}
\mathsf{Eq}(R\Omega).
\]
Its image is the derived family profile
\[
\mathcal C_A(\mathcal Q_{d,j})
\subseteq
\mathsf{Eq}(R\Omega).
\]

For a context morphism
\[
\kappa:A'\to A,
\qquad
A'=A\kappa,
\]
naturality of $\mathsf D$ sends each relation
\[
E\in\mathsf{Eq}(R\Omega)
\]
to
\[
R(\kappa)^\sharp E
\in
\mathsf{Eq}(R\Omega').
\]
Applying this relation pullback elementwise to subsets gives the restriction map needed below.

\paragraph{Size convention.}
Fix a universe in which the context
category and indexed receiver families
used below are small. Assume that the
relevant collections
\(\mathsf{Eq}(R\Omega)\) of internal
equivalence relations are sets and
are stable under the specified
pullbacks. This holds in ordinary
\(\mathbf{Set}\) semantics; in a
general semantic category it follows,
for example, from the corresponding
well-poweredness and universe
assumptions.

\begin{definition}[Contextual receiver-kernel presheaf]
\label{def:semantic-profile-presheaf}
For a fixed finite or otherwise declared collection of aligned receiver indices $j$, define
\[
\boxed{
\mathsf{Sem}(A)
:=
\prod_j
\mathcal P\!\left(
\mathsf{Eq}(R\Omega)
\right)
}
\]
for each context $A:\Omega\to X$.

For
\[
\kappa:A'\to A,
\]
define
\[
\mathsf{Sem}(\kappa):
\mathsf{Sem}(A)\to\mathsf{Sem}(A')
\]
by applying
\[
R(\kappa)^\sharp
\]
elementwise in every receiver coordinate. Then
\[
\boxed{
\mathsf{Sem}:
\mathsf{Ctx}_X^{op}\to\mathbf{Set}
}
\]
is a presheaf.
\end{definition}

\begin{proof}
Identity pullback is the identity on internal equivalence relations, and
\[
R(\kappa\lambda)^\sharp
=
R(\lambda)^\sharp R(\kappa)^\sharp
\]
for composable context maps. Elementwise application preserves those equalities in each powerset coordinate, hence in the product.
\end{proof}

The presheaf records, for each producer
context, the sets of predecessor
equivalence relations realized by
the declared receiver families.
It does not retain the receivers'
full output maps, their represented
factorizations, or additional
continuations.

\subsection{Nominal architecture labels acquire semantics through one natural transformation}

Let
\[
\underline D:
\mathsf{Ctx}_X^{op}\to\mathbf{Set}
\]
be the constant presheaf with value $D$.

At each context $A:\Omega\to X$, define
\[
\Sigma_A:D\to\mathsf{Sem}(A)
\]
by
\[
\boxed{
\Sigma_A(d)
:=
\bigl(
\mathcal C_A(\mathcal Q_{d,j})
\bigr)_j.
}
\]
Thus $\Sigma_A(d)$ is the complete distinction-family profile assigned to nominal schema $d$ at the declared aligned receivers in context $A$.

\begin{theorem}[Architecture semantics transformation]\label{thm:architecture-semantics}
The maps $\Sigma_A$ assemble into a natural transformation
\[
\boxed{
\Sigma:
\underline D
\Longrightarrow
\mathsf{Sem}.
}
\]
For every context morphism
\[
\kappa:A'\to A,
\]
\[
\boxed{
\Sigma_{A'}
=
\mathsf{Sem}(\kappa)\Sigma_A.
}
\]
\end{theorem}

\noindent\emph{Proof in Appendix~\ref{app:categorical-proof-details}.}

This natural transformation generates the architecture-space theory. Contextual label equivalence is its kernel, effective contextual architecture is its image, and their reindexing follows from its naturality.

\subsection{Contextual equivalence is the kernel of the semantics map}

For a fixed context $A$, define
\[
d\sim_A d'
\]
only as shorthand for equality under $\Sigma_A$:
\[
\boxed{
d\sim_A d'
\quad\Longleftrightarrow\quad
\Sigma_A(d)=\Sigma_A(d').
}
\]
Equivalently, $\sim_A$ is the kernel pair of the Set map
\[
\Sigma_A:D\to\mathsf{Sem}(A).
\]

The equivalence \(d\sim_A d'\)
compares nominal receiver architectures
only through the declared families of
predecessor-kernel relations after
composition with producer \(A\).
It does not assert equality of their
represented operations, parameter
spaces, optimization dynamics,
computational costs, or unrestricted
receiver capabilities.

This family-level equivalence is
also different from equality of
receiver kernels at individual
parameter states. Part~%
\ref{part:trained-refinement}
uses the latter when following a
particular learned computation.

For a context morphism
\[
\kappa:A'\to A,
\qquad A'=A\kappa,
\]
naturality gives
\[
\Sigma_{A'}
=
\mathsf{Sem}(\kappa)\Sigma_A.
\]
Consequently,
\[
\boxed{
d\sim_A d'
\quad\Longrightarrow\quad
d\sim_{A'}d'.
}
\]

Receiver schemas that agree on the
predecessor distinctions induced in
one context continue to agree after
further precomposition. Additional
schema pairs may become equivalent
because the new producer presents
a more restricted family of
reachable states.

\subsection{The effective architecture presheaf is an image}

The presheaf category
\[
[\mathsf{Ctx}_X^{op},\mathbf{Set}]
\]
has pointwise images. Factor $\Sigma$ there as
\[
\underline D
\twoheadrightarrow
\mathfrak A_D
\hookrightarrow
\mathsf{Sem}.
\]

\begin{definition}[Effective architecture presheaf]\label{def:effective-architecture-presheaf}
Define
\[
\boxed{
\mathfrak A_D
:=
\operatorname{im}\Sigma.
}
\]
Pointwise,
\[
\boxed{
\mathfrak A_D(A)
=
\operatorname{im}\Sigma_A.
}
\]
\end{definition}

Because images in a Set-valued presheaf category are computed pointwise, the ordinary quotient-image correspondence gives
\[
\boxed{
\mathfrak A_D(A)
\cong
D/\ker\Sigma_A
=
D/\!\sim_A.
}
\]
The effective architecture quotient is therefore derived from the semantics map. The primary object is the realized image of nominal architecture labels in contextual semantics.

A context-independent downstream quotient is a special case: it requires $\mathfrak A_D$ to be label-compatibly equivalent to a constant presheaf on the contexts under consideration. If the image fibers change, no single quotient of $D$ describes effective downstream identity throughout that region of context space.

\paragraph{Totalization.}
The standard Grothendieck construction totalizes the image presheaf into a discrete fibration over producer contexts; Appendix~\ref{sec:exact-quotients} gives the construction. Its fibers are the effective architecture classes defined by \(\Sigma\) in each context.

\subsection{A continuous injective witness}\label{subsec:attention-context-witness}

To calculate how producer context changes downstream architectural identity, compare a local receiver and a one-head attention receiver under two injective presentations of the same predecessor state.

Let
\[
V=\mathbb R^2,
\qquad
\Omega=\mathbb R^2,
\qquad
X=V^2.
\]
Write
\[
H=(h_1,h_2),
\qquad
h_1,h_2\in V.
\]
Define two contexts
\[
\boxed{
A_{\mathrm{enc}}(u,v)
=
\bigl((u,v),(0,0)\bigr)
}
\]
and
\[
\boxed{
A_{\mathrm{sep}}(u,v)
=
\bigl((u,0),(0,v)\bigr).
}
\]
Both maps are injective.

For receiver $1$, let the local schema have the one-member family
\[
Q_{\mathrm{loc}}(H)=h_1.
\]
Let the attention schema use the coarse residual receiver family
\[
Q^{\mathrm{att}}_\theta(H)
=
\bigl(h_1,t_{1,\theta}(H)\bigr),
\]
where $t_{1,\theta}$ is a one-head attention output acting on the displayed two-token state. The family contains a zero-value state $\theta_0$, for example $W_V=0$, with
\[
t_{1,\theta_0}(H)=0,
\]
and a uniform-average state $\theta_{\mathrm{avg}}$ with
\[
t_{1,\theta_{\mathrm{avg}}}(H)
=
\frac{h_1+h_2}{2}.
\]
In this unnormalized two-token attention schema, the latter is realized by
\[
W_Q=W_K=0,\qquad W_V=I_V,
\]
with identity output map, zero bias, and both sources admissible. Appendix~\ref{app:dimension-labelled-transformer} specifies the complete PreNorm operator, which normalizes tokens before value projection. Theorem~\ref{thm:training-induced-receiver-separation} proves the corresponding contextual separation using that exact normalized operator and exhibits a gradient-descent update between the contexts.

We use one aligned receiver, so $\Sigma_A(d)$ is simply the image set
\[
\mathcal C_A(\mathcal Q_d)
\subseteq
\mathsf{Eq}(\Omega).
\]

\begin{theorem}[Injective-context nonconstancy]\label{thm:injective-attention-nonconstant}
For the two-schema label set
\[
D=\{\mathrm{loc},\mathrm{att}\},
\]
both $A_{\mathrm{enc}}$ and $A_{\mathrm{sep}}$ are injective, but
\[
\boxed{
|\mathfrak A_D(A_{\mathrm{enc}})|=1,
\qquad
|\mathfrak A_D(A_{\mathrm{sep}})|=2.
}
\]
Hence the two fibers are not label-compatibly isomorphic, and the effective architecture presheaf $\mathfrak A_D$ is nonconstant even across these information-preserving contexts.
\end{theorem}

\begin{proof}
Under $A_{\mathrm{enc}}$,
\[
Q_{\mathrm{loc}}A_{\mathrm{enc}}(u,v)
=
(u,v)
\]
is injective. Every attention state retains this same local vector as its first output component, so every
\[
Q^{\mathrm{att}}_\theta A_{\mathrm{enc}}
\]
is also injective. Therefore
\[
\mathcal C_{A_{\mathrm{enc}}}(\mathcal Q_{\mathrm{loc}})
=
\mathcal C_{A_{\mathrm{enc}}}(\mathcal Q_{\mathrm{att}})
=
\{\Delta_\Omega\}.
\]
Thus
\[
\Sigma_{A_{\mathrm{enc}}}(\mathrm{loc})
=
\Sigma_{A_{\mathrm{enc}}}(\mathrm{att}),
\]
and the image has one element.

Under $A_{\mathrm{sep}}$,
\[
Q_{\mathrm{loc}}A_{\mathrm{sep}}(u,v)
=
(u,0),
\]
so its distinction relation identifies exactly states with the same $u$.

For $\theta_0$,
\[
Q^{\mathrm{att}}_{\theta_0}A_{\mathrm{sep}}(u,v)
=
\bigl((u,0),(0,0)\bigr),
\]
which has the same distinction relation. For $\theta_{\mathrm{avg}}$,
\[
Q^{\mathrm{att}}_{\theta_{\mathrm{avg}}}A_{\mathrm{sep}}(u,v)
=
\left(
(u,0),
\left(\frac u2,\frac v2\right)
\right),
\]
which is injective: the first component determines $u$ and the second coordinate of the attention output determines $v$. Hence
\[
\mathcal C_{A_{\mathrm{sep}}}(\mathcal Q_{\mathrm{att}})
\]
contains both the local $u$-partition and the diagonal relation, while
\[
\mathcal C_{A_{\mathrm{sep}}}(\mathcal Q_{\mathrm{loc}})
\]
contains only the former. Therefore
\[
\Sigma_{A_{\mathrm{sep}}}(\mathrm{loc})
\ne
\Sigma_{A_{\mathrm{sep}}}(\mathrm{att}),
\]
and the image has two elements.
\end{proof}

The nominal local and attention receiver
families are held fixed, and both producers
preserve every predecessor distinction.
Under $A_{\mathrm{enc}}$, the local token
already carries $u$ and $v$, so the two
receiver families have the same distinction
profile. Under $A_{\mathrm{sep}}$, attention
can acquire $v$ from the second token
while the local receiver cannot.
Thus the effective equivalence class
of a receiver schema is not determined
by the nominal schema together with
the information preserved upstream.
The producer presentation determines
which distinctions the fixed receiver
family can use.

The result makes the admissible comparison of computational
roles depend on the represented producer context. Causal
abstraction compares lower-level mechanisms with higher-level
algorithms using intervention-sensitive correspondences
\citep{geiger2025causalabstraction}. The freedom allowed
in constructing such correspondences matters: sufficiently
expressive nonlinear alignment maps can trivialize tests
of mechanistic correspondence
\citep{sutter2025nonlinear}. In the present theorem, the
receiver families and their labels are fixed before the
producer context is varied. Both producers remain injective,
yet the resulting contextual profiles have different
cardinalities. An unrestricted change of coordinates could
transport information between the presentations, but it
would also change the receiving computation being compared.
The nonconstancy of the effective architecture presheaf
records this dependence exactly.

The Set specialization in Section~\ref{sec:exact-quotients} explains the general mechanism. Pullback along an injection is surjective on equivalence relations but need not be injective. An injective context preserves its predecessor states while restricting the ambient region on which downstream relations are compared. Different ambient receiver relations can therefore agree on one embedded image and disagree on another.

\subsection{A lossy context converts the same calculus into a task barrier}

The injective witness establishes nonconstancy of architecture semantics. A separate lossy context shows how the same distinction machinery produces quantitative task limitations.

Restrict the predecessor domain to
\[
\Omega_{\mathrm b}=\{0,1\}^2
\]
and define
\[
A_{\mathrm b}(u,v)
=
\bigl((u,0),(u,0)\bigr).
\]
Let
\[
P_{\mathrm b}(u,v)=u.
\]
The complete reachable state factors through $P_{\mathrm b}$, and the retained local token has the same induced distinction relation. Hence $P_{\mathrm b}$ is receiver-sufficient.

\begin{proposition}[Broadcast collapse and exact task barrier]\label{prop:broadcast-barrier}
For every attention parameter state,
\[
\ker(Q^{\mathrm{att}}_\theta A_{\mathrm b})
=
\ker P_{\mathrm b}.
\]
Therefore
\[
\mathcal C_{A_{\mathrm b}}(\mathcal Q_{\mathrm{att}})
=
\{\ker P_{\mathrm b}\},
\qquad
E_{A_{\mathrm b},\mathcal Q_{\mathrm{att}}}
=
\ker P_{\mathrm b}.
\]

For target
\[
f(u,v)=v
\]
with absolute error, every continuation after every such attention interface satisfies
\[
\boxed{
\sup_{(u,v)\in\Omega_{\mathrm b}}
|v-\widehat v_\theta(u,v)|
\ge
\frac12.
}
\]
If $(u,v)$ is uniform on $\Omega_{\mathrm b}$ and $Y=v$, every measurable squared-loss predictor after the family satisfies
\[
\boxed{
\mathbb E|Y-\widehat Y_\theta|^2
\ge
\frac14.
}
\]
Both constants are sharp at the information level.
\end{proposition}

\begin{proof}
Receiver sufficiency gives
\[
\ker(Q^{\mathrm{att}}_\theta A_{\mathrm b})
=
\ker P_{\mathrm b}
\]
by Proposition~\ref{prop:receiver-sufficient-collapse}. Within each $P_{\mathrm b}$-fiber, $v$ takes both values $0$ and $1$, so
\[
\Delta_{P_{\mathrm b}}(f)=1.
\]
The deterministic bound follows from Theorem~\ref{thm:deterministic-barrier}.

Under the uniform distribution,
\[
\mathbb E[Y\mid u]=\frac12,
\qquad
\operatorname{Var}(Y\mid u)=\frac14.
\]
The Bayes floor follows from Corollary~\ref{cor:uniform-bayes}. Constant prediction $1/2$ attains both information-level bounds.
\end{proof}

The injective and broadcast examples answer different questions. The first proves that contextual architecture identity can change without information loss. The second shows the quantitative consequence when a context actually removes a task-relevant distinction. Keeping them separate is essential to the architecture/information distinction developed throughout the paper.

\section{Computational identity and contextual architecture}
\label{sec:part1-conclusion}

A represented process supports several different
comparisons. Equality of its realized composite
concerns the boundary-to-boundary function;
equality of a prefix kernel concerns the
predecessor distinctions identified at a
selected cut. Neither comparison determines
the represented factorization or the
restricted continuations available from its
intermediate interface. The continuation-transport
theorem gives the condition under which
a semantic re-presentation also preserves
those restricted computations.

The effective architecture construction
applies this distinction to nominal receiver
types. A semantics map assigns each type
its family of predecessor-distinction
relations in a producer context; its kernel
determines contextual equivalence of the
labels, and its image determines the
effective architecture classes. The injective
local--attention witness establishes
\[
\boxed{
\ker A_{\mathrm{enc}}
=
\ker A_{\mathrm{sep}}
=
\Delta_\Omega,
\qquad
|\mathfrak A_D(A_{\mathrm{enc}})|
\ne
|\mathfrak A_D(A_{\mathrm{sep}})|.
}
\]
The producer changes effective receiver
equivalence without destroying any
predecessor distinction. Nominal architectural
identity and effective architectural identity
at the declared comparison grain are
therefore different.

This also specifies what it means for
a computational description to be sufficient.
A quotient can retain every distinction
required by one target and still omit
the operations needed by a restricted
receiver. A composite function can determine
the output and still omit the represented
cuts needed to explain how that output
was produced. Coarsening is justified by
the scientific quantity that survives it,
rather than by functional equality alone.

The theory therefore does not seek a uniquely
privileged decomposition of a computation.
Different represented descriptions can be
adequate for different scientific questions
without being interchangeable. Their adequacy
is determined by which computational relations
and observables survive the proposed
identifications. The appropriate criterion
of identity is consequently part of the
scientific investigation, rather than a
context-independent premise imposed before it.

Part~\ref{part:trained-refinement} introduces
additional structure: a concrete neural
parameterization, an implemented optimizer,
and the resulting transition law.
The nominal Transformer architecture
can remain unchanged while training
alters producer presentations, receiver
accessibility, and the future behavior
of the computational roles established here.

\part{Training-Dependent Computational Semantics in Transformers}
\label{part:trained-refinement}

\section{From contextual computation to learned computational roles}
\label{sec:part2-intro}

Part~\ref{part:represented-architecture} studies represented producer--receiver relations without assuming a neural network or training rule. We now return to the deep-learning system that motivates DLE and supply the additional structure needed for a dynamical analysis: a concrete Transformer, parameter space, data/target law, differentiation, and an implemented optimizer. At a fixed cut the producer presents an activation; the continuation determines which distinctions can be used downstream and propagates the loss covector acting on the parameters. Learning changes these presentations and continuations while the declared wiring can remain unchanged.

The evolution of a realized function is already known to
depend on how that function is parameterized. Deep linear
networks exhibit nonlinear, depth-dependent learning
trajectories despite representing linear maps
\citep{saxe2014exact}, and depth can alter optimization
through the factorization of those maps
\citep{arora2018acceleration}. Neural tangent kernel theory
expresses another aspect of this dependence: parameter
gradients induce a geometry governing movement in function
space \citep{jacot2018ntk}. We will encounter these effects
inside a resolved Transformer. The additional observable
tracked here is the computational relation at a selected
cut: training changes the features arriving at a receiver,
the continuation acting on them, and the distinctions
available under its specified computational restrictions.

Fix the dimension-labelled PreNorm residual Transformer of Appendix~\ref{app:dimension-labelled-transformer}, a data/target law, and an implemented optimizer. Depth, carrier widths, heads, masks, parameter sharing, and arithmetic remain fixed during each training experiment. A cut $c$ therefore has two descriptions: the family of computational relations available to its receiver architecture, and the actual relation realized at the current parameter state. The changing latter description is the subject of this part.

The forward and reverse computations
produce the update signals, and the
causal optimizer turns them into a
closed-loop training kernel.

We begin with a two-token PreNorm
receiver whose computational relations
change when a declared upstream
producer is trained. Its matrix-valued
extension gives different learning
rates to different represented
directions. These are exact
learning experiments for the specified
producer and receiving interface,
not general training laws for
arbitrary Transformer stacks.

We then train a Q/K coordinate in an
explicit attention realization and
derive its source-allocation dynamics.
Coupled producer--attention learning
and an input-conditioned construction
show how both parameter updates and
the presented input features change
receiver accessibility.

Finally, we determine which
complete-state distinctions predict
these contextual changes, before
comparing different architectures
under training.

\section{Learned contextual semantics of a represented cut}
\label{sec:learned-cut-semantics}

Fix a Transformer architecture, including its carrier types, parameter-sharing relations, and implemented data interface. At an internal cut $c$, a compatible parameter state $\theta$ determines a producer $P_{c,\theta}:\mathcal X\to X_c$, the local represented computation, and the downstream continuation $S_{c,\theta}:X_c\to Y$. These are restrictions of \emph{one} globally parameterized network. Their roles depend on the cut: moving the cut can transfer an operation from the receiver continuation into the producer history without changing the complete network.

Part~\ref{part:represented-architecture} compares \emph{families} of admissible receiver processes. Its context-indexed object $\Sigma_P(d)$ concerns the semantics of every receiver admitted by the nominal architecture $d$ after producer $P$. Training, in contrast, visits particular compatible parameter states $\theta_t$. The effective role of the resulting individual receiver is not identified with the image of the entire parameter family. Both descriptions are useful: one describes the repertoire available to an architecture under a context, the other the computation realized at a particular training state.

For a nonempty predecessor domain
\(D\subseteq\mathcal X\) and a
receiver probe
\(Q_{c,\theta}:X_c\to V_c\),
define the receiver kernel on \(D\) by
\begin{equation}
\label{eq:statewise-context-shadow}
E^D_{c,\theta,Q}
=
\ker\bigl(
(Q_{c,\theta}\circ P_{c,\theta})|_D
\bigr).
\end{equation}

This kernel records exact
indistinguishability at the selected
receiver. It does not determine the
complete represented cut or its
available continuations.

Bounded-receiver error measures
a different property: what a
restricted continuation can recover
from the presented state.
Predictive information introduces
the training kernel and asks
which present-state distinctions
determine future values of a
chosen observable.

For a finite declared test domain $D=\{x_1,\ldots,x_N\}$ and a finite family $\mathcal Q$ of typed receiver probes, the concrete profile
\begin{equation}\label{eq:finite-semantic-probe}
\sigma_{c,D,\mathcal Q}(\theta)
=\Bigl(\bigl(Q_{c,\theta}(P_{c,\theta}(x_i))\bigr)_{Q\in\mathcal Q,\,i\le N},
\bigl(\mathbf1\{QP_{c,\theta}(x_i)=QP_{c,\theta}(x_j)\}\bigr)_{Q,i,j}\Bigr)
\end{equation}
is a measurable observable whenever the evaluated producer and receiver maps depend measurably on $\theta$ in their finite-dimensional carriers. One may instead retain only its equivalence matrices to study distinctions. A finite probe is a specified grain of computational semantics, not a reconstruction of every represented relation.

For a complete training state $z=(\theta,\omega)$, put $\sigma_c(z)=\sigma_{c,D,\mathcal Q}(\theta)$ or choose another explicitly measurable contextual probe. Section~\ref{sec:training-closed-loop} constructs the complete transition kernel $K_t$, which then generates the law of the evolving semantic profile. The question is how the actual training updates change the producer and receiver contexts appearing in this observable.

\section{The resolved Transformer as a training-time system}
\label{sec:transformer-interface-calculus}

The Transformer attention mechanism originates with
\citet{vaswani2017attention}; placing LayerNorm inside
the residual branches yields the PreNorm organization
studied by \citet{xiong2020layernorm}.
Appendix~\ref{app:dimension-labelled-transformer}
specifies the dimension-labelled PreNorm system used here,
including its operation types and wiring.
Its residual carrier is
\[
X_\ell=(\R^d)^S,
\]
with global width \(d\) forced by the skip topology. Block \(\ell\) has arbitrary declared head count \(h_\ell\), query/key dimension \(q_\ell\), value dimension \(v_\ell\), feed-forward hidden depth \(n_\ell\), common hidden width \(w\), activation maps, masks, temperatures, biases, LayerNorm parameters, and parameter-sharing relations. No relation such as \(h_\ell v_\ell=d\) is imposed unless it is explicitly part of the architecture coordinate.

For one untied block, the exact count of parameter slots is derived from these typed maps:
\[
\boxed{
\begin{aligned}
P_\ell
={}&
2h_\ell d(q_\ell+v_\ell)
+
2h_\ell q_\ell
+
h_\ell v_\ell
\\
&+
2dw
+
(n_\ell-1)w^2
+
n_\ell w
+
6d
+
P^\phi_\ell.
\end{aligned}}
\]
Appendix~\ref{app:exact-block-parameter-count} derives every slot, including biases and LayerNorm parameters. If tensors share coordinates within a block, the slot count must be quotiented by the corresponding parameter identifications to obtain the number of independent trainable variables. When block parameter coordinates and boundary coordinates are disjoint, the complete model count is
\[
P_{\mathrm{tot}}=P_\partial+\sum_{\ell=0}^{L-1}P_\ell.
\]
For tied parameters, this sum counts coordinate occurrences rather than independent trainable coordinates. The exact parameter count is the dimension of the globally shared parameter locus $\Theta_{\mathrm{tot}}$; Appendix~\ref{app:exact-block-parameter-count} makes that distinction explicit. Scalar parameter count is a many-to-one
summary of the architecture. It does
not determine the internal operator
factorization, the exposed carriers,
or the parameter derivatives entering
training.

Proposition~\ref{prop:fine-structure-not-recoverable}
constructs two distinct PreNorm block
architectures with
\[
d=2,\qquad h=q=n=1,
\]
and internal dimensions
\[
(v,w)=(1,2)
\qquad\text{and}\qquad
(v,w)=(2,1).
\]
Both have exactly \(33\) parameters and admit
parameter states realizing the identity map,
yet their internal value and feed-forward
carriers differ. Thus the realized block map,
even together with its parameter count,
does not determine the represented
architecture.

The residual stream is the carried state of the block. Attention and feed-forward computation branch away from that carrier and return through explicit residual additions. This wiring matters because a branch can reorganize, compress, or expose information without yet replacing the complete residual state. Whole-state distinction and branch-relative accessibility therefore evolve on different schedules inside the same block.

\begin{figure}[tb]
\centering
\begin{tikzpicture}[
 x=1cm,y=1cm,>=Latex,
 stage/.style={draw,rounded corners=2pt,fill=black!3,minimum height=6mm,align=center,font=\small,inner xsep=4pt},
 plus/.style={draw,circle,inner sep=1.6pt,font=\small},
 state/.style={font=\small},
 flow/.style={->,thick},
 carry/.style={->,line width=1.15pt}
]
\node[state] (H) at (0,0) {$H$};
\node[stage] (ln1) at (1.6,1.05) {LN$_1$};
\node[stage] (att) at (4.0,1.05) {Q/K/V, scores,\\softmax, aggregation};
\node[plus] (sum1) at (6.25,0) {$+$};
\node[state] (U) at (7.1,0) {$U$};
\node[stage] (ln2) at (9.0,1.05) {LN$_2$};
\node[stage] (ffn) at (11.25,1.05) {FFN hidden\\and output};
\node[plus] (sum2) at (13.25,0) {$+$};
\node[state] (Hp) at (14.4,0) {$H^+$};
\draw[carry] (H) -- (sum1);
\draw[flow] (H.north) |- (ln1.west);
\draw[flow] (ln1) -- (att);
\draw[flow] (att.east) -| (sum1.north);
\draw[carry] (sum1) -- (U) -- (sum2);
\draw[flow] (U.north) |- (ln2.west);
\draw[flow] (ln2) -- (ffn);
\draw[flow] (ffn.east) -| (sum2.north);
\draw[carry] (sum2) -- (Hp);
\end{tikzpicture}
\caption{Two residual merges in one PreNorm Transformer block. At a fine cut inside either branch, the carried residual remains separately present. Whole-state identification can occur at the two merges, while normalization, attention routing, hidden FFN computation, and their receiving gradients have finer, receiver-relative semantics. Backpropagation transports covectors through the reversed represented graph.}
\label{fig:resolved-prenorm-graph}
\end{figure}

The usual block formula is already a coarse object. A PreNorm block contains a finite represented computation graph with LayerNorm suboperations, affine Q/K/V maps, scores, softmax, source-resolved weighted values, source aggregation, head projection, head aggregation, the first residual merge, the full feed-forward branch, and the second residual merge. Appendix~\ref{app:maximal-cut-relativity} resolves this graph into the distributive lattice
\[
\mathfrak F_\ell=J(\mathcal O_\ell)
\]
of legal live frontiers. Each frontier \(I\in\mathfrak F_\ell\) determines an exact factorization
\[
X_\ell
\xrightarrow{Q_{\ell,I}}
X_{\ell,I}
\xrightarrow{S_{\ell,I}}
X_\ell,
\qquad
S_{\ell,I}Q_{\ell,I}=B_\ell.
\]
The corresponding cut is therefore an object of \(\operatorname{Fact}_{\mathcal A}(B_\ell)\), not an auxiliary visualization of the block.

At every frontier, the represented
factorization supplies a producer prefix
and receiver suffix. The associated
semantic data include the complete-state
kernel pair, its exact quotient, the
permitted continuation family, target
variation, and the induced parameter-family
relations. Where differentiability is
declared, the frontier also determines
the pullback of downstream loss covectors.

Appendix~\ref{app:maximal-cut-relativity}
retains these quantities across all
legal frontiers:
\[
\Lambda_\ell^{\max}
=
\bigl(
\Sigma_{\ell,I}^{\mathrm{cut}}
\bigr)_{I\in\mathfrak F_\ell}.
\]
The retained data distinguish branch
access, complete residual-state
persistence, and the local information
needed for gradient calculations.

\subsection{Residual carries localize whole-state information loss}
\label{sec:residual-carry-profile}

The skip topology makes the whole-state distinction profile unusually rigid. Before the first residual merge, every frontier contains the original residual state \(H_\ell\) as a displayed factor. The frontier prefix therefore has the form
\[
H\longmapsto(H,\eta(H))
\]
and is injective. After the first residual merge and before the second, every frontier contains
\[
U_\ell=R_{\ell,1}(H_\ell)
\]
as a displayed factor and therefore has the same kernel as the first residual-stage map. The terminal frontier has the kernel of the complete block.

Writing
\[
R_{\ell,1}(H)=H+T_\ell(H),
\qquad
R_{\ell,2}(U)=U+F_\ell(U),
\qquad
B_\ell=R_{\ell,2}R_{\ell,1},
\]
the exact result is
\begin{theorem}[Exact whole-state plateaus in a PreNorm block]\label{thm:one-block-whole-state-plateaus}
\begin{equation}
\label{eq:prenorm-information-profile}
\boxed{
\begin{aligned}
E^{\mathrm{whole}}_{\ell,I}&=\Delta_{X_\ell}
&&\text{before the first residual merge},\\
E^{\mathrm{whole}}_{\ell,I}&=\ker R(R_{\ell,1})
&&\text{between the two residual merges},\\
E^{\mathrm{whole}}_{\ell,\mathrm{out}}&=\ker R(B_\ell)
&&\text{after the second residual merge}.
\end{aligned}}
\end{equation}
There are at most two strict whole-state distinction-growth events in one block, at its residual merges.
\end{theorem}
Appendix~\ref{app:maximal-cut-relativity} gives the complete frontier proof.

At arbitrary fine grain, LayerNorm, Q/K/V projection, score formation, softmax, source aggregation, head aggregation, and hidden feed-forward computation can change branch-relative distinctions, metric geometry, and receiver accessibility. They do not create a new complete-state equivalence class while the relevant residual carrier is still explicitly present. Exact whole-state distinction can enlarge at no more than two events per block: the two residual additions.

The residual-stream account of Transformer circuits treats
attention and feed-forward components as operations reading
from and writing to a common carried representation
\citep{elhage2021framework}. The frontier calculation gives
that account an exact distinction-level consequence. A
projection or normalization may identify branch inputs,
but its identification does not become a whole-state
identification while the bypassed residual state remains
available. The residual additions are precisely the
frontiers where the carried complete state changes.
Consequently, the number of internal branch contractions
and the number of complete-state distinction changes
measure different properties of the same block.

Thus
\[
\boxed{
\text{branch inaccessibility}
\neq
\text{whole-state information loss}.
}
\]
This is the concrete Transformer instance of the receiver relativity established in Part~\ref{part:represented-architecture}.

\subsection{Every internal frontier is a cut of the complete Transformer}
\label{sec:block-cut-insertion}

The fine block analysis is already global. Let
\[
A_\ell=B_{\ell-1}\cdots B_0,
\qquad
S_{\ell+1}=B_{L-1}\cdots B_{\ell+1}.
\]
An internal frontier \((Q_{\ell,I},S_{\ell,I})\) is inserted into the complete model as
\[
\boxed{
C_{\ell,I}^{[0,L-1]}
=
\bigl(
Q_{\ell,I}A_\ell,
S_{\ell+1}S_{\ell,I}
\bigr).
}
\]
The producer history and receiver continuation of the local computation therefore depend on where the block sits in the full Transformer. For an interior block of a segment containing at least three blocks, both sides change simultaneously when the surrounding context changes. Appendix~\ref{app:maximal-cut-relativity} gives the exact two-block, three-block, and arbitrary finite-segment formulas.

At a differentiable output-gated frontier, the corresponding local gradient has the producer--receiver form
\[
\widehat G_{\ell,I}^{[0,L-1]}
=
\lambda_{\ell,I}^{[0,L-1]}
\phi_{\ell,I}
\bigl(A_\ell(H_0)\bigr)^\top,
\]
where the feature comes from the producer prefix and the covector is pulled backward through the complete receiver suffix. The full reverse-mode derivatives, including score and softmax branches, are developed in Section~\ref{sec:resolved-training-derivatives}.

\subsection{Fine-cut gradients and training observations}

The optimizer receives observations generated by a training protocol: data, loss, gradients, and any auxiliary signals admitted by the declared interface. A branch can identify information while the complete residual stream still distinguishes its inputs. Its gradients nevertheless depend on that branch because the downstream loss covector passes through its receiver path. Appendix~\ref{app:maximal-cut-relativity} retains the finite-cut and multiblock compositions; the following derivatives turn those represented relations into the actual observations entering the optimizer.

\section{Fine cuts, residual information, and multiblock context}
\label{sec:transformer-multiblock-context}

The represented graph distinguishes five mechanisms that the ordinary block formula contracts into a single arrow. LayerNorm identifies shifts in the all-ones direction; projected query/key coordinates have transformations preserving their score pairing; softmax identifies rowwise constant changes of scores; source and head aggregation identify separately represented contributions having the same sum; and residual addition identifies different bypass--branch pairs with the same merged state. These identifications occur at different interfaces. A branch's information quotient therefore does not determine the distinction relation of the full carried residual state.

For a stabilized LayerNorm with nonvanishing coordinate gains, at the prescribed centered-input grain,
\[
N_\varepsilon(x)=N_\varepsilon(x')\quad\Longleftrightarrow\quad x'-x\in\operatorname{span}\{\mathbf1\}.
\]
At a softmax score row,
\[
\operatorname{softmax}(s)=\operatorname{softmax}(s')
\quad\Longleftrightarrow\quad s'-s=c\mathbf1.
\]
At a source-aggregation cut, a collection of source contributions $(z_j)_j$ is replaced by $\sum_jz_j$; the lost distinction is any zero-sum variation $\sum_j\delta z_j=0$. In every case the operation and the receiver grain, rather than the word ``information,'' specify the equivalence being measured. Appendix~\ref{app:maximal-cut-relativity} derives the exact kernels and compatibility conditions for the fully typed attention and FFN interfaces.

The one-block plateau theorem now extends
across a sequence of represented blocks.
Internal branch contractions remain distinct
from complete-state identifications, and
the residual merges delimit the possible
growth events of the whole-state distinction
relation.

For a segment of \(N\) blocks, the local
plateau structure therefore composes into
at most \(2N+1\) whole-state distinction
plateaus, while the producer and receiver
contexts of each internal frontier acquire
the surrounding block prefixes and suffixes.

\begin{theorem}[Cut relativity across a Transformer segment]
\label{thm:n-block-cut-relativity}
For a segment $\mathcal B_{a:b}=B_b\cdots B_a$, an internal frontier $(Q_{j,I},S_{j,I})$ of block $j$ has producer and receiver
\[
P_{j,I}^{[a,b]}=Q_{j,I}A_{a:j},\qquad
S_{j,I}^{[a,b]}=S_{j:b}S_{j,I},
\]
where $A_{a:j}=B_{j-1}\cdots B_a$ and $S_{j:b}=B_b\cdots B_{j+1}$, with identity maps for empty products. Its predecessor distinction relation is the pullback
\[
E_{j,I}^{[a,b]}=R(A_{a:j})^\sharp\ker R(Q_{j,I}).
\]
Downstream composition changes the receiver continuation and, on differentiable sectors, pulls back the loss covector through the later blocks. For an $N=b-a+1$ block segment, every full carried frontier belongs to one of $2N+1$ whole-state distinction plateaus; hence there are at most $2N$ strict whole-state identification events.
\end{theorem}

The pullback follows by taking the kernel of $Q_{j,I}A_{a:j}$. The receiver expression follows by composing the remaining suffix. The plateau count follows by applying Theorem~\ref{thm:one-block-whole-state-plateaus} successively to each block; the complete cut construction and associated target-variation equations appear in Appendix~\ref{app:maximal-cut-relativity}.

\subsection{Three blocks expose both external contexts}

Consider a cut $I$ inside the middle block of three consecutive PreNorm blocks. Its complete context is
\begin{equation}\label{eq:main-three-block-context}
\boxed{\ C_{\ell+1,I}^{(3)}=\bigl(Q_{\ell+1,I}B_\ell,\;B_{\ell+2}S_{\ell+1,I}\bigr).\ }
\end{equation}
The internal operation has not changed. The preceding block changes its producer, and the following block changes its receiver. At an output-gated differentiable cut, the same organization appears in the parameter gradient:
\begin{equation}\label{eq:main-three-block-gradient}
\widehat G_{\ell+1,I}^{(3)}=
\left[DR(S_{\ell+1,I})^*DR(B_{\ell+2})^*\lambda_{\ell+3}\right]
\phi_{\ell+1,I}(B_\ell(H_\ell))^\top.
\end{equation}
The learned upstream block determines the feature supplied to the held-fixed middle component; the learned downstream block determines the covector acting back on that component. Appendix~\ref{app:three-block-cut-relativity} proves the general interface formulas and extends them across arbitrary segments. These are two distinct ways of changing the role of the fixed middle operation: an upstream update changes the features on which it acts, whereas a downstream update changes which features matter to the loss. Their product is the local gradient passed to the optimizer in Section~\ref{sec:training-closed-loop}.

\section{Forward and backward organization of a PreNorm block}
\label{sec:resolved-training-derivatives}

At each internal cut, the forward computation produces the values consumed by the next operation. Reverse differentiation propagates the downstream loss covector through the same cut. Tied parameters receive the sum of these contributions from every occurrence. The resulting gradient is an observable of the complete represented computation, with its producer and receiver fixed by the cut.

\subsection{The two residual stages and their Jacobians}
\label{sec:training-block-jacobian}
Write a block on the residual carrier $X=(\mathbb R^d)^S$ as
\begin{equation}\label{eq:train-block-factorization}
U=H+A_\theta(H),\qquad H^+=U+F_\theta(U),\qquad B_\theta=({\rm Id}+F_\theta)\circ({\rm Id}+A_\theta).
\end{equation}
Here $A_\theta$ includes the first PreNorm, query/key/value production, masked scores, softmax, source and head aggregation, and attention output projection. The map $F_\theta$ includes the second PreNorm and the declared FFN. These names abbreviate the explicit operator sequence in Appendix~\ref{app:dimension-labelled-transformer}; they do not replace that sequence with an arbitrary two-map network.

On a differentiable parameter/input sector, the state Jacobian factors exactly:
\begin{equation}\label{eq:train-block-J}
D_HB_\theta=(I+D_UF_\theta)(I+D_HA_\theta).
\end{equation}
For an output covector $\lambda^+$ the covector entering the first stage is
\begin{equation}\label{eq:train-block-adjoint}
\lambda_H=(I+D_HA_\theta)^*(I+D_UF_\theta)^*\lambda^+.
\end{equation}
The order matters: forward composition uses the first stage followed by the second, and reverse differentiation traverses those stages in reverse. Nondifferentiable activation boundaries require a declared derivative or subgradient convention in the observation law; these formulas apply on the differentiable sectors.

The residual identity terms make the backward computation different from differentiating an isolated attention or FFN branch. They also explain the whole-state information plateaus of Section~\ref{sec:residual-carry-profile}. The complete carried state remains an explicit factor before each merge, but the gradient delivered to a branch depends on its receiving suffix and can vanish even where that branch still computes a nonconstant map.

\subsection{The attention gradient follows the resolved graph}
\label{sec:training-attention-derivative}
For one head, suppress the head/block indices and write, on an admissible source row,
\begin{equation}\label{eq:training-attention-forward}
q_i=W_Qn_i+b_Q,\quad k_j=W_Kn_j+b_K,\quad v_j=W_Vn_j+b_V,\quad
p_{ij}=\frac{\exp(s_{ij})}{\sum_{j'\in M_i}\exp(s_{ij'})},\quad
z_i=\sum_{j\in M_i}p_{ij}v_j,
\end{equation}
where $n_i$ is the first normalized token, the score $s_{ij}$ contains the declared query/key pairing, temperature, and mask, and $M_i$ is the allowed source set. Equation~\eqref{eq:training-attention-forward} compresses only notation: the weighted values $p_{ij}v_j$ remain separate produced quantities at their fine cuts.

Let $\bar z_i$ be the covector arriving from the head-output receiver. The direct source and routing variations are
\begin{equation}\label{eq:training-attention-pullback}
\bar v_j=\sum_i p_{ij}\bar z_i,\qquad
\bar p_{ij}=\langle \bar z_i,v_j\rangle,\qquad
\bar s_{ij}=p_{ij}\left(\bar p_{ij}-\sum_{j'\in M_i}p_{ij'}\bar p_{ij'}\right).
\end{equation}
The score derivative then propagates to $q_i,k_j$ through the declared bilinear score law, and Q/K/V affine derivatives propagate to normalized token features and shared weight matrices. For any shared parameter $\vartheta$, its gradient is the sum over all uses of $\vartheta$ in the unrolled graph. This is an exact reverse-mode calculation, not a statement that a scalar attention matrix summarizes the complete training interface.

The softmax identity $\sum_j\bar s_{ij}=0$ follows immediately. An additive common shift of a row of logits is unobservable to the forward conditional and annihilated by its local differential. Whether a parameter transformation generating such a forward invariance also preserves the optimizer update is a separate question, taken up in Section~\ref{sec:predictive-symmetry}.

\subsection{The producer--receiver pairing at output-gated cuts}
\label{sec:backprop-cut-frame}
At an internal cut with producer prefix $P$ and receiver suffix $S$, let $h=P(x)$ be the produced state and $\lambda=D S(h)^*\lambda_{\rm out}$ the receiving covector. For a linear output gate $W$ applied to a feature $\phi(h)$, the local parameter derivative takes the form
\begin{equation}\label{eq:training-producer-receiver-gradient}
\boxed{\nabla_W\widehat\ell=\lambda\,\phi(h)^\top.}
\end{equation}

\begin{example}[Changing both external contexts changes a local training signal]
\label{ex:two-sided-gradient-contexts}
Let $x=y=e_1\in\R^2$ and consider a trainable local gate $W\in\R^{2\times2}$ inside
\[
F_{P,S,W}(x)=SWPx,\qquad
\ell(W)=\tfrac12\lVert SWPx-y\rVert^2,
\]
where $P,S$ are fixed invertible producer and receiver matrices. For every $P,S$, the composite predictor at $W=0$ is the zero map, yet
\begin{equation}\label{eq:two-sided-gradient-example}
\nabla_W\ell=S^\top(SWPx-y)(Px)^\top,
\qquad \nabla_W\ell\big|_{W=0}=-(S^\top y)(Px)^\top.
\end{equation}
Writing $J=\left(\begin{smallmatrix}0&1\\1&0\end{smallmatrix}\right)$, four contexts give
\[
\begin{array}{c|cc}
 &S=I&S=J\\\hline
P=I &-e_1e_1^\top&-e_2e_1^\top\\
P=J &-e_1e_2^\top&-e_2e_2^\top
\end{array}.
\]
All four gates start at the same parameter $W=0$ and realize the same composite function, but their learning signals point along different parameter directions. The producer changes the feature $Px$; the receiver changes the covector $S^\top y$. This elementary gate isolates the same two-sided dependence that the resolved Transformer produces with normalized attention, residual propagation, and shared-parameter accumulation.
\end{example}

The feature and the covector are generated on opposite sides of the selected cut. When a gate is tied across sites or layers, the complete gradient is the sum of such contributions over its occurrences. Equation~\eqref{eq:training-producer-receiver-gradient} is a specific output-gate formula; general Q/K parameters involve the additional routing derivatives in \eqref{eq:training-attention-pullback}.

For a full $L$-block model, the prefix and suffix modify the internal cut to\[\bigl(Q_{\ell,I}B_{\ell-1}\cdots B_0,\; B_{L-1}\cdots B_{\ell+1}S_{\ell,I}\bigr).\]Therefore two equal internal vectors in two different network contexts can supply different gradients, even if their immediate branch map agrees. This is Part I's producer/receiver dependence manifested in the learning observation process.

These forward/backward observations feed the optimizer. Its state is determined by which distinctions between past observations change its future actions.

\section{The optimizer and the closed-loop training process}
\label{sec:training-closed-loop}

Fix a depth $L$, dimension-labelled architecture coordinate $c_L$, data/target law, loss/readout, initialization rule, and arithmetic implementation. Together they specify the training experiment. The current parameter state determines the forward Transformer computation and its backward signals through Section~\ref{sec:resolved-training-derivatives}. Sampling data, random masks where declared, and stochastic automatic differentiation yield the optimizer observation kernel
\begin{equation}\label{eq:training-observation-kernel}
\Gamma^S_{L,c_L,t}:\Theta_{L,c_L}\rightsquigarrow\mathsf{Obs}_{L,c_L}.
\end{equation}
For SGD-like experiments a realization of this observation contains a sampled loss and gradient. For other protocols it may also contain auxiliary statistics, control signals, and explicit numerical-implementation information. The observation type must be fixed before any optimizer law is defined.

\subsection{A general optimizer is a causal transducer}
\label{sec:causal-optimizer-main}

The architecture determines what training can observe; the optimizer determines how that observed history is turned into action. A typed optimizer scheme $\mathbf O$ assigns a causal stochastic response law to every declared observation/action type. For a pre-action history
\[
h_t=(o_0,u_0,\ldots,o_{t-1},u_{t-1}),
\]
the optimizer is a family of kernels
\[
\boxed{
\mathbf O_t:
\mathsf H_t\times\mathsf{Obs}
\rightsquigarrow
\mathsf{Act},
\qquad
u_t\sim\mathbf O_t(\cdot\mid h_t,o_t).
}
\]
The rule is causal: the next action depends only on the observed past and present, never on future observations. SGD, momentum, Adam/AdamW, clipping rules, adaptive preconditioners, schedule-dependent methods, and long-memory schemes are different causal transducers on declared interfaces. The architecture class is unchanged.

Optimizer state is derived from this behavior. Two admissible histories are equivalent when every admissible future observation continuation induces the same conditional law of all future optimizer actions:
\[
h\sim_{\mathbf O}h'
\quad\Longleftrightarrow\quad
\mathbf O_h^{[k]}(\cdot\mid\mathbf o)
=
\mathbf O_{h'}^{[k]}(\cdot\mid\mathbf o)
\]
for every finite horizon \(k\) and every admissible future observation string \(\mathbf o\). The optimizer causal-state space is
\[
\boxed{
\Omega_{\mathbf O}
=
\mathsf H/\!\sim_{\mathbf O}.
}
\]

This is the predictive-state construction familiar from
input--output computational mechanics. An
$\epsilon$-transducer identifies input--output histories
when they induce the same conditional future output laws
under future input sequences
\citep{barnett2015transducer}. Here the inputs are
admissible future optimizer observations and the outputs
are optimizer actions. The equivalence
$\sim_{\mathbf O}$ therefore retains exactly the past
distinctions that can change future optimizer behavior.
Appendix~\ref{app:optimizer-causal-state-E} proves the
corresponding canonical state realization for the
declared optimizer interface: future actions and the
next causal state factor through
$\omega=[h]_{\mathbf O}$. Every exact optimizer-state
realization must distinguish histories belonging to
different causal-state classes.

The notation $\Omega_{\mathbf O}$ therefore has a derived meaning. It is not a container into which unspecified persistent variables are placed. For memoryless SGD the behavioral state can be trivial once the external schedule is included in the observation interface. Momentum requires the velocity distinction because it changes future actions under the same future gradients. Adam requires the moment distinctions for the same reason. Different implementations count as the same optimizer state exactly when no admissible future training signal can make them act differently.

\subsection{Architecture and optimizer determine the closed-loop training process}

Let
\[
\mathfrak R_{L,c_L}(\theta,u)
\]
be the implemented parameter update, including the declared arithmetic and numerical implementation. The complete derived training state is
\[
\mathcal Z_{L,c_L}^{\mathbf O}
=
\Theta_{L,c_L}\times\Omega_{\mathbf O}.
\]
The observation kernel \(\Gamma\), optimizer action kernel, causal-state update, and implementation compose to one closed-loop kernel
\[
K^{\mathbf O}_{L,c_L,t}:
\mathcal Z_{L,c_L}^{\mathbf O}
\rightsquigarrow
\mathcal Z_{L,c_L}^{\mathbf O}.
\]
Appendix~\ref{app:derived-closed-loop-training-E} gives the explicit integral formula. Once the architecture coordinate, data/target law, initialization protocol, implementation, observation interface, and optimizer scheme are fixed, the learning experiment is determined. Training convergence is therefore a consequence of the resulting closed-loop dynamics.

\paragraph{Data and controller state.} The transition kernel below includes every persistent variable that affects future observation or action laws. If minibatches and stochastic masks are freshly drawn from a law conditional on $(\theta,t)$, they enter the observation kernel $\Gamma_t(o\mid\theta)$ directly. If the sampler, environment, or controller has persistent internal memory, its behaviorally relevant state is part of $Z$ and its update is composed into $K_t$. Omitting persistent exogenous state does not give a Markov kernel on $(\theta,\omega)$ alone.

\subsection{The exact observable law is not generally autonomous}
\label{sec:training-observable-law}

Fix a measurable evaluation receiver $\Phi:Z\to E$ on the complete closed-loop state $Z=\Theta_{L,c_L}\times\Omega_{\mathbf O}$. Examples include population risk, a finite vector of task responses, the predictor modulo the declared task law, or a fine-cut statistic. The current observation law is $\zeta_t=\Phi_\#\mu_t$, where $\mu_{t+1}=\mu_tK_t$. For a measurable event $B\subseteq E$,
\begin{equation}\label{eq:training-pushforward}
\zeta_{t+1}(B)=\int_ZK_t(z,\Phi^{-1}B)\,\mu_t(dz).
\end{equation}
Although $\zeta_t$ is known from $\mu_t$, the integrand in \eqref{eq:training-pushforward} need not be constant within the fibers of $\Phi$. Two complete-state laws with the same current observable image may therefore produce different next observable images. A closed transition law on $E$ exists for arbitrary initial complete-state laws only under an additional kernel-factorization condition.

For a semantic observation $\Phi=\sigma_{c,D,\mathcal Q}$, Equation~\eqref{eq:training-pushforward} is the exact evolution of the selected cut's computational-role profile. The frozen-receiver routing construction in Section~\ref{sec:semantic-shadow-nonclosure} demonstrates that this semantic image need not be autonomous.

For deterministic gradient descent, the obstruction can already occur in a two-weight active FFN branch with equal present predictor and different future loss (Section~\ref{sec:ffn-predictive-example}). Optimizer memory adds a separate possible obstruction: two histories producing the same current parameters may cause different next actions through different velocity or moment state. The behavioral optimizer quotient above addresses action prediction from histories; it is not automatically the smallest quotient that predicts a selected task observable of the complete closed loop.

Equation~\eqref{eq:training-pushforward} makes the information problem explicit. An effective state for future observations must distinguish complete states whenever their transition laws produce different future observable behavior.

\section{Training changes effective receiver roles at fixed architecture}
\label{sec:training-induced-role}

The contextual nonconstancy of Part~I compares producer contexts while holding the receiver architectures fixed. The following construction realizes that change along a parameter update. Its attention receiver is the first PreNorm attention of Appendix~\ref{app:dimension-labelled-transformer}, with the local residual state explicitly retained at the cut.

\begin{theorem}[Training-induced contextual receiver separation in a PreNorm interface]
\label{thm:training-induced-receiver-separation}
Let $d\ge3$ and choose orthonormal $e_1,e_2\in\mathbf1^\perp\subset\R^d$. On $D=\R^2$, fix the two-token producer family
\begin{equation}\label{eq:learned-routing-producer}
P_s(u,v)=\bigl(u e_1+(1-s)v e_2,\;s v e_2\bigr),\qquad s\in\R.
\end{equation}
It is realized by one fixed, linear-in-input producer architecture with a single shared parameter $s$ and complementary coefficients $1-s,s$. Fix the local receiver $Q_{\rm loc}(h_1,h_2)=h_1$ and the residual-augmented attention receiver
\begin{equation}\label{eq:prenorm-comparison-receiver}
Q_{\rm att}(h_1,h_2)
=\Bigl(h_1,\tfrac12N_\varepsilon(h_1)+\tfrac12N_\varepsilon(h_2)\Bigr),
\end{equation}
where $N_\varepsilon$ is the first stabilized PreNorm LayerNorm with unit gain, zero shift, and $\varepsilon>0$. The attention feature uses a fully admitted two-source row, $W_Q=W_K=0$, $W_V=I_d$, and a dimension-compatible identity value readout. No downstream parameter changes.

Every $P_s$ is injective. At $s=0$, $Q_{\rm loc}P_s$ and $Q_{\rm att}P_s$ both have diagonal predecessor kernels. At $s=1$, $Q_{\rm loc}P_s$ identifies exactly inputs with equal $u$, while $Q_{\rm att}P_s$ remains injective. Thus the effective local/attention distinction classes change from one to two at the declared kernel grain while the architecture, the two receiver maps, and all predecessor distinctions of the producer remain fixed.

The change occurs in one exact gradient-descent step on the same represented graph. Add the fixed training readout $T(h_1,h_2)=\langle h_2,e_2\rangle$ on the datum $(u,v)=(0,1)$ and squared target loss $\ell_s=\tfrac12(T(P_s(0,1))-1)^2$. Update only the shared producer coordinate $s$ with Euclidean learning rate $\eta=1$. Then $s_0=0$ gives $s_1=1$.
\end{theorem}
\begin{proof}
If $P_s(u,v)=P_s(u',v')$, the $e_1$ coefficient of the first token gives $u=u'$. Summing the $e_2$ coefficients of the two tokens gives $v=v'$, independently of $s$; hence $P_s$ is injective. At $s=0$ the first token $u e_1+v e_2$ determines $(u,v)$, so the local receiver is injective and the attention receiver, which retains that token, is also injective.

At $s=1$ the first token is $u e_1$, so the local receiver identifies all points with the same $u$. The attention receiver additionally gives $\frac12N_\varepsilon(u e_1)+\frac12N_\varepsilon(v e_2)$. Recover $u$ from the first component and subtract its normalized contribution. Stabilized LayerNorm is injective on the centered line $\R e_2$ by Lemma~\ref{lem:layernorm-radial-injective}; the remainder therefore determines $v$. The attention composite is injective. The zero score maps give equal masked-softmax weights $1/2$ on the two allowed sources, so this is an exact PreNorm attention feature, not an unnormalized averaging schema.

Finally $T(P_s(0,1))=s$ and $\ell_s=\tfrac12(s-1)^2$. Its derivative in the sole trainable coordinate is $s-1$, so $s_1=s_0-\eta(s_0-1)=1$ for $s_0=0$, $\eta=1$. The complementary routing weights belong to one shared one-dimensional parameter manifold; the update is taken in that coordinate. The readout is a fixed branch of the same represented computation and contributes no change to the comparison receivers.
\end{proof}

For a complete optimizer state $z=(\theta,\omega)$, take a measurable finite probe $\sigma_c(z)$ of the represented cut, such as the distinction matrices of the two fixed receivers on a finite test domain. If $\mu_{t+1}=\mu_tK_t$, the semantic-profile law $\eta_{c,t}=(\sigma_c)_\#\mu_t$ obeys
\begin{equation}\label{eq:semantic-law-pushforward}
\eta_{c,t+1}(B)=\int_ZK_t\bigl(z,\sigma_c^{-1}(B)\bigr)\,\mu_t(dz).
\end{equation}
The equation follows from the implemented training kernel. It propagates the distribution of contextual receiver roles, while the complete state retains the upstream and downstream information that the selected profile may forget.

The theorem exhibits a change of \emph{effective receiver identity}, not a change in architecture or a loss of predecessor information. The structural receiver pair is fixed; learning changes the upstream presentation on which each receiver acts. For $s\ne1$, the local receiver remains injective, while its inverse sensitivity to the $v$ coordinate grows as $|1-s|^{-1}$ near $s=1$.

\begin{proposition}[Bounded-receiver accessibility along the learned producer family]
\label{prop:bounded-receiver-accessibility}
Fix $u$ and compare $v\in\{-1,1\}$. If a scalar local continuation $g:\R^d\to\R$ is $L$-Lipschitz in Euclidean norm and predicts the target $v$ from the first token of $P_s(u,v)$, then
\begin{equation}\label{eq:bounded-receiver-error}
\max_{v\in\{-1,1\}}\bigl|g(u e_1+(1-s)v e_2)-v\bigr|
\ge\max\{0,1-L|1-s|\}.
\end{equation}
\end{proposition}
\begin{proof}
The two target values differ by $2$, while their first-token states differ by $2(1-s)e_2$. Their scalar predictions therefore differ by at most $2L|1-s|$. The triangle inequality bounds the sum of the two absolute prediction errors below by $2-2L|1-s|$, proving the result.
\end{proof}

With the same loss and $0<\eta<1$, an ordinary gradient-descent trajectory from $s_0=0$ satisfies
\begin{equation}\label{eq:scalar-routing-learning-trajectory}
s_t=1-(1-\eta)^t,
\qquad
\operatorname{err}_{\max}(t)\ge\max\{0,1-L(1-\eta)^t\}.
\end{equation}
The local receiver remains exactly injective at every finite $t$, while the lower bound on its best bounded-sensitivity recovery tends to $1$. The one-step transition at $\eta=1$ proves an exact change of receiver distinction class; the ordinary subunit-step trajectory reveals a continuously developing loss of accessibility before that limiting class is reached. Section~\ref{sec:matrix-routing} derives the corresponding anisotropic dynamics for a vector of represented features.

Exact kernel equivalence and receiver-bounded usability record different aspects of the same learned presentation. Before $s=1$, the first token still mathematically contains $v$; recovering it with a bounded receiver becomes increasingly costly as $s$ approaches the separation context.

\section{Vector-valued organization and learned attention allocation}
\label{sec:vector-routing-learned-attention}

The scalar producer of Theorem~\ref{thm:training-induced-receiver-separation} moves one coordinate between two tokens. Its exact distinction relation changes when the local coefficient vanishes. With several coordinates, training can redistribute the accessibility of different directions at different rates, even while every intermediate representation remains injective. The same distinction between information and usable computation appears when the attention scores themselves learn.

\subsection{An exact matrix-valued learned producer}
\label{sec:matrix-routing}

Let $a,k\ge1$ and choose a residual dimension $d\ge a+k+1$. Fix orthogonal linear isometric embeddings
\[
J_u:\R^a\hookrightarrow\mathbf1^\perp\subset\R^d,
\qquad
J_v:\R^k\hookrightarrow\mathbf1^\perp\subset\R^d,
\qquad J_u^*J_v=0.
\]
The architecture has two residual tokens. For $T\in\R^{k\times k}$, define the producer
\begin{equation}\label{eq:matrix-routing-producer}
P_T(u,v)=\bigl(J_u u+J_v(I-T)v,\ J_vTv\bigr).
\end{equation}
The two token projections $I-T$ and $T$ share one trainable matrix $T$. Their sum is the identity, so the represented parameterization enforces the conservation of the total embedded input. Fix the local receiver $Q_{\mathrm{loc}}(h_1,h_2)=h_1$ and the same residual-augmented, uniform-score PreNorm receiver as in Equation~\eqref{eq:prenorm-comparison-receiver}:
\[
Q_{\mathrm{att}}(h_1,h_2)
=\bigl(h_1,\tfrac12N_\varepsilon(h_1)+\tfrac12N_\varepsilon(h_2)\bigr).
\]
Both receivers have fixed parameters throughout the experiment.

\begin{theorem}[Spectrally resolved learned accessibility in a fixed PreNorm interface]
\label{thm:matrix-routing-spectral-accessibility}
Let $v$ be centered with finite positive-definite covariance $C=\E[vv^\top]$. Train $T$ alone by full-batch Euclidean gradient descent on the fixed second-token readout $J_v^*h_2=Tv$ with target $v$ and loss
\begin{equation}\label{eq:matrix-routing-loss}
\mathcal L(T)=\tfrac12\E\lVert Tv-v\rVert^2
=\tfrac12\operatorname{tr}\bigl((T-I)C(T-I)^\top\bigr).
\end{equation}
Initialize $T_0=0$ and take $0<\eta<\lambda_{\max}(C)^{-1}$. Then:
\begin{enumerate}[label=(\roman*)]
\item $P_T$ is injective for every $T$, and the fixed attention receiver $Q_{\mathrm{att}}P_T$ is injective. The local receiver identifies exactly the pairs with equal $u$ and equal $(I-T)v$.
\item The exact training trajectory is
\begin{equation}\label{eq:matrix-routing-spectral-solution}
\nabla_T\mathcal L=(T-I)C,\qquad
T_{t+1}=T_t-\eta(T_t-I)C,
\qquad I-T_t=(I-\eta C)^t.
\end{equation}
Thus $Q_{\mathrm{loc}}P_{T_t}$ remains injective at every finite $t$, but its sensitivity to a $C$-eigenvector $w_i$ of eigenvalue $c_i>0$ is $(1-\eta c_i)^t$.
\item Fix $u$ and the two inputs $v=\pm w_i$, where $\lVert w_i\rVert=1$. If an $L$-Lipschitz scalar local continuation $g$ predicts $\pm1$ from the first token, then
\begin{equation}\label{eq:matrix-routing-Lipschitz-bound}
\max_{v=\pm w_i}\bigl|g(J_uu+J_v(I-T_t)v)-\langle w_i,v\rangle\bigr|
\ \ge\ \max\bigl\{0,1-L(1-\eta c_i)^t\bigr\}.
\end{equation}
As $t\to\infty$, $T_t\to I$, and the local receiver identifies every pair having the same $u$, whereas the fixed attention receiver still distinguishes all $(u,v)$.
\end{enumerate}
\end{theorem}
\begin{proof}
From the token sum, $h_1+h_2=J_uu+J_vv$. Orthogonal projection onto $J_u$ and $J_v$ recovers $u,v$, proving that $P_T$ is injective for every $T$. The local receiver gives $J_u u+J_v(I-T)v$, so orthogonality proves the stated kernel relation. For the attention receiver, $h_1$ is explicitly retained. Its second output then determines $N_\varepsilon(h_2)$ by subtracting $N_\varepsilon(h_1)/2$. Stabilized LayerNorm with unit gain and zero shift is injective on the centered residual subspace: there it equals $x/(\varepsilon+\lVert x\rVert^2/d)^{1/2}$, whose radial magnitude is strictly increasing. Hence the second token $h_2$ is recovered, and so are $u,v$ by the token sum.

Differentiating \eqref{eq:matrix-routing-loss} gives $(T-I)C$. Therefore $I-T_{t+1}=(I-T_t)(I-\eta C)$, and $T_0=0$ proves the exact formula in \eqref{eq:matrix-routing-spectral-solution}. Since $0<\eta c_i<1$ for every eigenvalue, $I-T_t$ is invertible for every finite $t$ and tends to zero. On the pair $\pm w_i$, the two local token values are separated by $2(1-\eta c_i)^tJ_vw_i$ in Euclidean norm. Their targets differ by $2$. Applying the Lipschitz condition and the triangle inequality to the two prediction errors proves \eqref{eq:matrix-routing-Lipschitz-bound}. The limiting kernel statements follow from $T_t\to I$ and the already proved attention injectivity.
\end{proof}

The covariance eigenvectors are the directions in which the learned producer reorganizes its two tokens. If $c_i>c_j$, then under the stated step-size regime the first token loses bounded-receiver accessibility in direction $w_i$ faster than in direction $w_j$. For $L>1$, the bound in \eqref{eq:matrix-routing-Lipschitz-bound} becomes positive after the exact threshold
\begin{equation}\label{eq:accessibility-timescale}
t>\frac{\log L}{-\log(1-\eta c_i)}.
\end{equation}
The rates are selected by the input second moment and the learning rule. Every finite training step retains all predecessor distinctions in both the producer and the local receiver. The changing quantity is the effort needed by a receiver of declared sensitivity to use them.

Data-dependent mode dynamics are known for deep linear
networks \citep{saxe2014exact}. For controlled linear
Transformers, \citet{mainali2026exact} derive exact SGD
dynamics and covariance-dependent stages in the emergence
of in-context learning. The present model asks a different
question: how the learned producer changes recoverability
for a receiver with a fixed Lipschitz budget. The
covariance eigenvalues set the attenuation rates in
Equation~\eqref{eq:matrix-routing-spectral-solution},
while the producer and local receiver remain injective at
every finite step.
Equation~\eqref{eq:matrix-routing-Lipschitz-bound}
gives the resulting receiver-error barrier, and
Theorem~\ref{thm:exact-learned-accessibility-law}
identifies the optimal two-point recovery error.

The first-token norm need not be held fixed in this construction. The attention receiver is the exact stabilized PreNorm receiver, and its injectivity follows from the normalization map itself, not from treating normalization as an identity. The producer $P_T$ is a declared upstream linear-in-input module feeding that interface; realizing the same shared $T$ parameterization inside an arbitrary stack of standard Transformer blocks is a separate architectural implementation question.

\subsection{An actual Q/K update changes which source is usable}
\label{sec:learned-qk-attention}

The preceding producer changes while attention weights remain uniform. We now hold the producer fixed and train a query coordinate in a genuine two-source PreNorm attention row. This isolates a different dynamical mechanism: the attention receiver itself changes the weights with which it combines source information.

\begin{proposition}[Exact learned Q/K routing and receiver attenuation]
\label{prop:learned-qk-attenuation}
Let $d\ge5$ and choose four mutually orthonormal centered vectors $e_1,e_2,f_1,f_2\in\mathbf1^\perp\subset\R^d$. On the four-point predecessor domain $(a,b)\in\{-1,1\}^2$, take two fixed source tokens
\begin{equation}\label{eq:qk-input-tokens}
h_1(a)=e_1+a f_1,\qquad h_2(b)=e_2+b f_2.
\end{equation}
Use stabilized PreNorm LayerNorm with unit gain, zero shift, and $\varepsilon>0$, a fully admitted two-source attention row at token $1$, and temperature $\tau>0$. Put $\kappa=(\varepsilon+2/d)^{-1/2}$. Choose the fixed key and value projections and the one-parameter query projection on the centered basis by
\[
W_K e_1=\kappa^{-1},\quad W_K e_2=-\kappa^{-1},\quad
W_K f_1=W_K f_2=0,
\qquad
W_Q(\alpha)e_1=\alpha\kappa^{-1},\quad W_Q(\alpha)f_1=0,
\]
\[
W_Ve_1=W_Ve_2=0,\qquad
W_Vf_1=\kappa^{-1}b_1,\quad W_Vf_2=\kappa^{-1}b_2,
\]
where $b_1,b_2$ are orthonormal vectors in the two-dimensional value carrier; all unspecified coefficients and biases vanish. Then the query and keys on the first row are $q_1=\alpha$, $k_1=1$, $k_2=-1$, and the attended branch feature is
\begin{equation}\label{eq:qk-value-response}
z_\alpha(a,b)=p(\alpha)a b_1+[1-p(\alpha)]b b_2,
\qquad p(\alpha)=\frac{1}{1+\exp(-2\alpha/\tau)}.
\end{equation}
Train only the coordinate $\alpha$ on the datum $(a,b)=(1,1)$ using the fixed branch readout $z_\alpha$ and squared target loss $\mathcal L(\alpha)=\tfrac12\lVert z_\alpha(1,1)-b_1\rVert^2$. Euclidean gradient descent with $\eta>0$ satisfies
\begin{equation}\label{eq:qk-attention-gd}
\mathcal L(\alpha)=[1-p(\alpha)]^2,
\qquad
\alpha_{t+1}=\alpha_t+\frac{4\eta}{\tau}p(\alpha_t)[1-p(\alpha_t)]^2.
\end{equation}
For every finite initialization, $\alpha_t$ strictly increases to $+\infty$ and $p(\alpha_t)\to1$. For a fixed $a$, the residual-augmented attention receiver $(h_1,z_\alpha)$ distinguishes $b=\pm1$ at every finite step. If an $L$-Lipschitz scalar continuation predicts $b$ from this receiver, its worst-case absolute error on those two inputs obeys
\begin{equation}\label{eq:qk-receiver-accessibility}
\operatorname{err}_{\max}(t)\ge\max\{0,1-L[1-p(\alpha_t)]\}.
\end{equation}
Thus training of the actual query--key routing coordinate continuously suppresses access to the second source, even while softmax retains a strictly positive weight for that source at every finite parameter state.
\end{proposition}
\begin{proof}
Both input tokens have zero mean and squared norm $2$, so normalization is exactly $N_\varepsilon(h_i)=\kappa h_i$ throughout the declared input domain. The chosen projections therefore yield the stated queries, keys, and source values, with normalized scores $\pm\alpha/\tau$. This proves \eqref{eq:qk-value-response}. Orthogonality of $b_1,b_2$ gives $\mathcal L=(1-p)^2$. Since $p'=2p(1-p)/\tau$, differentiation gives $\mathcal L'=-4p(1-p)^2/\tau$, establishing the update.

Every increment in \eqref{eq:qk-attention-gd} is positive. If the increasing sequence $\alpha_t$ had a finite limit, continuity would make its increments converge to a strictly positive constant, a contradiction. Thus $\alpha_t\to+\infty$ and $p(\alpha_t)\to1$. For fixed $a$ and the two signs of $b$, the first token is identical and the attention feature values differ by $2(1-p)b_2$. Their target labels differ by $2$, so the Lipschitz inequality and triangle inequality give \eqref{eq:qk-receiver-accessibility}. Since $0<p(\alpha_t)<1$ at every finite time, the two receiver outputs are distinct throughout training.
\end{proof}

Unlike the uniform-score learned-producer example, the attention weights here are learned through the Q/K and softmax derivative of Equation~\eqref{eq:training-attention-pullback}. The experiment rewards the first source and suppresses the second; it is a model of task-selected receiver allocation. The exact distinction relation does not change at any finite step, while bounded continuation access does. That separation between formal recoverability and practical receiver sensitivity is a second dynamical manifestation of DLE, realized inside an actual normalized attention row. The normalization-compatible four-point input domain is necessary here: varying raw token magnitudes arbitrarily would vary LayerNorm factors and invalidate the claimed fixed score law.

\subsection{Exact accessible error and the time law of learned attention}
\label{sec:exact-receiver-accessibility}

The preceding bounds are attained on their two-point comparison domains. This allows receiver accessibility to be followed as an exact observable of training rather than inferred solely from a lower bound. For a receiver representation $q:D\to V$, target $y:D\to\R$, and admissible sensitivity $L$, define
\begin{equation}\label{eq:exact-accessibility-observable}
\mathcal E_L(q;y,D)=\inf_{\operatorname{Lip}(g)\le L}\sup_{x\in D}|y(x)-g(q(x))|.
\end{equation}
The receiver class is all real-valued $L$-Lipschitz maps on the displayed Euclidean carrier. Changing that class changes the observable; it does not change the exact predecessor kernel.

\begin{theorem}[Exact two-point accessibility and asymptotic attention allocation]
\label{thm:exact-learned-accessibility-law}
For the declared two-point comparison domains and Euclidean receiver metrics in Theorem~\ref{thm:matrix-routing-spectral-accessibility} and Proposition~\ref{prop:learned-qk-attenuation}, the optimal errors are
\begin{equation}\label{eq:exact-two-point-accessibility}
\boxed{\begin{aligned}
\mathcal E_{L,i}^{\mathrm{producer}}(t)
  &=\max\{0,1-L(1-\eta c_i)^t\},\\
\mathcal E_L^{\mathrm{attention}}(t)
  &=\max\{0,1-L[1-p(\alpha_t)]\}.
\end{aligned}}
\end{equation}
For the attention update of Equation~\eqref{eq:qk-attention-gd}, every $\eta,\tau>0$ and every finite initial $\alpha_0$ satisfy
\begin{equation}\label{eq:qk-time-asymptotic}
\boxed{\quad
1-p(\alpha_t)\sim\frac{\tau}{4\sqrt{\eta t}},\qquad
\mathcal E_L^{\mathrm{attention}}(t)
=1-\frac{L\tau}{4\sqrt{\eta t}}+o(t^{-1/2})
\quad(t\to\infty).}
\end{equation}
For each producer eigendirection, the corresponding attenuation is exactly geometric. Attention learning in this model produces algebraic attenuation without imposing an asymptotic law on the training dynamics.
\end{theorem}
\begin{proof}
Consider any two receiver values $z_+,z_-$ at distance $2a$ whose target labels are $+1,-1$. A scalar $L$-Lipschitz continuation satisfies $|g(z_+)-g(z_-)|\le2La$, hence its maximum prediction error is at least $\max\{0,1-La\}$. Conversely, assign values $g(z_\pm)=\pm\min\{1,La\}$. The two assigned values satisfy the Lipschitz constraint, and a scalar $L$-Lipschitz extension to the entire Euclidean carrier exists, for example by the usual infimum extension formula. The optimal error is therefore exactly $\max\{0,1-La\}$. Apply this with $a=(1-\eta c_i)^t$ for the local matrix-routing receiver, and with $a=1-p(\alpha_t)$ for the residual-augmented attention receiver at fixed first source input. This proves \eqref{eq:exact-two-point-accessibility}.

For the attention asymptotic, write $Y_t=\exp(4\alpha_t/\tau)$. Proposition~\ref{prop:learned-qk-attenuation} establishes $\alpha_t\to\infty$. Since $p_t=\sqrt{Y_t}/(1+\sqrt{Y_t})$ and $1-p_t=(1+\sqrt{Y_t})^{-1}$, the exact update yields
\[
Y_{t+1}=Y_t\exp\!\left(
\frac{16\eta}{\tau^2}\frac{\sqrt{Y_t}}{(1+\sqrt{Y_t})^3}
\right).
\]
As $Y_t\to\infty$, the exponent is $(16\eta/\tau^2)Y_t^{-1}+o(Y_t^{-1})$, hence
\[
Y_{t+1}-Y_t\longrightarrow\frac{16\eta}{\tau^2}.
\]
Taking Ces\`aro averages gives $Y_t/t\to16\eta/\tau^2$. Consequently $1-p_t=(1+\sqrt{Y_t})^{-1}\sim\tau/(4\sqrt{\eta t})$. For fixed $L$, eventually $L(1-p_t)<1$, so substitution into the exact optimal-error formula gives the second expression in \eqref{eq:qk-time-asymptotic}.
\end{proof}

Training analyses of softmax attention identify phases,
head specialization, and data-dependent behavior
\citep{huang2024incontext,chen2024softmaxdynamics,
he2025incontext}. \citet{vasudeva2025attention} establish
implicit-bias results and rates for attention
sparsification in a classification setting. Our
PreNorm Q/K calculation instead measures the optimal
recovery error of a downstream $L$-Lipschitz receiver.
The two source values remain distinguishable at every
finite training step, but the learned allocation obeys
$1-p_t\sim\tau/(4\sqrt{\eta t})$ and makes the
optimal two-point error approach one at the corresponding
algebraic rate. The matrix producer yields a geometric
accessibility law under a different exact training
mechanism. Neither law requires an earlier change in
the exact predecessor-equivalence relation.

\subsection{Coupled learning of a producer and an attention receiver}
\label{sec:coupled-producer-qk}

The preceding examples changed the producer or the attention receiver separately. Their gradients interact when both are trained through one task. A constant-norm token construction keeps the exact PreNorm and score calculations tractable while allowing both parameter sets to move.

\begin{proposition}[Exact two-parameter producer--attention coadaptation]
\label{prop:coupled-producer-attention}
Let $d\ge7$ and take six mutually orthonormal centered directions $e_1,e_2,f_1,g_1,f_2,g_2\in\mathbf1^\perp\subset\R^d$. A two-token producer with one learned coordinate $s$ is
\begin{equation}\label{eq:coupled-producer-tokens}
\begin{aligned}
h_1(a;s)&=e_1+a(\cos s\,f_1+\sin s\,g_1),\\
h_2(b;s)&=e_2+b(\cos s\,f_2+\sin s\,g_2),
\end{aligned}\qquad (a,b)\in\{-1,1\}^2.
\end{equation}
Let the first-row PreNorm Q/K projections be those of Proposition~\ref{prop:learned-qk-attenuation}, with the fixed keys taking values $1,-1$, the query taking value $\alpha$, and every signal direction annihilated by the Q/K projections. Let the fixed value map send $f_i\mapsto\kappa^{-1}b_i$ and $g_i\mapsto\kappa^{-1}c_i$, where $b_1,c_1,b_2,c_2$ are orthonormal value directions and $\kappa=(\varepsilon+2/d)^{-1/2}$. All position directions are annihilated by the value map. Thus, with $p=\sigma(2\alpha/\tau)$, the branch feature is
\begin{equation}\label{eq:coupled-receiver-feature}
z_{s,\alpha}(a,b)=p\,a(\cos s\,b_1+\sin s\,c_1)
 +(1-p)\,b(\cos s\,b_2+\sin s\,c_2).
\end{equation}
Train $s$ and $\alpha$ jointly on the datum $(a,b)=(1,1)$ with target $b_1$ and squared loss. Then the exact loss and simultaneous Euclidean gradient-descent update are
\begin{equation}\label{eq:coupled-producer-qk-dynamics}
\boxed{\begin{aligned}
\mathcal L(s,\alpha)&=1-p+p^2-p\cos s,\\
s_{t+1}&=s_t-\eta_s p_t\sin s_t,\\
\alpha_{t+1}&=\alpha_t-\frac{2\eta_\alpha}{\tau}p_t(1-p_t)
(2p_t-1-\cos s_t).
\end{aligned}}
\end{equation}
The producer's update speed depends on the attention weight, while the direction and magnitude of the query update depend on the learned producer angle. On the invariant slice $s=0$, the query update reduces exactly to Equation~\eqref{eq:qk-attention-gd}. For $0<s<\pi$, producer gradient descent moves $s$ toward $0$ for sufficiently small positive $\eta_s$. For fixed $s$, query gradient descent increases $\alpha$ precisely when $p<(1+\cos s)/2$.
\end{proposition}
\begin{proof}
Every token in Equation~\eqref{eq:coupled-producer-tokens} has zero mean and squared norm $2$, independent of $s,a,b$. Stabilized LayerNorm is therefore exactly multiplication by $\kappa$ throughout this parameterized predecessor domain. The fixed Q/K maps produce scores $\pm\alpha/\tau$, and the fixed value projection produces Equation~\eqref{eq:coupled-receiver-feature}. The four value directions are orthonormal, so the loss is
\[
\tfrac12\bigl(p^2+(1-p)^2+1-2p\cos s\bigr)
=1-p+p^2-p\cos s.
\]
Its derivatives are $\partial_s\mathcal L=p\sin s$ and
$\partial_\alpha\mathcal L=(2p-1-\cos s)(2/\tau)p(1-p)$.
The displayed update follows. If $s=0$, the query derivative is
$-(4/\tau)p(1-p)^2$. The other assertions follow from the signs of the two derivatives; taking $\eta_s p\le1$ keeps $s\in[0,\pi]$ from crossing below zero.
\end{proof}

This model exhibits a two-sided feedback mechanism without replacing PreNorm by an unnormalized surrogate. The producer changes the direction of the values presented to attention. The attention allocation then weights the training signal acting back on the producer. Simultaneously, the producer angle changes the query gradient, including the threshold at which attention switches from strengthening to reducing its preference for the first source. Its exact state is the coupled pair $(s,\alpha)$; the separate producer and Q/K laws are controlled slices rather than a universal reduction of their interaction.

\subsection{Effective dynamics of coupled producer and attention learning}
\label{sec:coupled-effective-dynamics}

The interaction in Proposition~\ref{prop:coupled-producer-attention} has an exact reduced description. Write $c=\cos s$ and $q=1-p$. Its loss becomes
\begin{equation}\label{eq:coupled-loss-decomposition}
\mathcal L(c,p)=q^2+p(1-c).
\end{equation}
The first term is error from allocating attention to the other source; the second is the mismatch between the first-source value direction and the target, weighted by its attention allocation. For a fixed producer orientation the loss-minimizing allocation is
\begin{equation}\label{eq:coupled-optimal-allocation}
p_*(c)=\tfrac12(1+c).
\end{equation}
This preferred allocation moves as the producer learns. The movement is not imposed as a training schedule: it is obtained by minimizing the declared squared loss at each producer orientation.

\begin{proposition}[Exact reduced state of discrete coupled gradient descent]
\label{prop:coupled-exact-reduced-state}
For any finite $s,\alpha$, positive temperature $\tau$, and step sizes $\eta_s,\eta_\alpha$, the update in Equation~\eqref{eq:coupled-producer-qk-dynamics} descends to the autonomous state $(c,p)=(\cos s,\sigma(2\alpha/\tau))$. Set $\delta=\eta_s p\sqrt{1-c^2}$. Then
\begin{equation}\label{eq:coupled-reduced-discrete}
\boxed{\begin{aligned}
c'&=c\cos\delta+\sqrt{1-c^2}\sin\delta,\\
p'&=\sigma\!\left(\operatorname{logit}p-
\frac{4\eta_\alpha}{\tau^2}p(1-p)(2p-1-c)\right).
\end{aligned}}
\end{equation}
The reduction identifies angles with the same cosine and retains the exact future $(c_t,p_t)$ trajectory for every such initial pair.
\end{proposition}
\begin{proof}
Since $s'=s-\eta_s p\sin s$, the cosine difference formula gives
$\cos s'=c\cos(\eta_s p\sin s)+\sin s\sin(\eta_s p\sin s)$.
The first trigonometric factor is even in $\sin s$, and the second product is also even. Replacing $|\sin s|$ by $\sqrt{1-c^2}$ gives the first displayed update even for negative $\sin s$. The query update gives
$2\alpha'/\tau=\operatorname{logit}p-(4\eta_\alpha/\tau^2)p(1-p)(2p-1-c)$.
Applying the sigmoid yields the second update. Both depend only on $(c,p)$.
\end{proof}

The continuous-time gradient-flow companion resolves the long-time behavior analytically. It is a separate, specified training law, obtained from $\dot s=-\gamma_s\partial_s\mathcal L$ and $\dot\alpha=-\gamma_\alpha\partial_\alpha\mathcal L$, rather than a substitution for the exact discrete experiment.

\begin{theorem}[Alignment, allocation and the late-time coupled regime]
\label{thm:coupled-gradient-flow-asymptotic}
For $\gamma_s,\gamma_\alpha,\tau>0$, let $-1<c_0<1$ and $0<p_0<1$. The coupled gradient flow induces
\begin{equation}\label{eq:coupled-gradient-flow-cp}
\boxed{\begin{aligned}
\dot c&=\gamma_s p(1-c^2),\\
\dot p&=\frac{4\gamma_\alpha}{\tau^2}
 p^2(1-p)^2(1+c-2p).
\end{aligned}}
\end{equation}
In the open domain
\((-1,1)\times(0,1)\),
the attention-allocation nullcline is
\[
p=\frac{1+c}{2}.
\]
There is no interior producer-alignment
nullcline, since
\(\dot c=\gamma_s p(1-c^2)>0\).
The producer-alignment nullclines
\(c=\pm1\), together with \(p=0\),
belong to the boundary of the
compactified \((c,p)\) domain. Every interior solution converges to $(c,p)=(1,1)$. Moreover $1-c(t)$ decays exponentially and
\begin{equation}\label{eq:coupled-flow-asymptotic}
\boxed{\quad
1-p(t)\sim\frac{\tau}{4\sqrt{\gamma_\alpha t}},
\qquad
\mathcal L(t)\sim\frac{\tau^2}{16\gamma_\alpha t}
\quad(t\to\infty).}
\end{equation}
The moving attention nullcline is not a curve of joint equilibria: $\dot c>0$ at every interior point.
\end{theorem}
\begin{proof}
The chain rule, $\partial_s\mathcal L=p\sin s$, $\partial_\alpha\mathcal L=(2/\tau)p(1-p)(2p-1-c)$, and $dp/d\alpha=(2/\tau)p(1-p)$ yield Equation~\eqref{eq:coupled-gradient-flow-cp}. The gradient-flow dissipation is
\[
\frac{d\mathcal L}{dt}
=-\gamma_s p^2\sin^2s
-\gamma_\alpha\left[\frac{2}{\tau}p(1-p)(2p-1-c)\right]^2\le0.
\]
Since $p>0$ in the interior, $c$ increases. Put $r(c)=(1+c)/2$ and $m=\min\{p_0,r(c_0)\}>0$. At $p=m\le r(c(t))$, the $p$ derivative is nonnegative. By scalar comparison, $p(t)\ge m$ at all times. Hence
$\dot c\ge\gamma_s m(1-c^2)$; equivalently, $\operatorname{artanh}c(t)\ge \operatorname{artanh}c_0+\gamma_s mt$.
Thus $c(t)\to1$ and $1-c(t)=O(e^{-2\gamma_s mt})$.

The curve $p=r(c)$ is moving upward. At a crossing, $\dot p=0$ but $\dot r=\dot c/2>0$, so $p-r$ can cross only from positive to negative. If a trajectory ever lies below $r(c)$, it remains there and $p$ is subsequently nondecreasing; its limit cannot be less than one because the right-hand side of its equation is strictly positive at $(c,p)=(1,p_\infty)$ whenever $0<p_\infty<1$. If it never lies below the curve, then $p(t)\ge r(c(t))\to1$. Thus in either case $p\to1$.

Let $q=1-p$ and $a=4\gamma_\alpha/\tau^2$. Its exact evolution is
$\dot q=-a p^2q^2[2q-(1-c)]\ge -2a q^3$.
Consequently $(q^{-2})'\le4a$, so $q(t)\ge(q_0^{-2}+4at)^{-1/2}$. Combining this algebraic lower bound with exponential decay of $1-c$ gives $(1-c)/q\to0$. Now
\[
\frac{d}{dt}q^{-2}=2a p^2\left(2-\frac{1-c}{q}\right)\longrightarrow4a.
\]
Integrating and taking Ces\`aro averages yields $q^{-2}/t\to4a$, that is $q\sim (4at)^{-1/2}=\tau/(4\sqrt{\gamma_\alpha t})$. Finally
$\mathcal L=q^2+p(1-c)$ and $1-c$ decays exponentially, proving the second asymptotic.
\end{proof}

\begin{figure}[H]
\centering
\includegraphics[width=.85\linewidth]{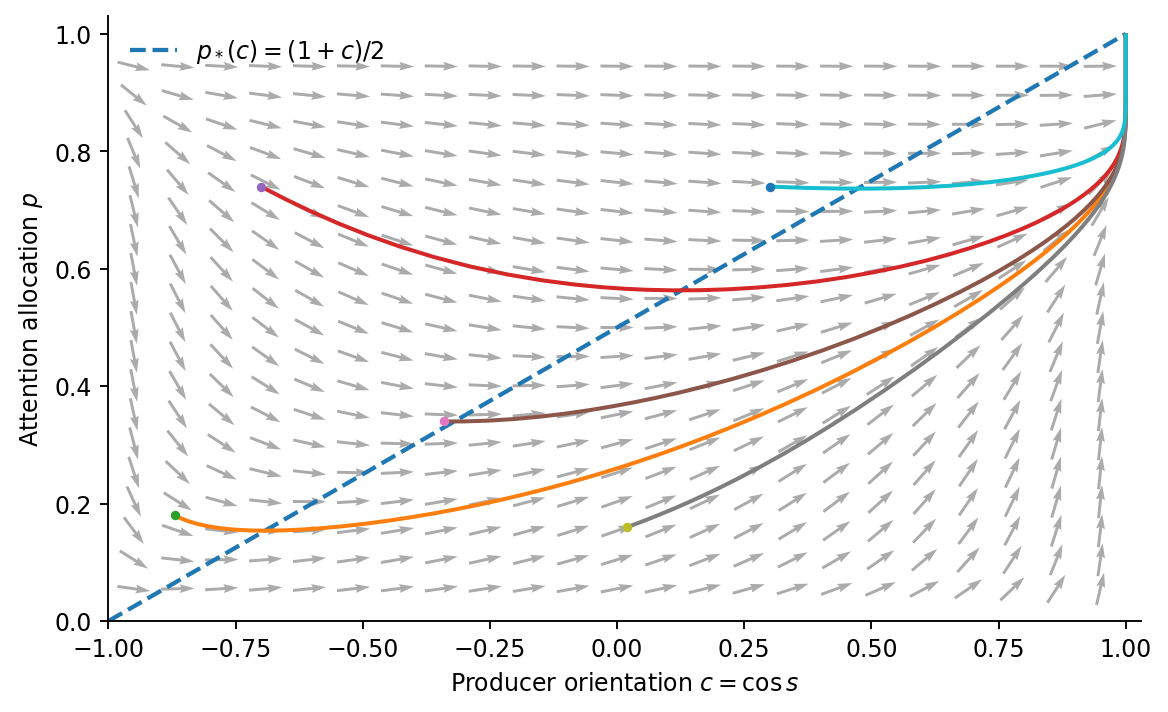}
\caption{The derived gradient-flow field for the coupled producer orientation $c=\cos s$ and attention allocation $p$. The dashed curve is the moving allocation optimum $p_*(c)=(1+c)/2$, not a set of joint equilibria. Trajectories approach aligned producer features and saturated source allocation at different rates. Parameters used for illustration: $\gamma_s=\gamma_\alpha=\tau=1$.}
\label{fig:coupled-phase-portrait}
\end{figure}

In Figure~\ref{fig:coupled-phase-portrait}, trajectories move toward producer alignment while attention follows a preferred allocation $p_*(c)=(1+c)/2$ that itself shifts as $c$ increases. The curve $p=p_*(c)$ is an attention nullcline, not a family of joint fixed points. Producer alignment becomes exponentially accurate once attention retains a positive share of that source; the softmax parameter continues toward saturation with an algebraic tail. Proposition~\ref{prop:coupled-exact-reduced-state} gives the exact discrete-time reduction, while Theorem~\ref{thm:coupled-gradient-flow-asymptotic} gives the stated continuous-flow asymptotics.

\subsection{Producer-dependent queries: representation changes attention scores}
\label{sec:content-dependent-qk}

In Proposition~\ref{prop:coupled-producer-attention}, query and key projections ignore the producer's rotating signal directions. The attention weight is therefore independent of $s$ at fixed $\alpha$, even though their training gradients interact. A different shared query projection gives a second mechanism: upstream presentation directly changes the scores computed by the unchanged attention receiver.

\begin{theorem}[Exact input-conditioned PreNorm routing under joint learning]
\label{thm:content-dependent-qk}
Use the two-token producer and six centered directions of Equation~\eqref{eq:coupled-producer-tokens}, with $d\ge7$ and $\kappa=(\varepsilon+2/d)^{-1/2}$. Retain the fixed shared key and value projections of Proposition~\ref{prop:coupled-producer-attention}. For a fixed $\beta\ne0$, modify the \emph{shared} query projection by
\[
W_Q(\alpha)e_1=\alpha\kappa^{-1},\qquad
W_Q(\alpha)f_1=\beta\kappa^{-1},\qquad
W_Q(\alpha)g_1=0,
\]
with zero coefficients on the other basis directions; the keys remain $W_K e_1=\kappa^{-1}$ and $W_K e_2=-\kappa^{-1}$, annihilating signal directions. All these maps act on the exactly normalized tokens. For each $(a,b)\in\{-1,1\}^2$, the first attention row has weight
\begin{equation}\label{eq:content-conditioned-weights}
\boxed{p_a(s,\alpha)=\sigma\!\left(\frac{2(\alpha+\beta a\cos s)}{\tau}\right),}
\end{equation}
and value feature
\begin{equation}\label{eq:content-conditioned-feature}
z_{s,\alpha}(a,b)=
p_a\,a(\cos s\,b_1+\sin s\,c_1)
+(1-p_a)\,b(\cos s\,b_2+\sin s\,c_2).
\end{equation}
For fixed $a$, the optimal error of an $L$-Lipschitz scalar continuation recovering $b=\pm1$ from the residual-augmented attention receiver is
\begin{equation}\label{eq:conditional-accessibility-error}
\boxed{\mathcal E_L^{(b\mid a)}(s,\alpha)=\max\{0,1-L(1-p_a(s,\alpha))\}.}
\end{equation}
Train the producer angle $s$ and query parameter $\alpha$ on $(a,b)=(1,1)$ with branch target $b_1$ and squared loss. Write $c=\cos s$, $p=p_+(s,\alpha)$, and
\begin{equation}\label{eq:content-A}
A(c,p)=\frac{2}{\tau}p(1-p)(2p-1-c).
\end{equation}
Then the exact loss and simultaneous Euclidean gradient-descent updates are
\begin{equation}\label{eq:content-joint-gd}
\boxed{\begin{aligned}
\mathcal L(s,\alpha)&=1-p+p^2-p\cos s,\\
s'&=s-\eta_s\sin s\,[p-\beta A(c,p)],\\
\alpha'&=\alpha-\eta_\alpha A(c,p).
\end{aligned}}
\end{equation}
At fixed $\alpha$, changing the producer changes the attention scores; at fixed parameter values, the two input regimes $a=\pm1$ generally give different attention weights and bounded-receiver access. Setting $\beta=0$ recovers the earlier coupled model.
\end{theorem}
\begin{proof}
The six selected centered directions are orthonormal. Each token has squared norm $2$ for every input and every angle, so stabilized PreNorm acts as multiplication by $\kappa$. The displayed \emph{linear}, shared projections thus give first-row query $q_1=\alpha+\beta a\cos s$, keys $k_1=1,k_2=-1$, and normalized source values $a(\cos s\,b_1+\sin s\,c_1)$ and $b(\cos s\,b_2+\sin s\,c_2)$. The two-source softmax yields Equation~\eqref{eq:content-conditioned-weights} and the weighted sum yields Equation~\eqref{eq:content-conditioned-feature}. For fixed $a$, changing $b$ leaves the first residual token, scores, and first-source value unchanged; the receiver outputs for $b=\pm1$ are separated by $2(1-p_a)$. The exact two-point extension argument of Theorem~\ref{thm:exact-learned-accessibility-law} proves Equation~\eqref{eq:conditional-accessibility-error}.

On the training datum the value directions are orthonormal, hence the stated loss. Differentiation gives
$\partial_\alpha p=(2/\tau)p(1-p)$ and
$\partial_s p=-(2\beta/\tau)p(1-p)\sin s$.
Since $\partial_p\mathcal L=2p-1-\cos s$, the chain rule gives
$\partial_\alpha\mathcal L=A(c,p)$ and
$\partial_s\mathcal L=p\sin s-\beta A(c,p)\sin s$.
The updates follow exactly. The claim about regimes follows from $p_+\ne p_-$ whenever $\beta\cos s\ne0$, because the sigmoid is strictly increasing.
\end{proof}

In this model the same Q/K receiver assigns different weights to the second source depending on what the learned producer places in the first token. This is a change of \emph{quantitative accessibility} at a fixed cut, not by itself a change of the exact predecessor kernel or of the complete represented semantics. The two input-conditioned access measurements will also determine an exact predictive state for their future evolution.

\begin{figure}[H]
\centering
\includegraphics[width=.85\linewidth]{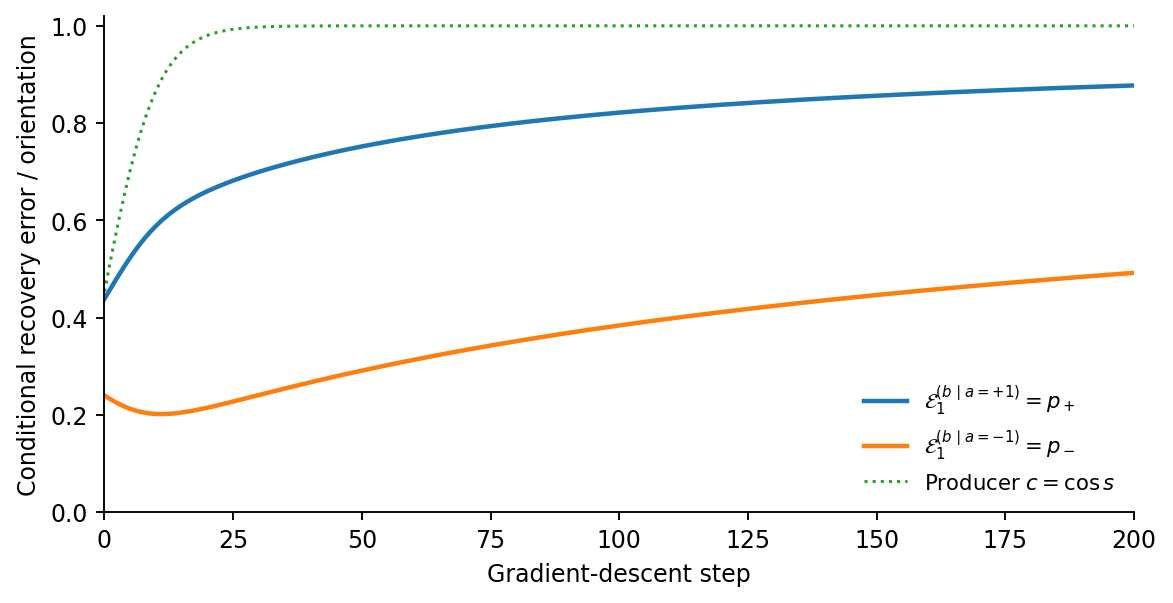}
\caption{Illustrative exact discrete learning trajectory of the producer-dependent Q/K model, showing $p_+$ and $p_-$, which equal the optimal unit-Lipschitz recovery errors for the second source on the two input regimes. The same learned attention parameters produce different receiver accessibility because the upstream representations differ.}
\label{fig:content-accessibility}
\end{figure}

The learned producer and learned Q/K examples isolate two ways to change the computations accessible at a cut. The covariance spectrum governs the geometric attenuation of directions redistributed between tokens, whereas the softmax gradient creates an algebraic late-time attenuation of one source. In the first coupled model, the producer affects the query gradient through target alignment while the scores are independent of producer orientation at fixed query parameters. In the content-dependent model, changing the producer modifies those scores directly. Both change bounded continuation accessibility; neither numerical change alone implies a new exact predecessor-kernel class. The reported error is the optimum over the declared $L$-Lipschitz receiver family, not necessarily the error attained by a particular trained downstream module.

\section{Input regimes and the evolution of semantic profiles}
\subsection{Input-regime dependence in one fixed represented computation}
\label{sec:input-regime-semantics}

The same change of receiver role can be induced by an input condition rather than a parameter update. Fix a binary routing input $b\in\{0,1\}$, restrict $|v|\le1$, and let $M>1$. A fixed ReLU gate computes
\begin{equation}\label{eq:binary-relu-route}
g(b,v)=\operatorname{ReLU}(v-M(1-b))-\operatorname{ReLU}(-v-M(1-b)).
\end{equation}
For $b=0$ this equals $0$, and for $b=1$ it equals $v$. The fixed producer
\begin{equation}\label{eq:input-conditional-producer}
P(u,v,b)=\bigl(u e_1+(v-g(b,v))e_2,\;g(b,v)e_2\bigr)
\end{equation}
therefore restricts on the two nontrivial input regimes
$D_b=\{(u,v,b):u\in\R,\ |v|\le1\}$
to the corresponding $P_0$ and $P_1$. Under these restrictions the same fixed local and PreNorm attention receivers have coincident predecessor distinction kernels on $D_0$ and distinct kernels on $D_1$. The network parameters and the represented operation graph do not change with $b$; the producer presentation does. Comparing two single activation values would not establish this result: each restriction contains a family of predecessor inputs whose receiver-induced identifications can be compared.

\subsection{Input-conditioned access in a learned attention receiver}
\label{sec:input-conditioned-access}

A fixed ReLU producer gives the exact-kernel contrast above. A second mechanism appears in the producer-dependent Q/K construction of Theorem~\ref{thm:content-dependent-qk}: with fixed network parameters, its two input regimes $a=+1$ and $a=-1$ have source weights
\[
p_\pm(s,\alpha)=\sigma\!\left(\frac{2(\alpha\pm\beta\cos s)}{\tau}\right).
\]
They are unequal whenever $\beta\cos s\ne0$. On either restricted regime the predecessor family still contains both values $b=\pm1$. At unit Lipschitz sensitivity, the exact optimal error for recovering $b$ is $\mathcal E_1^{(b\mid a)}=p_a$. The same fixed query/key/value projection therefore exhibits distinct quantitative computational accessibility on the two input conditions. Here the difference concerns \emph{bounded-receiver error}; the earlier ReLU construction gives the separate, stronger claim of different exact predecessor identifications.

\subsection{A current receiver-kernel profile
need not predict its next value}
\label{sec:semantic-shadow-nonclosure}

Define the coarse semantic observable $\chi(s)$ to be the pair of predecessor kernel relations of $Q_{\rm loc}P_s$ and $Q_{\rm att}P_s$ on $D$. At $s=0$ and $s=2$, both receivers are injective and $\chi(0)=\chi(2)$; at $s=1$ the local receiver identifies distinct $v$ values and $\chi(1)\ne\chi(2)$.

Use the same training readout $T(P_s(0,1))=s$, but set its squared-loss target to $2$ and the gradient-descent step to $\eta=\tfrac12$. The exact update is $s'=(s+2)/2$. Hence
\begin{equation}\label{eq:semantic-nonclosure-example}
\chi(0)=\chi(2),\qquad
\chi\bigl((0+2)/2\bigr)=\chi(1)\ne\chi(2)=\chi\bigl((2+2)/2\bigr).
\end{equation}
The current receiver-kernel profile
does not determine its next value
under the fixed training rule.
Although
\(\chi(0)=\chi(2)\), the two
parameter states have different
successor profiles.

The complete represented producers
\(P_0\) and \(P_2\) are different.
The result establishes nonclosure
of the selected kernel-profile
observable, not equivalence of the
full represented computations.
Section~\ref{sec:derived-predictive}
determines the additional
information required for exact
prediction under the training kernel.

\section{Factorized FFN learning:
identical functions, different futures}
\label{sec:ffn-predictive-example}
An active FFN branch gives a case where the present predictor identifies two parameter realizations that gradient descent subsequently separates. The branch is a two-parameter specialization of the fully resolved Transformer, and the update is computed exactly.

\begin{example}[Two-weight active FFN path]\label{ex:ffn-path}
Use one token, a residual width $d\ge2$, and an FFN with one active hidden unit. Fix the attention output to zero and all parameters except two FFN weights. Choose a nondegenerate residual input and fixed LayerNorm gain/shift so that the input to this hidden unit is $1$; choose its scalar activation to be ReLU on its positive branch. The incoming and outgoing FFN weights are $v$ and $u$ respectively. With a fixed linear output readout subtracting the residual baseline, the predictor on the single retained example is
\[f_{u,v}=uv\quad (v>0).\]
More strongly, on an arbitrary input whose fixed incoming preactivation is $a(x)$, the unit contributes $u\operatorname{ReLU}(v a(x))=uv\operatorname{ReLU}(a(x))$ for $v>0$ (with this unit's bias fixed at zero). Therefore parameter pairs having the same positive product $uv$ realize the same predictor function, not just the same loss on the retained example. This is a two-parameter slice of the fully resolved FFN in ABF Appendix~A, not a replacement architecture. Take squared target loss $R(u,v)=\frac12(uv)^2$ and exact full-batch gradient descent with step size $\eta>0$. Gradient descent is one permitted specialization of Appendix D's arbitrary typed causal optimizer.
\end{example}

The actual optimizer update is
\begin{equation}\label{eq:ffn-gd}
u'=u-\eta uv^2,\qquad v'=v-\eta u^2v.
\end{equation}
The two states $(u,v)=(1,1)$ and $(2,1/2)$ both realize $f=1$ and have risk $1/2$. At $\eta=0.1$ their next products are, respectively, $0.81$ and $0.585$, so their next risks are $0.32805$ and $0.1711125$. Thus \emph{even identical predictors need not determine identical next losses}: the optimizer distinguishes their parameter realizations. In particular, equality of function or current loss is not a strong Markov lumping for this training rule.

\begin{theorem}[Exact two-statistic closure on the active FFN slice]\label{thm:ffn-two-stat}
On any portion of the slice where the ReLU remains active, put
\[r=(uv)^2,\qquad q=u^2+v^2.\]
The squared-risk response and the exact gradient-descent update satisfy the closed polynomial system
\begin{equation}\label{eq:ffn-two-stat}
\boxed{\begin{aligned}
r'&=r\bigl(1-\eta q+\eta^2r\bigr)^2,\\
q'&=q\bigl(1+\eta^2r\bigr)-4\eta r,\\
R&=\tfrac12r.
\end{aligned}}
\end{equation}
Hence the joint future loss sequence is exactly determined by $(r,q)$ as long as the trajectory remains in the active FFN slice. On a subset with $r>0$ and $1-\eta q+\eta^2r>0$, $q$ is recovered from the present and next risks:
\begin{equation}\label{eq:ffn-q-recover}
q=\frac{1+\eta^2r-\sqrt{r'/r}}{\eta}.
\end{equation}
Thus, on such a region, the information encoded by $q$ cannot be discarded while retaining exact next-risk prediction for all starting states. On the zero-risk fiber $r=0$, all gradient-descent updates vanish and the value of $q$ is irrelevant to every future risk.
\end{theorem}
\begin{proof}
Write $p=uv$. Equation~\eqref{eq:ffn-gd} gives
\[p'=(u-\eta pv)(v-\eta pu)=p(1-\eta(u^2+v^2)+\eta^2p^2),\]
and
\[u'^2+v'^2=(u^2+v^2)-4\eta p^2+\eta^2p^2(u^2+v^2).\]
Squaring the first identity proves the $r$ recurrence; the second gives the $q$ recurrence. Substituting $R=r/2$ proves closure. Under the stated positivity condition the nonnegative square root of $r'/r$ is the parenthesized factor, giving \eqref{eq:ffn-q-recover}. If $r=0$, both gradient components in \eqref{eq:ffn-gd} vanish.
\end{proof}

The additional statistic has a direct interpretation
in learning-dynamics theory. For the retained scalar
predictor $f(u,v)=uv$, its empirical neural tangent
kernel is
\[
K_{u,v}
=\|\nabla_{(u,v)}f\|^2
=u^2+v^2=q.
\]
The parameter dependence of tangent-kernel geometry
is central to the function-space description of
gradient learning \citep{jacot2018ntk}. Here it is
visible in its smallest nontrivial factorized form:
the same current predictor $f$ can have different $q$
and therefore a different next predictor. Under
gradient flow the relation is particularly transparent,
\[
\dot f=-qf,\qquad \dot q=-4f^2.
\]
The discrete system in
Equation~\eqref{eq:ffn-two-stat} gives the exact
finite-step counterpart, with $r=f^2$ and
$\|\nabla R\|^2=rq$. Thus $(r,q)$ closes future loss
prediction on the active branch, while the present
predictor alone does not. This is the
parameterization-dependent learning behavior found
more generally in deep factorized networks
\citep{saxe2014exact}.

For momentum or Adam, the action depends additionally on their stored velocity or moments; clipping changes the update map itself. The predictive coordinates must then be derived from that optimizer's complete kernel. Equation~\eqref{eq:ffn-two-stat} is the closure for the specified Euclidean gradient-descent branch.

\subsection{An exact matrix-factorization closure on a fixed activation sector}
\label{sec:matrix-factorization-training}

The same realization dependence appears in a two-factor matrix branch. On a specified fixed activation sector the trainable map is $M=UV$; changes of activation pattern change that factorization and its derivatives. The following result isolates the fixed-sector dynamics.

\begin{proposition}[Gram-completed evolution of a two-factor branch]\label{prop:gram-completed-evolution}
Let $U\in\mathbb R^{m\times k}$ and $V\in\mathbb R^{k\times n}$, put $M=UV$, and suppose the loss has the form $\mathcal L(U,V)=\ell(M)$ with differentiable $\ell$. Define $G=\nabla_M\ell(M)$, $A=UU^\top$, and $B=V^\top V$. Under simultaneous Euclidean gradient descent with step $\eta$,
\begin{equation}\label{eq:matrix-gd-factor}
U'=U-\eta GV^\top,\qquad V'=V-\eta U^\top G,
\end{equation}
the three matrices $(M,A,B)$ obey the exact closed evolution
\begin{equation}\label{eq:matrix-gd-closure}
\boxed{\begin{aligned}
M'&=M-\eta(GB+AG)+\eta^2GM^\top G,\\
A'&=A-\eta(MG^\top+GM^\top)+\eta^2GBG^\top,\\
B'&=B-\eta(M^\top G+G^\top M)+\eta^2G^\top AG.
\end{aligned}}
\end{equation}
Consequently the realized matrix $M$ need not determine its next value under parameter-factor gradient descent, but the Gram-completed state $(M,A,B)$ does. The closure dimension is inherited from the declared matrix-factorized regime; no claim of minimality is made.
\end{proposition}
\begin{proof}
The product rule gives $\nabla_U\mathcal L=GV^\top$ and $\nabla_V\mathcal L=U^\top G$. Substitute \eqref{eq:matrix-gd-factor} into $M'=U'V'$, $A'=U'U'^\top$, and $B'=V'^\top V'$, expand, and use $UV=M$, $UU^\top=A$, and $V^\top V=B$. Every resulting term depends only on $(M,A,B)$ through $G=\nabla\ell(M)$.
\end{proof}

The factors affect function-space motion through the
Gram matrices $A=UU^\top$ and $B=V^\top V$.
In the update for $M=UV$, the terms $AG$ and $GB$
precondition the composite-map gradient on its two
sides. This is the mechanism by which factorization
alters optimization even when it does not enlarge the
represented linear-function class
\citep{arora2018acceleration}. Related
reparameterization analyses demonstrate that
parameter-space geometric descriptions need not be
invariant under transformations preserving the
realized function \citep{dinh2017sharp}.
Here the exact recurrence identifies the additional
state required for prediction: two realizations with
the same $M$ can have different next matrices, whereas
$(M,A,B)$ has closed evolution under the declared
gradient-descent rule. Adding gates, normalization,
or optimizer memory changes the update and requires
recomputing that closure.

\section{Predictive information generated by the actual optimizer}\label{sec:derived-predictive}
\subsection{Training kernel and observable}
The learned-routing construction in
Section~\ref{sec:semantic-shadow-nonclosure} gives
complete training states with the same present
contextual distinction profile and different future
profiles. The FFN example gives the corresponding
failure for predictor and risk. Both raise a
predictive-state question: which distinctions must
be retained from the complete training state to
determine the future of a selected observable?

Computational mechanics identifies histories by
the future observation distributions they induce
\citep{shalizi2001computational}. Predictive state
representations similarly organize a dynamical
system around predictions of future observable
tests \citep{littman2001predictive}. The training
experiment supplies a concrete transition kernel
on the complete parameter--optimizer state, while
the computational cut supplies an observable that
need not close under that transition. We use the
kernel to derive the information required for
future prediction. Depending on the selected
observable, the resulting state describes task
performance, receiver accessibility, or the
evolution of a contextual computational role.

Fix $(L,c_L)$ and a realized training dataset $S$. Write $Z=\Theta_{L,c_L}\times\Omega_{\mathbf O;L,c_L}$, $K_t=K^{S,\mathbf O}_{L,c_L,t}$, and let $\Phi:Z\to E$ be a selected measurable observable. For computational organization, choose the finite cut profile $\sigma_{c,D,\mathcal Q}$ of Equation~\eqref{eq:finite-semantic-probe}, or its distinction-equivalence component; for task performance, choose risk or predictor responses. These are the objects of Section~\ref{sec:training-closed-loop} and Appendix D, with $K_t$ the composite of the Transformer observation kernel $\Gamma_t^S$, the optimizer's action and memory kernels $U_{\mathbf O},M_{\mathbf O}$, and the implemented action map $\mathfrak R$. For finite response evaluations one may take $E=\mathbb R^{q+1}$; the general construction works for a declared measurable output space $E$. The kernels $K_t$ may depend on training time; each is obtained from the specified Transformer--optimizer experiment.

We work first with bounded measurable functions of the observable. Unbounded risk coordinates can be used whenever the corresponding conditional expectations are finite. Let $\mathcal B(Z)$ be the full-state sigma-algebra and $B_b(\mathcal H)$ denote the bounded real-valued functions measurable with respect to $\mathcal H\subseteq\mathcal B(Z)$.

\subsection{Backward closure of observable information}

\begin{definition}[Backward predictive closure]
\label{def:forced-hierarchy}
For each training time $t$ put
\begin{equation}\label{eq:forced-hierarchy}
\mathcal H_t^{(0)}=\sigma(\Phi),\qquad
\mathcal H_t^{(n+1)}=\sigma\!\left(\mathcal H_t^{(n)},\ \{K_t f:f\in B_b(\mathcal H_{t+1}^{(n)})\}\right),
\qquad
\mathcal H_t^{(\infty)}=\sigma\left(\bigcup_{n\ge0}\mathcal H_t^{(n)}\right).
\end{equation}
Here $(K_tf)(z)=\int f(z')K_t(z,dz')$. Each step adds the distinctions needed to predict an additional finite horizon of retained observations.
\end{definition}

\begin{theorem}[Minimal dynamically stable information]\label{thm:forced-minimal}
For every $t$, the sigma-algebra $\mathcal H_t^{(\infty)}$ contains $\sigma(\Phi)$ and satisfies
\begin{equation}\label{eq:forced-stable}
K_t B_b(\mathcal H_{t+1}^{(\infty)})\subseteq B_b(\mathcal H_t^{(\infty)}).
\end{equation}
If another family $\{\mathcal G_t\}_{t\ge0}$ contains $\sigma(\Phi)$ and obeys $K_tB_b(\mathcal G_{t+1})\subseteq B_b(\mathcal G_t)$ for all $t$, then $\mathcal H_t^{(\infty)}\subseteq\mathcal G_t$ for every $t$. Thus \eqref{eq:forced-hierarchy} is the smallest backward-kernel-stable information family containing the evaluation receiver.
\end{theorem}
\begin{proof}
The inclusions $\mathcal H_t^{(n)}\subseteq\mathcal H_t^{(n+1)}$ are immediate. If $f$ is measurable at some finite level $\mathcal H_{t+1}^{(n)}$, then $K_tf$ is measurable at level $\mathcal H_t^{(n+1)}$. The bounded functions $f$ for which $K_tf$ is measurable with respect to $\mathcal H_t^{(\infty)}$ form a vector space closed under bounded monotone pointwise limits. The monotone-class theorem extends this property from the union of the finite-level sigma-algebras to all of $B_b(\mathcal H_{t+1}^{(\infty)})$, proving \eqref{eq:forced-stable}. For minimality, induction gives $\mathcal H_t^{(n)}\subseteq\mathcal G_t$: the case $n=0$ is assumed, and backward stability supplies the induction step. Take the generated sigma-algebra.
\end{proof}

The hierarchy gives the smallest information family stable under the backward transition operators. If it is generated by a measurable quotient $q_t:Z\to Y_t$ and the induced transition probabilities form a measurable kernel, that quotient carries the retained observable dynamics. The existence and dimension of such a state space are properties of the given kernel.

\begin{corollary}[Exact marginal closure whenever the quotient is realized]\label{cor:realized-predictive}
Suppose $\mathcal H_t^{(\infty)}=\sigma(q_t)$ and the fiber-constant conditional transition probabilities admit a measurable induced kernel $\widehat K_t:Y_t\rightsquigarrow Y_{t+1}$. For every input measure $\mu_t$,
\[
\nu_t=(q_t)_\#\mu_t,\qquad
\nu_{t+1}=\nu_t\widehat K_t,\qquad
\zeta_t=(\widehat\Phi_t)_\#\nu_t,
\]
where $\Phi=\widehat\Phi_t\circ q_t$. This is exact at each training time and places no restriction on $\nu_t$'s subsequent behavior.
\end{corollary}
\begin{proof}
For any measurable $B\subseteq Y_{t+1}$, $K_t(z,q_{t+1}^{-1}(B))$ is $\sigma(q_t)$-measurable by Theorem~\ref{thm:forced-minimal}; by the stated realizability hypothesis it equals $\widehat K_t(q_t(z),B)$. Integrate against $\mu_t$ and apply pushforward.
\end{proof}

\subsection{The predictive quotient of a learned semantic trajectory}
\label{sec:routing-predictive-quotient}

The distinction profile in Section~\ref{sec:semantic-shadow-nonclosure} gives a direct application of the backward construction to a training-dependent computational role. Let $f(s)=(s+2)/2$ be the exact full-batch gradient-descent update and $\chi(s)$ the pair of local and attention predecessor-kernel relations on the two-token domain. The attention receiver is injective for every $s$; the local receiver is injective precisely when $s\ne1$. Thus $\chi$ has an exceptional value at $s=1$ and one common value at all other real parameter states.

\begin{proposition}[Countable exact predictive quotient of learned receiver semantics]
\label{prop:semantic-countdown-quotient}
Put $a_n=2-2^n$ for $n=0,1,2,\ldots$ and $E_*=\R\setminus\{a_n:n\ge0\}$. For the deterministic training kernel $K(s,\cdot)=\delta_{f(s)}$ and observable $\Phi=\chi$, the coarsest complete-state partition preserving the entire future semantic-profile path consists of
\begin{equation}\label{eq:semantic-countdown-partition}
\boxed{\ \{a_0\},\ \{a_1\},\ \{a_2\},\ldots,\ E_*\ }.
\end{equation}
It is also a strong Markov quotient. Label the exceptional classes by $n\in\N_0$ and the complementary class by $\infty$. Its transition is the deterministic countdown
\begin{equation}\label{eq:semantic-countdown-transition}
n\mapsto n-1\ (n\ge1),\qquad 0\mapsto\infty,\qquad
\infty\mapsto\infty.
\end{equation}
The current semantic profile has only two values, but exact prediction of its complete future requires countably many classes of initial parameters.
\end{proposition}
\begin{proof}
Iteration gives $f^t(s)=2+2^{-t}(s-2)$, and $f^t(s)=1$ if and only if $s=2-2^t=a_t$. For an initial condition $a_n$, $\chi(f^t(a_n))$ is exceptional at exactly $t=n$; for $s\in E_*$ it is never exceptional. The future profile sequences are therefore pairwise different for all $a_n$ and agree for every state in $E_*$. This proves that the displayed partition is the coarsest future-path-equivalence partition. The update sends $a_n$ to $a_{n-1}$ for $n\ge1$, sends $a_0=1$ to $3/2\in E_*$, and maps $E_*$ into $E_*$: any predecessor of $a_n$ is $a_{n+1}$, already outside $E_*$. Hence the transition law is constant on each class, giving the stated strong Markov quotient. The quotient map is Borel measurable, since its classes are countable singletons and their Borel complement.
\end{proof}

This is the predictive information required by the learned receiver-identity observable itself. Each singleton $a_n$ has the same current generic semantic profile unless $n=0$, yet differs by the number of training steps until the exceptional local/attention separation occurs. The time index is information about the present \emph{parameter state}, not an additional architectural operation. Under an initialization law without atoms at $\{a_n:n\ge0\}$, this countable exceptional set is visited with probability zero at every finite training time. The displayed partition is the globally exact predictive quotient on the declared real parameter state space; the predictive quotient of an initialization-restricted experiment can be smaller. By contrast, the bounded-accessibility profile varies along ordinary training trajectories. Under the alternative learned producer trajectory of Theorem~\ref{thm:training-induced-receiver-separation}, the relevant predictive partition changes because the optimizer law changes.

\subsection{A complete predictive state from two input regimes}
\label{sec:two-regime-accessibility-prediction}

The content-dependent attention model of Theorem~\ref{thm:content-dependent-qk} supplies a continuously changing semantic-accessibility observable, in contrast with the exceptional exact-kernel event of Proposition~\ref{prop:semantic-countdown-quotient}. For unit receiver sensitivity, define
\[
\Psi(s,\alpha)=(\mathcal E_1^{(b\mid+1)},\mathcal E_1^{(b\mid-1)})=(p_+,p_-).
\]
Both components are observable receiver-accessibility errors on nontrivial two-point input regimes. Their joint dynamics close without requiring the full angle $s$.

\begin{theorem}[Two-regime predictive closure and one-regime insufficiency]
\label{thm:two-regime-predictive-state}
Let $\beta\ne0$, $\tau>0$, $p_\pm\in(0,1)$ and write $\ell_\pm=\operatorname{logit}(p_\pm)$. The two-regime observable $\Psi=(p_+,p_-)$ uniquely determines the reduced state $(\alpha,c)=(\alpha,\cos s)$ by
\begin{equation}\label{eq:two-regime-inversion}
\boxed{\quad
\alpha=\frac\tau4(\ell_++\ell_-),\qquad
c=\frac\tau{4\beta}(\ell_+-\ell_-).
\quad}
\end{equation}
On the realizable range $|c|\le1$, the exact gradient-descent evolution of Theorem~\ref{thm:content-dependent-qk} descends to an autonomous map of $\Psi$. With $p=p_+$, $A=A(c,p)$, $H=p-\beta A$, and $\delta=\eta_sH\sqrt{1-c^2}$, its reduced update is
\begin{equation}\label{eq:content-reduced-dynamics}
\boxed{\quad
c'=c\cos\delta+\sqrt{1-c^2}\sin\delta,
\qquad \alpha'=\alpha-\eta_\alpha A(c,p),
\quad}
\end{equation}
followed by $p_\pm'=\sigma(2(\alpha'\pm\beta c')/\tau)$. Thus the present two-regime profile determines its entire future profile under the declared training law. One regime alone is not sufficient in general.
\end{theorem}
\begin{proof}
Taking logits in Equation~\eqref{eq:content-conditioned-weights} gives
$\ell_\pm=2(\alpha\pm\beta c)/\tau$. Their sum and difference prove Equation~\eqref{eq:two-regime-inversion}. The gradient update has $s'=s-\eta_sH\sin s$ with $H$ depending on $(c,p_+)$ alone. Applying the same even-in-$\sin s$ cosine identity used in Proposition~\ref{prop:coupled-exact-reduced-state} gives Equation~\eqref{eq:content-reduced-dynamics}. The update of $\alpha$ also depends only on $(c,p_+)$. Hence states with identical present $(p_+,p_-)$ have identical future profiles by induction.

To show that one regime is insufficient, take $\beta=1$, $\tau=2$, $\eta_s=\eta_\alpha=0.1$. The distinct states $(s,\alpha)=(0,-1)$ and $(s,\alpha)=(\pi/2,0)$ both have $p_+=1/2$, but direct substitution into the exact update gives respectively
\[
p_+'\simeq0.5062496745\qquad\text{and}\qquad p_+'\simeq0.5124921921.
\]
Their second-regime measurements are $p_-=(1+e^2)^{-1}$ and $p_-=1/2$, respectively. Thus $p_+$ alone does not descend to an autonomous update, while the two-regime profile does.
\end{proof}

The inverse in Equation~\eqref{eq:two-regime-inversion} reconstructs the producer orientation $c$ and query parameter $\alpha$, not the angle $s$ or the entire represented computation. Because the profile is one-to-one with the reduced state, it is an exact deterministic strong predictive state for this observable. The two measurements are an invertible observable coordinate chart for $(\alpha,\cos s)$ on the realizable range. Observing the receiver under distinct input contexts therefore reveals the reduced state needed to predict these same accessibility measurements. This establishes dynamical sufficiency for the selected profile, not a reconstruction of the entire represented computation or a general minimum-state theorem. A single context hides a distinction the optimizer uses on the next step.

\subsection{Different prediction targets}
The countable semantic countdown is an exact strong Markov quotient, but a quotient grouping complete states by equal future output-path laws need not itself be strongly lumpable. For a finite implemented training system the coarsest strong Markov quotient is obtained by refining the observable partition until transition probabilities into each block are constant. Appendix~\ref{app:predictive-reduction-details} gives the counterexample, exact partition algorithm, and proof. These are distinct prediction contracts: current observable equivalence, full observable-path equivalence, and Markov sufficiency of a reduced state.

\section{Marginal closure and symmetries of the training process}
\label{sec:predictive-extensions-main}
The full future path of contextual semantics is a stronger prediction target than its one-time marginal law. For a finite complete-state Markov kernel $K$, the minimal invariant linear space generated by indicators of present observable classes is
\[
\mathcal V=\operatorname{span}\{K^n\mathbf1_{\{\Phi\in B\}}:n\ge0,\ B\subseteq\Phi(Z)\}.
\]

This is an invariant-observable construction for
the training Markov operator. Koopman approaches
study dynamical systems through the evolution
of functions of their states, with finite-dimensional
invariant subspaces providing closed observable
descriptions \citep{brunton2016koopman}.
Here the starting functions are indicators of the
selected computational or task observable, and
the subspace is generated by their iterates under
the actual training kernel. The construction
therefore determines the observable coordinates
needed for exact marginal evolution instead of
assuming a closed reduced coordinate system
in advance.

Writing $Kb_j=\sum_i A_{ji}b_i$ in a basis gives the exact evolution $m_{t+1,j}=\sum_i A_{ji}m_{t,i}$ for $m_{t,i}=\mu_t(b_i)$. The induced coefficient matrix need not be stochastic, and this calculation preserves observable marginals rather than their joint paths. Theorem~\ref{thm:linear-closure} and its constructive algorithm appear in Appendix~\ref{app:predictive-reduction-details}.

A different reduction identifies parameter/optimizer states related by a symmetry of the \emph{actual training experiment}. A group action $g$ must preserve both the selected semantic observable and the learning kernel:
\[
\Phi(gz)=\Phi(z),\qquad K_t(gz,gA)=K_t(z,A).
\]
Then training cannot distinguish states in the same orbit through future values of $\Phi$. Attention-head relabelings or FFN-unit permutations yield this reduction when masks, observations, optimizer memory, and implemented updates respect the same relabeling. The exact criterion and proof are in Proposition~\ref{prop:symmetry-lumping}. Function-preserving transformations that change gradient-descent updates do not satisfy it.

\section{Same-function architectures
and new tangent directions}
\label{sec:refinement-innovations}

An identity block can be inserted into a residual Transformer without changing its realized function. Likewise, new hidden FFN units can be installed behind a zero output gate. These constructions place the same predictor inside two distinct represented architectures. Their available tangent directions and optimizer dynamics can then be compared from a common initial function.

\subsection{The identity fiber gives an exact same-function lift}
\label{sec:identity-fiber}

For an inserted attention branch, set every output projection and the attention output bias to zero. The branch update vanishes while its query, key, value, routing, and source-resolved internal states remain free. For an inserted feed-forward branch, set only the final output map and bias to zero. The branch output again vanishes while every hidden computation remains free. With both residual branches closed in this way, the inserted block is exactly the identity,
\[
B_{\mathrm{id}}(H)=H,
\qquad
D_HB_{\mathrm{id}}=I.
\]
The fine architecture can therefore contain new internal computation while realizing exactly the coarse predictor and preserving first-order state propagation through the inserted block.

Opening an output gate makes the latent computation transverse to the same-function fiber. For example,
\[
D_{W_{n+1}}F_{\mathrm{ff}}[\dot W_{n+1}]
=
\dot W_{n+1}z_n,
\]
and an attention output gate gives the analogous term \(\dot O_a z_a\). The latent states \(z_n\) and \(z_a\) are producer features for newly available predictor variations.

Let $c=(c_c,c_f)$ be a declared pair of admissible dimension-labelled coordinates for depths $L$ and $mL$, with $D_Lc:=c_c$ and $D_{mL}c:=c_f$. These are projections of the comparison datum, not depth-RG rescalings. For a type-compatible architecture comparison
\[
\mathcal T_{L,D_Lc}
\longrightarrow
\mathcal T_{mL,D_{mL}c},
\]
write
\[
J^{\Theta}_{m,L,c}:
\Theta_{L,D_Lc}
\longrightarrow
\Theta_{mL,D_{mL}c}
\]
for the resulting parameter lift. It preserves the realized predictor,
\[
r_{mL,D_{mL}c}\circ J^{\Theta}_{m,L,c}
=
r_{L,D_Lc}.
\]
This lift is defined only where the fine architecture admits the required type-correct representative of the coarse predictor. The closed-gate identities above construct this representative for identity-block insertion. Other width/refinement schemes require their own specified lift.

The optimizer part of the comparison is not defined by an optimizer-specific instruction such as ``copy the moments.'' Let
\[
\mathfrak J^{\mathrm{hist}}_{m,L,c}
:
\mathsf H_{L,D_Lc}
\longrightarrow
\mathsf H_{mL,D_{mL}c}
\]
be the declared history lift obtained by applying the parameter comparison on inherited coordinates and the neutral identity-fiber convention on inserted coordinates and observations. When this lift respects optimizer behavioral equivalence, Appendix~\ref{app:optimizer-state-lift-E} proves that it descends uniquely to
\[
J^{\Omega}_{m,L,c}:
\Omega_{\mathbf O;L,D_Lc}
\longrightarrow
\Omega_{\mathbf O;mL,D_{mL}c}.
\]
The complete comparison lift is therefore
\[
\boxed{
J_{m,L,c}
=
J^{\Theta}_{m,L,c}\times J^{\Omega}_{m,L,c}.
}
\]
The optimizer is free to respond differently after refinement. The lift supplies a common state; it does not impose commutation with training.

\subsection{Internal widening gives a nontrivial exact same-function family}
\label{sec:width-comparison-fiber}

The same construction is not restricted to inserting whole identity blocks. Internal hidden capacity can grow while the residual carrier and the realized predictor remain fixed. For a feed-forward map
\[
F_{\mathrm{ff}}(x)=W_2\phi(W_1x+b_1)+b_2
\]
with hidden width \(w\), choose \(w'>w\) and define
\[
W_1'
=
\begin{bmatrix}
W_1\\ A
\end{bmatrix},
\qquad
b_1'
=
\begin{bmatrix}
b_1\\ a
\end{bmatrix},
\qquad
W_2'
=
\begin{bmatrix}
W_2 & 0
\end{bmatrix},
\qquad
b_2'=b_2.
\]
The added hidden coordinates may compute arbitrary features, but the zero output columns make
\[
F_{\mathrm{ff}}'(x)=F_{\mathrm{ff}}(x)
\]
exactly. Opening the new output columns later exposes additional function-changing directions. Attention admits the same pattern: additional heads may be introduced with arbitrary internal query/key/value parameters and zero columns in the output projection. The residual carrier is unchanged, so these widenings preserve the boundary type required by the comparison map.

The zero-output-column construction widens the FFN with identical initial predictor. Changing the residual carrier width requires a separate boundary map, and comparing training additionally requires a lift of the optimizer's behavioral state.

\subsection{Fine structure changes available training directions}
At a common predictor, the inherited coarse tangent variations form a subspace of the fine-model variations. Their quotient
\[
N=V_f/I(V_c)
\]
records directions newly available through the finer represented architecture. Opening a previously zeroed FFN output gate exposes $\dot W z$; its downstream relevance is determined by the receiver covector $\lambda$ through $\langle\lambda,\dot Wz\rangle$. The actual globally realizable directions are obtained by restricting one \emph{global} parameter tangent to the selected cut variables. Local compatibility alone does not supply a global realization. Appendix~\ref{app:architecture-innovation-details} constructs this tangent image and proves the finite attainable-improvement bound on a regular parameter path.

\subsection{Learning and refinement need not commute}
The fine and coarse architectures can realize the same current predictor while the two optimizer processes apply different updates. Let $K_{c,t}$ and $K_{f,t}$ be their complete training kernels, and $J_t$ a declared coarse-to-fine kernel matching the states. In the forward-kernel convention, exact dynamical compatibility is
\[
K_{c,t}J_{t+1}=J_tK_{f,t}.
\]
The signed discrepancy between
these routes measures the failure
of training and architectural
refinement to commute.
Appendix~\ref{app:architecture-innovation-details}
derives its finite-time Duhamel
expansion and gives conditions
under which a fine-model tangent
direction is not inherited from
the coarse architecture and
produces a strict first-order
decrease in the selected objective.

Function-preserving network morphisms show how one
architecture can be transformed into another without
changing its realized predictor at the moment of
transformation \citep{wei2016networkmorphism}.
The geometry of the resulting parameterizations
need not agree, even when their functions do
\citep{dinh2017sharp}. The kernel comparison above
asks what happens after training resumes.
Equality of the present predictor fixes an
initial observational relation; the intertwining
condition tests whether the two implemented
learning processes preserve that relation over
subsequent updates. Its failure identifies
a dynamical distinction between architectures
that present-function equivalence does not
determine.

\section{Semantic trajectories across architectures}\label{sec:predictive-architecture-comparison}

Each architecture has its own
complete parameter--optimizer state
and training transition kernel.
A same-function lift matches
initial predictors, but need not
match the receiver-kernel profiles
or evaluated responses at selected
internal cuts.

To compare their subsequent
observable behavior, declare
corresponding measurable cut
probes \(\sigma_c\) and \(\sigma_f\).
For a comparison kernel
\(\Lambda_t\), the condition
\[
\Lambda_t
\bigl(
z,
\sigma_f^{-1}
\{\sigma_c(z)\}
\bigr)=1
\]
requires the matched states
to have equal selected observables.

Choosing the full finite cut probes
compares both evaluated receiver
responses and their equality
relations. Choosing only the
kernel-profile components compares
the corresponding indistinguishability
relations. Choosing task risk
instead compares task performance.

The backward information family $\mathcal H_t^{(\infty)}$ is derived separately from each architecture's transition kernel and selected observable. The state realizations of those predictive families may differ even where a same-function lift exists. Their future paths agree under the explicit intertwining conditions below.

\begin{proposition}[Exact observable-path intertwining criterion]\label{prop:intertwining-test}
Let $\Lambda_t:Z_L\rightsquigarrow Z_{mL}$ be a \emph{declared comparison kernel} whose parameter marginal agrees with the declared same-function lift. Suppose
\begin{equation}\label{eq:intertwining-test}
\Lambda_t K_{mL,t}=K_{L,t}\Lambda_{t+1}
\quad\text{and}\quad
\Lambda_t\!\left(z,\Phi_{mL}^{-1}\{\Phi_L(z)\}\right)=1
\quad(z\in Z_L).
\end{equation}
The comparison kernel preserves the retained observable almost surely at each state and intertwines the training transitions. Consequently, the fine experiment initialized from $\mu_L\Lambda_0$ and the coarse experiment have identical finite-dimensional joint laws of the selected $\Phi$-observations at matched training times. Taking $\Phi$ to be task risk yields equality of risk paths; taking $\Phi$ to be the declared contextual-semantic probe yields equality of its observable paths.
\end{proposition}
\begin{proof}
Kernel intertwining transports the complete law at every step by induction. Let $M_{\varphi\circ\Phi}$ denote multiplication by a bounded observable test. Pathwise preservation gives the operator identity $\Lambda_tM_{\varphi\circ\Phi_{mL}}=M_{\varphi\circ\Phi_L}\Lambda_t$. Interleave multiplication operators with consecutive optimizer kernels in the finite-dimensional joint-moment formula; moving each multiplication operator and transition through the comparison kernel yields equality of all products of bounded observable tests. Taking indicators gives equality of finite-dimensional laws.
\end{proof}

The corresponding identity for \emph{expectations only}, $\Lambda_t(\varphi\circ\Phi_{mL})=\varphi\circ\Phi_L$, is insufficient for the operator manipulation above unless it entails the stronger almost-sure preservation condition. ABF's deterministic same-function parameter lift preserves risk pathwise at the moment of comparison; what is not automatic is its intertwining with training on the two optimizer-state spaces.

\section{Learned organization and its dynamics}
\label{sec:part2-conclusion}

The resolved Transformer makes the
producer--receiver relations of Part~I
dynamical. At a selected parameterized
cut, the upstream computation supplies
the features on which the operation acts,
and the downstream continuation supplies
the covectors that determine its local
gradient. The optimizer updates the
represented processes on both sides.

The controlled learning models resolve
several mechanisms by which accessibility
changes. Matrix producer learning attenuates
represented directions at rates determined
by the data covariance spectrum.
Q/K-softmax learning produces an algebraic
attention-saturation law. In coupled learning,
producer alignment changes the allocation
preferred by attention, while attention
changes the learning force acting on the
producer. Direct input dependence of Q/K
scores provides a further mechanism:
the same attention parameters induce
different allocations under different
learned producer presentations.

The resulting computational observables
need not determine their own future.
The FFN factorization examples show that
identical present functions can possess
different tangent geometry and subsequent
updates. The semantic countdown establishes
an analogous failure for a coarse contextual
distinction profile. In the two-regime
attention construction, observing
accessibility under two input contexts
instead reconstructs a closed reduced
dynamical state. These calculations identify
what the selected observables measure,
how their values evolve, and which
additional distinctions are required
to predict that evolution.
\section{Conclusion}
\label{sec:paper-conclusion}

A computational operation has a realized
function, but its role at a selected cut
also depends on the represented process
that supplies its inputs and the continuation
that receives its outputs. Composing a
represented process and passing to its
extensional realization discard different
information. A subsequent distinction
quotient retains which predecessor states
are identified, not the entire computational
organization. These descriptions therefore
support different criteria of computational
identity.

Part~I derives those criteria and their
dependence on context. Restricted continuation
families can distinguish two invertibly
related presentations even though both
preserve the same predecessor information.
More strongly, the effective equivalence
of nominal receiver architectures is the
kernel of a context-indexed semantics map.
The injective local--attention construction
proves that these effective architecture
classes can vary with producer presentation
without any loss of predecessor distinctions.

Part~II studies how these relations evolve
under learning. The resolved PreNorm
Transformer separates branch-relative
accessibility from complete residual-state
identifications and identifies the upstream
features and downstream covectors entering
local gradients. Exact producer-routing
and Q/K experiments then give distinct
time laws for loss of bounded-receiver
accessibility. Coupled producer--attention
learning shows how the surrounding computation
changes both the receiver's current behavior
and its update signal.

Present equivalence does not in general
imply predictive equivalence. Equal realized
functions can have different future training
behavior, and equal current contextual
profiles can require additional complete-state
distinctions to predict their evolution.
The two-input-regime attention construction
exhibits an exact reduced predictive state
for a selected accessibility observable.
The appropriate computational description
is therefore determined by what is being
compared or predicted: the realized function,
retained information, available continuations,
or future observable behavior under the
implemented learning rule.

\appendix
\section{Dimension-labelled PreNorm Transformers as fully resolved cut systems}
\label{app:dimension-labelled-transformer}

This appendix refines the Transformer object used in
Part~\ref{part:trained-refinement} to the finest
architectural grain needed to make every internal
carrier, affine map, bias, normalization statistic,
attention contraction, feed-forward nonlinearity,
residual merge, and parameter-count coordinate explicit.
The attention and PreNorm conventions follow
\citet{vaswani2017attention} and
\citet{xiong2020layernorm}; the general carrier
dimensions, parameter-sharing declarations, and
retained cut structure are fixed by the present model.
Contracting the additional cuts introduced below
recovers the represented path of
Section~\ref{sec:transformer-interface-calculus}.

Two homogeneity assumptions are imposed and no others are used in the dimension bookkeeping:
\begin{enumerate}
    \item every hidden affine layer in every feed-forward branch has the same hidden width \(w\);
    \item within a fixed Transformer block \(\ell\), all attention heads have the same query dimension, key dimension, and value dimension. The parameters themselves remain head-specific.
\end{enumerate}
The Euclidean dot-product score forces the query and key dimensions in a given block to agree; we denote their common value by \(q_\ell\). No relation is imposed between \(q_\ell\), the value dimension \(v_\ell\), the residual width \(d\), the head count \(h_\ell\), the feed-forward depth \(n_\ell\), or the common feed-forward width \(w\). In particular, no head-packing identity such as \(h_\ell v_\ell=d\) is assumed.

\subsection{Global model and block coordinates}
\label{app:transformer-global-coordinates}

Let \(S\) be the finite token set and let
\[
V\cong\R^d,
\qquad
X=V^S
\]
be the residual carrier. The complete Transformer core is the represented composite
\[
F
=
B_{L-1}\cdots B_1B_0,
\]
where every \(B_\ell:X\to X\) is a PreNorm residual block. The identity residual carries force the same residual carrier to be used by every block, hence the residual width \(d\) is global. This is a consequence of the declared skip topology rather than an additional width-homogeneity assumption.

For each block \(\ell\in\{0,\ldots,L-1\}\), define the discrete--dimensional architecture coordinate
\begin{equation}
\label{eq:block-coordinate-tuple}
\kappa_\ell
=
\bigl(
    h_\ell,
    q_\ell,
    v_\ell,
    n_\ell;
    d,w;
    \Phi_\ell,
    \tau_\ell,
    \mathcal M_\ell,
    \varepsilon_{1,\ell},
    \varepsilon_{2,\ell}
\bigr).
\end{equation}
Here
\begin{itemize}
    \item \(h_\ell\in\N\) is the number of attention heads;
    \item \(q_\ell\in\N\) is the common query/key dimension of those heads;
    \item \(v_\ell\in\N\) is their common value dimension;
    \item \(n_\ell\in\N\) is the number of hidden nonlinear stages in the feed-forward branch;
    \item \(d\in\N\) is the global residual width;
    \item \(w\in\N\) is the common feed-forward hidden width;
    \item \(\Phi_\ell=(\phi_{\ell,1},\ldots,\phi_{\ell,n_\ell})\) is the ordered tuple of activation maps;
    \item \(\tau_\ell=(\tau_{\ell,1},\ldots,\tau_{\ell,h_\ell})\) is the tuple of positive attention-score temperatures;
    \item \(\mathcal M_\ell=
    (\mathcal M_{\ell,a,i})_{a,i}\) is the declared attention-admissibility structure, where
    \[
    \varnothing\ne
    \mathcal M_{\ell,a,i}\subseteq S
    \]
    is the source set visible to receiver token \(i\)
    in head \(a\), for every active receiver row.
    \item \(\varepsilon_{1,\ell},\varepsilon_{2,\ell}>0\) are the two LayerNorm stabilizers.
\end{itemize}
No injectivity, smoothness, monotonicity, coordinatewise action, homogeneity, or equality across hidden stages is imposed on the activations. Each
\[
\phi_{\ell,r}:\R^w\to\R^w
\]
is an arbitrary declared map. If an activation itself carries trainable parameters, let \(\Theta^\phi_{\ell,r}\) be its parameter space and let
\[
p^\phi_{\ell,r}:=\dim \Theta^\phi_{\ell,r}
\]
be its exact trainable parameter count. Parameter-free activations are the special case \(p^\phi_{\ell,r}=0\).

The complete architecture coordinate is
\[
\kappa
=
\bigl(L,d,w;\kappa_0,\ldots,\kappa_{L-1}\bigr),
\]
with repeated appearances of \(d\) and \(w\) in \(\kappa_\ell\) understood as references to the global values.

\subsection{LayerNorm as a represented internal factorization}
\label{app:layernorm-cut-factorization}

For \(r\in\{1,2\}\), block \(\ell\), and one token state \(x\in\R^d\), let the trainable LayerNorm parameters be
\[
\gamma_{r,\ell},\beta_{r,\ell}\in\R^d.
\]
Define
\begin{align}
\mu_{r,\ell}(x)
&=
\frac{1}{d}\mathbf 1^\top x,
\\
c_{r,\ell}(x)
&=
x-\mu_{r,\ell}(x)\mathbf 1,
\\
\nu_{r,\ell}(x)
&=
\frac{1}{d}\norm{c_{r,\ell}(x)}_2^2,
\\
r_{r,\ell}(x)
&=
\bigl(\nu_{r,\ell}(x)+\varepsilon_{r,\ell}\bigr)^{-1/2},
\\
\bar x_{r,\ell}
&=
r_{r,\ell}(x)c_{r,\ell}(x),
\\
N_{r,\ell}(x)
&=
\gamma_{r,\ell}\odot\bar x_{r,\ell}+\beta_{r,\ell}.
\end{align}
All operations are applied tokenwise on \(X=(\R^d)^S\). Thus a LayerNorm call has the represented refinement
\begin{equation}
\label{eq:layernorm-fine-cut-path}
x
\longrightarrow
(x,\mu)
\longrightarrow
(x,\mu,c)
\longrightarrow
(x,\mu,c,\nu)
\longrightarrow
(x,\mu,c,\nu,r)
\longrightarrow
(x,\bar x)
\longrightarrow
N(x).
\end{equation}
Contracting the intermediate carriers gives the single normalization map used in Section~\ref{sec:transformer-interface-calculus}.

The branch-only normalization map has the exact translation invariance
\[
N_{r,\ell}(x+a\mathbf 1)
=
N_{r,\ell}(x)
\qquad
(a\in\R),
\]
because centering removes the constant direction. Further identifications may be introduced by zero coordinates of \(\gamma_{r,\ell}\). None of these branch-level identifications is automatically a whole-state identification when the residual carrier is retained alongside the normalization branch.

Each LayerNorm contributes exactly
\begin{equation}
\label{eq:layernorm-parameter-count}
P^{\mathrm{LN}}_{r,\ell}=2d
\end{equation}
trainable parameters, namely \(d\) gains and \(d\) shifts. The stabilizer \(\varepsilon_{r,\ell}\) is part of the declared architecture coordinate and is not counted as trainable unless it is itself placed in the trainable parameter space.

\subsection{Fully typed attention with affine query, key, and value maps}
\label{app:attention-full-types}

Let
\[
\widehat H_\ell=N_{1,\ell}(H_\ell),
\qquad
H_\ell=(h_{\ell,i})_{i\in S},
\qquad
\widehat H_\ell=(\widehat h_{\ell,i})_{i\in S}.
\]
For each head \(a\in\{1,\ldots,h_\ell\}\), define the trainable affine query, key, and value maps
\begin{align}
Q_{\ell,a}(x)
&=
W^Q_{\ell,a}x+b^Q_{\ell,a},
&
W^Q_{\ell,a}&\in\R^{q_\ell\times d},
&
b^Q_{\ell,a}&\in\R^{q_\ell},
\\
K_{\ell,a}(x)
&=
W^K_{\ell,a}x+b^K_{\ell,a},
&
W^K_{\ell,a}&\in\R^{q_\ell\times d},
&
b^K_{\ell,a}&\in\R^{q_\ell},
\\
V_{\ell,a}(x)
&=
W^V_{\ell,a}x+b^V_{\ell,a},
&
W^V_{\ell,a}&\in\R^{v_\ell\times d},
&
b^V_{\ell,a}&\in\R^{v_\ell}.
\end{align}
The equality of the query and key output dimensions is forced by the Euclidean score pairing and is not an independent architecture assumption.

For receiver token \(i\) and source token \(j\), set
\begin{align}
q_{\ell,a,i}
&=
Q_{\ell,a}(\widehat h_{\ell,i})\in\R^{q_\ell},
\\
k_{\ell,a,j}
&=
K_{\ell,a}(\widehat h_{\ell,j})\in\R^{q_\ell},
\\
v_{\ell,a,j}
&=
V_{\ell,a}(\widehat h_{\ell,j})\in\R^{v_\ell}.
\end{align}
For every admissible source \(j\in\mathcal M_{\ell,a,i}\), define the raw score
\begin{equation}
\label{eq:exact-attention-logit}
s_{\ell,a,i,j}
=
\frac{
q_{\ell,a,i}^{\top}k_{\ell,a,j}
}{
\tau_{\ell,a}
}.
\end{equation}
The attention probability is
\begin{equation}
\label{eq:exact-attention-softmax}
\alpha_{\ell,a,i,j}
=
\frac{
\exp s_{\ell,a,i,j}
}{
\sum_{j'\in\mathcal M_{\ell,a,i}}\exp s_{\ell,a,i,j'}
},
\qquad
j\in\mathcal M_{\ell,a,i},
\end{equation}
and \(\alpha_{\ell,a,i,j}=0\) for \(j\notin\mathcal M_{\ell,a,i}\). This realizes masking by restricting the softmax domain rather than by introducing formal \(-\infty\) values. The nonempty-row condition ensures that every
active softmax denominator is strictly positive,
since all admissible scores are finite and
\(\tau_{\ell,a}>0\). No empty-row fallback
operator is included in the declared architecture.

Retain the source-resolved value contribution
\begin{equation}
\label{eq:source-resolved-value}
c_{\ell,a,i,j}
=
\alpha_{\ell,a,i,j}v_{\ell,a,j}
\in\R^{v_\ell}.
\end{equation}
Source aggregation gives
\begin{equation}
\label{eq:head-source-aggregation}
z_{\ell,a,i}
=
\sum_{j\in S}c_{\ell,a,i,j}
\in\R^{v_\ell}.
\end{equation}

Write the ordinary multi-head output matrix as a block decomposition
\[
W^O_\ell
=
\begin{bmatrix}
O_{\ell,1}&\cdots&O_{\ell,h_\ell}
\end{bmatrix},
\qquad
O_{\ell,a}\in\R^{d\times v_\ell},
\]
with one trainable output bias
\[
b^O_\ell\in\R^d.
\]
The head-resolved projected contribution is
\begin{equation}
\label{eq:head-projected-contribution}
p_{\ell,a,i}
=
O_{\ell,a}z_{\ell,a,i}
\in\R^d,
\end{equation}
and the attention residual update is
\begin{equation}
\label{eq:attention-residual-update}
t_{\ell,i}
=
b^O_\ell
+
\sum_{a=1}^{h_\ell}p_{\ell,a,i}
\in\R^d.
\end{equation}
The first residual merge is
\begin{equation}
\label{eq:first-residual-merge-exact}
u_{\ell,i}
=
h_{\ell,i}+t_{\ell,i}.
\end{equation}

\subsection{Attention cut lattice and contraction map}
\label{app:attention-cut-lattice}

Let
\begin{align*}
\mathsf Q_\ell
&=
\bigl(q_{\ell,a,i}\bigr)_{a,i},
&
\mathsf K_\ell
&=
\bigl(k_{\ell,a,j}\bigr)_{a,j},
&
\mathsf V_\ell
&=
\bigl(v_{\ell,a,j}\bigr)_{a,j},
\\
\mathsf S_\ell
&=
\bigl(s_{\ell,a,i,j}\bigr)_{a,i,j},
&
\mathsf A_\ell
&=
\bigl(\alpha_{\ell,a,i,j}\bigr)_{a,i,j},
&
\mathsf C_\ell
&=
\bigl(c_{\ell,a,i,j}\bigr)_{a,i,j},
\\
\mathsf Z_\ell
&=
\bigl(z_{\ell,a,i}\bigr)_{a,i},
&
\mathsf P_\ell
&=
\bigl(p_{\ell,a,i}\bigr)_{a,i},
&
\mathsf T_\ell
&=
\bigl(t_{\ell,i}\bigr)_i.
\end{align*}
The fully retained attention branch is generated by the dependency diagram
\begin{equation}
\label{eq:fully-resolved-attention-dependency}
\widehat H_\ell
\longrightarrow
(\mathsf Q_\ell,\mathsf K_\ell,\mathsf V_\ell)
\longrightarrow
(\mathsf S_\ell,\mathsf V_\ell)
\longrightarrow
(\mathsf A_\ell,\mathsf V_\ell)
\longrightarrow
\mathsf C_\ell
\longrightarrow
\mathsf Z_\ell
\longrightarrow
\mathsf P_\ell
\longrightarrow
\mathsf T_\ell.
\end{equation}
Every displayed arrow is a genuine represented contraction:
\begin{enumerate}
    \item the score map contracts the separate query/key vectors to their pairings;
    \item softmax contracts raw logits to normalized source weights;
    \item multiplication with values produces source-resolved value contributions;
    \item \(\mathsf C_\ell\to\mathsf Z_\ell\) contracts the source index;
    \item \(\mathsf Z_\ell\to\mathsf P_\ell\) applies the headwise blocks of the output projection;
    \item \(\mathsf P_\ell\to\mathsf T_\ell\) contracts the head index and adds the output bias.
\end{enumerate}
The corresponding whole-state carried path is
\begin{equation}
\label{eq:fully-resolved-attention-carried-path}
\begin{aligned}
H_\ell
&\longrightarrow
(H_\ell,\widehat H_\ell)
\longrightarrow
(H_\ell,\mathsf Q_\ell,\mathsf K_\ell,\mathsf V_\ell)
\\
&\longrightarrow
(H_\ell,\mathsf S_\ell,\mathsf V_\ell)
\longrightarrow
(H_\ell,\mathsf A_\ell,\mathsf V_\ell)
\longrightarrow
(H_\ell,\mathsf C_\ell)
\\
&\longrightarrow
(H_\ell,\mathsf Z_\ell)
\longrightarrow
(H_\ell,\mathsf P_\ell)
\longrightarrow
(H_\ell,\mathsf T_\ell)
\longrightarrow
U_\ell.
\end{aligned}
\end{equation}
The first component makes every carried map before the residual sum injective. Thus any branch-level contraction before \(U_\ell\) can alter receiver accessibility or geometry without enlarging the whole-state distinction relation.

\subsection{Exact attention symmetries and cut-relative gauge directions}
\label{app:attention-gauge-directions}

The refined attention cuts expose exact equivalences that are invisible after contraction.

For any head \(a\) and any \(A\in GL(q_\ell)\), transform the affine query and key coordinates by
\begin{align}
W^Q_{\ell,a}&\mapsto A W^Q_{\ell,a},
&
b^Q_{\ell,a}&\mapsto A b^Q_{\ell,a},
\\
W^K_{\ell,a}&\mapsto A^{-\top}W^K_{\ell,a},
&
b^K_{\ell,a}&\mapsto A^{-\top}b^K_{\ell,a}.
\end{align}
Then
\[
q_{\ell,a,i}^{\top}k_{\ell,a,j}
\]
is unchanged for every \(i,j\), hence the score cut, softmax cut, and every downstream attention output are unchanged. The query/key carrier therefore contains a \(GL(q_\ell)\) re-presentation freedom that the score contraction forgets.

For fixed \((\ell,a,i)\), softmax has the exact additive invariance
\[
\operatorname{softmax}
\bigl((s_j+c)_{j\in\mathcal M_{\ell,a,i}}\bigr)
=
\operatorname{softmax}
\bigl((s_j)_{j\in\mathcal M_{\ell,a,i}}\bigr).
\]
Thus the logit cut contains one additive gauge direction per receiver/head admissible-source simplex. These are semantic identifications of the softmax receiver; they are not identifications of the preceding query/key represented computation.

Source aggregation further identifies source-resolved tensors satisfying
\[
\sum_j c_{\ell,a,i,j}
=
\sum_j c'_{\ell,a,i,j}
\qquad
\forall a,i,
\]
and head aggregation identifies projected head tuples satisfying
\[
\sum_a p_{\ell,a,i}
=
\sum_a p'_{\ell,a,i}
\qquad
\forall i.
\]
The residual merge finally identifies pair states whenever
\[
H_\ell+\mathsf T_\ell
=
H'_\ell+\mathsf T'_\ell.
\]
These are distinct quotient operations at distinct cuts.

\subsection{Exact attention carrier dimensions}
\label{app:attention-carrier-dimensions}

Let \(N=|S|\), and for each \((\ell,a,i)\) write
\[
m_{\ell,a,i}:=|\mathcal M_{\ell,a,i}|.
\]
The Euclidean dimensions of the principal attention carriers are
\begin{align}
\dim H_\ell
&=Nd,
\\
\dim \mathsf Q_\ell
&=Nh_\ell q_\ell,
&
\dim \mathsf K_\ell
&=Nh_\ell q_\ell,
&
\dim \mathsf V_\ell
&=Nh_\ell v_\ell,
\\
\dim \mathsf S_\ell
&=
\sum_{a=1}^{h_\ell}\sum_{i\in S}m_{\ell,a,i},
&
\dim \mathsf A_\ell
&=
\sum_{a=1}^{h_\ell}\sum_{i\in S}m_{\ell,a,i},
\\
\dim \mathsf C_\ell
&=
v_\ell
\sum_{a=1}^{h_\ell}\sum_{i\in S}m_{\ell,a,i},
\\
\dim \mathsf Z_\ell
&=Nh_\ell v_\ell,
&
\dim \mathsf P_\ell
&=Nh_\ell d,
&
\dim \mathsf T_\ell
&=Nd.
\end{align}
The softmax image itself lies in a product of simplices of intrinsic dimension
\begin{equation}
\label{eq:attention-simplex-intrinsic-dim}
\sum_{a=1}^{h_\ell}\sum_{i\in S}(m_{\ell,a,i}-1),
\end{equation}
although its ambient coordinate representation has the dimension displayed above.

\subsection{Fully resolved feed-forward branch with arbitrary activation maps}
\label{app:ffn-full-resolution}

Let
\[
\widehat U_\ell=N_{2,\ell}(U_\ell).
\]
The shared hidden width assumption gives one hidden carrier \(\R^w\) at every nonlinear stage. Define
\[
z_{\ell,0,i}:=\widehat u_{\ell,i}\in\R^d.
\]
For the first hidden stage,
\begin{align}
g_{\ell,1,i}
&=
W_{\ell,1}z_{\ell,0,i}+b_{\ell,1}
\in\R^w,
\\
z_{\ell,1,i}
&=
\phi_{\ell,1}(g_{\ell,1,i})
\in\R^w,
\end{align}
with
\[
W_{\ell,1}\in\R^{w\times d},
\qquad
b_{\ell,1}\in\R^w.
\]
For \(2\le r\le n_\ell\),
\begin{align}
g_{\ell,r,i}
&=
W_{\ell,r}z_{\ell,r-1,i}+b_{\ell,r}
\in\R^w,
\\
z_{\ell,r,i}
&=
\phi_{\ell,r}(g_{\ell,r,i})
\in\R^w,
\end{align}
where
\[
W_{\ell,r}\in\R^{w\times w},
\qquad
b_{\ell,r}\in\R^w.
\]
The output affine map is
\begin{equation}
\label{eq:ffn-output-affine-full}
f_{\ell,i}
=
W_{\ell,n_\ell+1}z_{\ell,n_\ell,i}
+
b_{\ell,n_\ell+1}
\in\R^d,
\end{equation}
with
\[
W_{\ell,n_\ell+1}\in\R^{d\times w},
\qquad
b_{\ell,n_\ell+1}\in\R^d.
\]
The second residual merge is
\begin{equation}
\label{eq:second-residual-merge-exact}
h_{\ell+1,i}
=
u_{\ell,i}+f_{\ell,i}.
\end{equation}

The branch-only path is
\begin{equation}
\label{eq:fully-resolved-ffn-branch}
\widehat U_\ell
\longrightarrow
\mathsf G_{\ell,1}
\longrightarrow
\mathsf Z_{\ell,1}
\longrightarrow
\cdots
\longrightarrow
\mathsf G_{\ell,n_\ell}
\longrightarrow
\mathsf Z_{\ell,n_\ell}
\longrightarrow
\mathsf F_\ell,
\end{equation}
where \(\mathsf G_{\ell,r}=(g_{\ell,r,i})_i\), \(\mathsf Z_{\ell,r}=(z_{\ell,r,i})_i\), and \(\mathsf F_\ell=(f_{\ell,i})_i\). The carried whole-state path is
\begin{equation}
\label{eq:fully-resolved-ffn-carried}
\begin{aligned}
U_\ell
&\longrightarrow
(U_\ell,\widehat U_\ell)
\longrightarrow
(U_\ell,\mathsf G_{\ell,1})
\longrightarrow
(U_\ell,\mathsf Z_{\ell,1})
\\
&\longrightarrow\cdots\longrightarrow
(U_\ell,\mathsf G_{\ell,n_\ell})
\longrightarrow
(U_\ell,\mathsf Z_{\ell,n_\ell})
\longrightarrow
(U_\ell,\mathsf F_\ell)
\longrightarrow
H_{\ell+1}.
\end{aligned}
\end{equation}
No property of \(\phi_{\ell,r}\) is required for the carried maps before the residual merge to remain injective: \(U_\ell\) itself remains an explicit component. Branch-relative distinction growth at \(\mathsf G_{\ell,r}\to\mathsf Z_{\ell,r}\) is determined exactly by the kernel pair of the declared activation \(\phi_{\ell,r}\).

\subsection{Fully resolved Transformer block cuts}
\label{app:maximally-refined-block-cut-frame}

Combining the two PreNorm branches gives the maximally retained block-level path
\begin{equation}
\label{eq:maximal-block-path}
\begin{aligned}
H_\ell
&\to
\mathsf{LN}^{(1)}_\ell
\to
(H_\ell,\mathsf Q_\ell,\mathsf K_\ell,\mathsf V_\ell)
\to
(H_\ell,\mathsf S_\ell,\mathsf V_\ell)
\to
(H_\ell,\mathsf A_\ell,\mathsf V_\ell)
\\
&\to
(H_\ell,\mathsf C_\ell)
\to
(H_\ell,\mathsf Z_\ell)
\to
(H_\ell,\mathsf P_\ell)
\to
(H_\ell,\mathsf T_\ell)
\to
U_\ell
\\
&\to
\mathsf{LN}^{(2)}_\ell
\to
(U_\ell,\mathsf G_{\ell,1})
\to
(U_\ell,\mathsf Z_{\ell,1})
\to\cdots\to
(U_\ell,\mathsf G_{\ell,n_\ell})
\\
&\to
(U_\ell,\mathsf Z_{\ell,n_\ell})
\to
(U_\ell,\mathsf F_\ell)
\to
H_{\ell+1},
\end{aligned}
\end{equation}
where \(\mathsf{LN}^{(r)}_\ell\) abbreviates the explicitly refined LayerNorm path in Equation~\eqref{eq:layernorm-fine-cut-path}, with the bypass carrier retained at every intermediate cut. Every coarser Transformer description used in the main text is obtained by composing consecutive arrows of Equation~\eqref{eq:maximal-block-path}.

Let \(E_{\ell,r}\) denote the whole-state distinction relation induced from the model input at an internal cut \(r\). Every carried cut before a residual sum has the same whole-state distinction relation as the input to that residual branch. Hence the exact whole-state chain refines Equation~\eqref{eq:prenorm-information-profile} to
\begin{equation}
\label{eq:maximal-whole-state-profile}
\boxed{
E_\ell
=
E_{\ell,r}
\quad
\text{for every retained carried cut }r\text{ before the first residual merge},
}
\end{equation}
then
\[
E_\ell\subseteq E_{\ell,U},
\]
followed by
\begin{equation}
\label{eq:maximal-whole-state-profile-ffn}
\boxed{
E_{\ell,U}
=
E_{\ell,r}
\quad
\text{for every retained carried cut }r\text{ before the second residual merge},
}
\end{equation}
and finally
\[
E_{\ell,U}\subseteq E_{\ell+1}.
\]
Therefore exact whole-state identification can be created inside a block only at the two residual additions. All earlier contractions can still change branch-relative information, geometry, accessible continuation, tangent innovations, or receiver-weighted work.

\subsection{Exact trainable parameter count of one block}
\label{app:exact-block-parameter-count}

The parameter count is derived from the typed maps above rather than imposed as a primitive coordinate.

For one head in block \(\ell\), query and key weights contribute \(2q_\ell d\), query and key biases contribute \(2q_\ell\), the value weight and bias contribute \(v_\ell d+v_\ell\), and the corresponding block of the output matrix contributes \(dv_\ell\). Summing over heads and adding the single multi-head output bias gives
\begin{equation}
\label{eq:exact-attention-parameter-count}
\boxed{
P^{\mathrm{att}}_\ell
=
2h_\ell q_\ell d
+
2h_\ell q_\ell
+
2h_\ell d v_\ell
+
h_\ell v_\ell
+
d.
}
\end{equation}

The two LayerNorms contribute
\begin{equation}
\label{eq:exact-two-ln-count}
P^{\mathrm{norm}}_\ell=4d.
\end{equation}

The feed-forward weights contribute
\[
dw
+
(n_\ell-1)w^2
+
wd,
\]
and the affine biases contribute
\[
n_\ell w+d.
\]
If the activation maps carry trainable parameter spaces, their contribution is
\[
P^\phi_\ell
:=
\sum_{r=1}^{n_\ell}p^\phi_{\ell,r}.
\]
Thus
\begin{equation}
\label{eq:exact-ffn-parameter-count}
\boxed{
P^{\mathrm{ff}}_\ell
=
2dw
+
(n_\ell-1)w^2
+
n_\ell w
+
d
+
P^\phi_\ell.
}
\end{equation}

Combining attention, normalization, and feed-forward parameter slots gives the exact untied block count
\begin{equation}
\label{eq:exact-block-total-count}
\boxed{
\begin{aligned}
P_\ell
={}&
2h_\ell d(q_\ell+v_\ell)
+
2h_\ell q_\ell
+
h_\ell v_\ell
\\
&+
2dw
+
(n_\ell-1)w^2
+
n_\ell w
+
6d
+
P^\phi_\ell.
\end{aligned}
}
\end{equation}
The \(6d\) term consists exactly of the attention output bias \(d\), the two LayerNorm gain/shift pairs \(4d\), and the feed-forward output bias \(d\).

The complete Transformer core count for \emph{untied} blocks is
\begin{equation}\label{eq:exact-transformer-core-count}
\boxed{P_{\mathrm{core}}^{\mathrm{untied}}(\kappa)=\sum_{\ell=0}^{L-1}P_\ell.}
\end{equation}
For any parameter tying within or between blocks, let $\Theta_{\rm core}^{\rm tied}$ be the compatible global parameter locus inside the product of untied block parameter-slot spaces. The number of independent trainable core coordinates is $\dim\Theta_{\rm core}^{\rm tied}$; the displayed sum counts local coordinate occurrences instead. For coordinate-identification ties, this dimension equals the number of equivalence classes of local coordinates. For more general smooth constraints it is the tangent-space dimension on the specified regular locus.
If the represented model includes trainable boundary maps such as embeddings, positional parameters, a final normalization, or a readout, collect their exact declared parameter set in \(\Theta_\partial\) and define
\[
P_\partial:=\dim\Theta_\partial.
\]
For disjoint untied boundary and core coordinates, the total count is
\begin{equation}\label{eq:exact-total-model-count}
\boxed{P_{\mathrm{tot}}^{\mathrm{untied}}=P_\partial+P_{\mathrm{core}}^{\mathrm{untied}}.}
\end{equation}
For any globally tied model, including boundary-to-core ties, its exact count is $P_{\mathrm{tot}}=\dim\Theta_{\mathrm{tot}}^{\mathrm{compatible}}$. This counts each independent coordinate once.

The scale--shape reconstruction below is a coordinate analysis of the \emph{untied} architecture family. It does not specify a training trajectory or assert an asymptotic scaling law.

\subsection{Scale--shape coordinates with all bias terms retained}
\label{app:scale-shape-coordinates}

The raw dimension tuple can be rewritten as one absolute width scale together with dimensionless shape coordinates. On the locus \(d>0\), define
\begin{equation}
\label{eq:dimension-ratio-coordinates}
\rho^q_\ell
:=
\frac{q_\ell}{d},
\qquad
\rho^v_\ell
:=
\frac{v_\ell}{d},
\qquad
\rho^w
:=
\frac{w}{d}.
\end{equation}
The head counts \(h_\ell\) and hidden depths \(n_\ell\) remain discrete architecture coordinates. The activation tuple, temperatures, masks, normalization stabilizers, and parameter-sharing declarations remain structural coordinates.

Write
\[
P^\phi
=
\sum_{\ell=0}^{L-1}P^\phi_\ell.
\]
For untied blocks, substituting
\[
q_\ell=\rho^q_\ell d,
\qquad
v_\ell=\rho^v_\ell d,
\qquad
w=\rho^w d
\]
into Equation~\eqref{eq:exact-transformer-core-count} gives the exact decomposition
\begin{equation}
\label{eq:exact-scale-shape-count}
\boxed{
P_{\mathrm{core}}^{\mathrm{untied}}-P^\phi
=
A(\rho,h,n)\,d^2
+
B(\rho,h,n)\,d,
}
\end{equation}
where
\begin{align}
A(\rho,h,n)
&=
\sum_{\ell=0}^{L-1}
\left[
2h_\ell(\rho^q_\ell+\rho^v_\ell)
+
2\rho^w
+
(n_\ell-1)(\rho^w)^2
\right],
\label{eq:shape-quadratic-coefficient}
\\
B(\rho,h,n)
&=
\sum_{\ell=0}^{L-1}
\left[
2h_\ell\rho^q_\ell
+
h_\ell\rho^v_\ell
+
n_\ell\rho^w
+
6
\right].
\label{eq:shape-linear-coefficient}
\end{align}
All bias and normalization terms are retained in \(B\). No large-width approximation has been taken.

For fixed discrete/structural coordinates and fixed activation-parameter count, the positive residual width is recovered exactly from the scalar core parameter count and the shape coordinates:
\begin{equation}
\label{eq:exact-d-from-P-and-ratios}
\boxed{
d
=
\frac{
-B
+
\sqrt{B^2+4A(P_{\mathrm{core}}^{\mathrm{untied}}-P^\phi)}
}{2A},
}
\end{equation}
whenever \(A>0\). The remaining widths are then
\begin{equation}
\label{eq:recover-dimensions-from-ratios}
q_\ell=\rho^q_\ell d,
\qquad
v_\ell=\rho^v_\ell d,
\qquad
w=\rho^w d.
\end{equation}
The actual architecture space is the integral locus on which these recovered dimensions are positive integers. Thus \((P_{\mathrm{core}}^{\mathrm{untied}},\rho)\) is a continuous coordinate chart for the ambient algebraic architecture variety, while realizable neural architectures occupy its arithmetic sublocus.

Equation~\eqref{eq:exact-scale-shape-count} makes the scalarization precise for untied block parameter coordinates. Parameter tying changes the independent-coordinate count and requires applying the shared-coordinate identifications before any inversion from total parameter number to width. The map
\[
\kappa
\longmapsto
P_{\mathrm{core}}^{\mathrm{untied}}
\]
forgets the ratio coordinates, head decomposition, hidden depth, activation choice, masks, temperatures, and all represented cut structure. Conversely, \(P_{\mathrm{core}}^{\mathrm{untied}}\) together with the full shape/structural coordinate recovers the dimension tuple through Equations~\eqref{eq:exact-d-from-P-and-ratios}--\eqref{eq:recover-dimensions-from-ratios}.

\subsection{Architecture constraint variety and its matroidal dependency structure}
\label{app:architecture-matroid}

Fix the discrete coordinates
\[
L,
\qquad
(h_\ell)_{\ell=0}^{L-1},
\qquad
(n_\ell)_{\ell=0}^{L-1},
\]
and fix the activation-parameter count \(P^\phi\) and all non-dimensional structural data. Consider the ambient coordinate set
\begin{equation}
\label{eq:architecture-ground-coordinates}
\mathcal E_{\mathrm{arch}}
=
\left
\{
P_{\mathrm{core}}^{\mathrm{untied}},d,w,
q_0,v_0,
\ldots,
q_{L-1},v_{L-1}
\right\}.
\end{equation}
The exact parameter-count relation is the polynomial constraint
\begin{equation}
\label{eq:architecture-count-polynomial}
\begin{aligned}
0
=
F_{\mathrm{arch}}
:={}&
P_{\mathrm{core}}^{\mathrm{untied}}-P^\phi
\\
&-
\sum_{\ell=0}^{L-1}
\Bigl[
2h_\ell d(q_\ell+v_\ell)
+2h_\ell q_\ell+h_\ell v_\ell
\\
&\hspace{8em}
+2dw+(n_\ell-1)w^2+n_\ell w+6d
\Bigr].
\end{aligned}
\end{equation}
Let \(\mathcal V_{\mathrm{arch}}\) be the affine variety cut out by \(F_{\mathrm{arch}}=0\) over a field of characteristic zero, with the physically realized neural architectures given by its positive integral points together with the fixed discrete structure.

The coordinate functions in \(\mathcal E_{\mathrm{arch}}\) determine an algebraic matroid
\[
M_{\mathrm{arch}}
:=
M(\mathcal V_{\mathrm{arch}}).
\]
A subset \(A\subseteq\mathcal E_{\mathrm{arch}}\) is independent when its coordinate functions are algebraically independent on \(\mathcal V_{\mathrm{arch}}\). A basis is a maximal generically free set of architecture coordinates; the remaining coordinate is algebraically determined by the parameter-count constraint. Because \(F_{\mathrm{arch}}\) is affine-linear in \(P_{\mathrm{core}}^{\mathrm{untied}}\) with unit coefficient, it is irreducible and defines a codimension-one hypersurface. Hence
\begin{equation}
\label{eq:architecture-matroid-rank}
\operatorname{rk}M_{\mathrm{arch}}
=
|\mathcal E_{\mathrm{arch}}|-1.
\end{equation}
Additional architectural identities, if later imposed, add polynomial constraints and can reduce this rank. For example, a head-packing condition \(h_\ell v_\ell=d\) would be an additional constraint; it is not assumed here.

The equalities forced by the represented wiring define a separate linear dependency system before parameter count is introduced. Identity residual composition identifies every residual carrier dimension with the same variable \(d\); the dot-product score identifies query and key carrier dimensions within each block; the shared-width condition identifies every feed-forward hidden carrier dimension with \(w\); and head homogeneity identifies the corresponding head dimensions within a block. These wiring equalities can be represented by an integer constraint matrix \(C_{\mathrm{wire}}\). Its row space records the independent dimension-identification constraints, while the quotient by that row space gives the free dimension-coordinate space on which Equation~\eqref{eq:architecture-count-polynomial} is imposed. Thus the full architecture-coordinate dependency has a linear wiring layer followed by the algebraic parameter-count layer.

\subsection{Cut-relative tangent dimensions and innovation subspace rank}
\label{app:innovation-matroid}

The dimension-labelled cut system also exposes a subspace-rank structure at the level relevant to refinement. Fix a same-function refinement \(\mathfrak a\to\mathfrak a'\) and a finite set \(J\) of retained inserted cuts. For each \(j\in J\), let
\[
\mathcal U_j
\subseteq
T_{r(\theta)}\mathcal H
\]
be the linear space of first-order predictor perturbations generated by opening the inserted output degrees of freedom at cut \(j\), after quotienting parameter directions in \(\ker Dr\). Define
\begin{equation}
\label{eq:innovation-rank-function}
r_{\mathrm{inn}}(A)
:=
\dim\operatorname{span}
\left(
\bigcup_{j\in A}\mathcal U_j
\right),
\qquad
A\subseteq J.
\end{equation}
Because \(r_{\mathrm{inn}}\) is a linear-subspace rank, it satisfies normalization, monotonicity, and submodularity and therefore defines a representable polymatroid rank on the cut labels \(J\). It is not in general a matroid rank on \(J\), since a single cut can expose more than one independent direction. Choosing basis-vector ground elements for each subspace yields an ordinary represented matroid on those expanded elements, while the cut-labelled rank in Equation~\eqref{eq:innovation-rank-function} remains the subspace polymatroid. Overlap among cut subspaces describes redundant exposed predictor directions without identifying one cut with one matroid element.

Receiver weighting adds a second reduction, but not by taking a linear subspace of ``nonzero-work'' directions, since that set is generally not closed under addition. If \(\lambda\) is the pulled-back loss covector, define the receiver-invisible hyperplane
\[
\ker\lambda
=
\{u:\langle\lambda,u\rangle=0\}
\]
and the active cut set
\begin{equation}
\label{eq:receiver-active-cut-set}
J_\lambda
=
\{j\in J:\mathcal U_j\not\subseteq\ker\lambda\}.
\end{equation}
Equivalently, the receiver-visible quotient of the innovation space at cut \(j\) is
\[
\overline{\mathcal U}^{(\lambda)}_j
=
\mathcal U_j/(\mathcal U_j\cap\ker\lambda).
\]
A cut belongs to \(J_\lambda\) exactly when this quotient is nonzero. The scalar first-order work is not a matroid or polymatroid rank; receiver weighting selects cut subspaces not annihilated by \(\lambda\). The gluing constraint of Section~\ref{sec:innovation-gluing} then further restricts locally available combinations to those lying in the image of the global section map. Architecture, receiver visibility, and global realizability are therefore three distinct reductions of the same cut-indexed innovation structure.

\subsection{Homogeneous low-dimensional slice of the architecture space}
\label{app:recover-main-text-transformer}

A familiar homogeneous Transformer family is obtained by the additional specialization
\[
h_\ell=h,
\qquad
q_\ell=v_\ell=d_h,
\qquad
n_\ell=n,
\qquad
\phi_{\ell,r}=\phi,
\qquad
P^\phi_\ell=0,
\]
for every block and hidden stage. Contracting the fine LayerNorm, query/key/value, score, softmax, preactivation, and activation cuts gives the corresponding low-dimensional block description. Under this specialization, Equation~\eqref{eq:exact-block-total-count} becomes
\begin{equation}
\label{eq:main-text-count-with-biases}
P_\ell
=
4hdd_h
+
3hd_h
+
2dw
+
(n-1)w^2
+
nw
+
6d.
\end{equation}
The leading quadratic part
\[
4hdd_h+2dw+(n-1)w^2
\]
is the leading quadratic block count; Equation~\eqref{eq:main-text-count-with-biases} supplies the exact affine-bias and normalization correction for the fully specified block above.

The homogeneous slice is therefore a projection of the fully resolved architecture:
\[
\boxed{
\begin{gathered}
\text{dimension-labelled represented Transformer}
\longrightarrow
\text{selected cut contraction}
\\
\longrightarrow
\text{architecture scale/shape coordinates}
\longrightarrow
\text{scalar parameter count}.
\end{gathered}
}
\]
The scalar \(P\) is the last coordinate in this chain. The represented cut system, its dimension constraints, and its ratio coordinates are the structures that determine what information has been discarded in passing to that scalar.

\section{Maximally resolved cut relativity, exact block matching, and multi-block composition}
\label{app:maximal-cut-relativity}

This appendix resolves the internal
computation graph of a dimension-labelled
PreNorm Transformer block and compares
its intermediate cuts.

For each legal computation frontier, we
retain the represented producer and
continuation, the induced kernel pair,
the available downstream computations,
and the quantities needed for
differentiation and training.

Composition then contracts the internal
interfaces to a block-level map.
The retained frontier data permit
questions about internal information
accessibility, residual-state persistence,
and learning gradients to be answered
after that contraction.

The same construction extends to
two-block, three-block, and arbitrary
finite block segments.

No additional architectural homogeneity assumptions are introduced here. The block coordinates, affine biases, LayerNorm parameters, arbitrary activation maps, masks, temperatures, parameter-sharing declarations, and exact parameter counts are those of Appendix~\ref{app:dimension-labelled-transformer}. In particular, the only width assumptions remain the shared feed-forward hidden width and homogeneous dimensional specification across heads within a block. The cut analysis itself does not require differentiability of the activation maps. Derivatives enter only when the receiver pullback or trained-gradient specialization is explicitly invoked.

\subsection{The fully resolved block is a finite acyclic represented process}
\label{app:maximal-cut-dag}

Fix one block index \(\ell\). Let
\[
X_\ell=(\R^d)^S
\]
be the residual carrier at the block input and output. The fully resolved block of Appendix~\ref{app:dimension-labelled-transformer} determines a finite acyclic represented computation graph
\[
\mathscr G_\ell.
\]
Its source wire is \(H_\ell\in X_\ell\), its sink wire is \(H_{\ell+1}\in X_\ell\), and its operation nodes are the primitive represented operations listed below.

The first PreNorm attention half contains, in dependency order,
\begin{enumerate}
    \item the first LayerNorm mean, centering, variance, inverse-scale, normalized-state, and affine gain/shift operations;
    \item the affine query, key, and value maps for every head;
    \item the admissible query--key score pairings, including the declared positive temperatures and masks;
    \item softmax on each admissible source simplex;
    \item multiplication of attention weights with value vectors;
    \item source aggregation inside each head;
    \item headwise output projection;
    \item head aggregation and addition of the multi-head output bias;
    \item the first residual addition.
\end{enumerate}
The second PreNorm feed-forward half contains
\begin{enumerate}
    \item the second LayerNorm factorization;
    \item for each hidden stage \(r\), the affine preactivation map and the arbitrary declared activation \(\phi_{\ell,r}\);
    \item the final affine feed-forward output map;
    \item the second residual addition.
\end{enumerate}
Fan-out is part of the represented wiring. Whenever one represented state is consumed by more than one later operation, the corresponding diagonal/product presentation is retained, so the graph records the actual simultaneous availability of that state to the downstream branches.

Let \(\mathcal O_\ell\) be the finite poset of primitive operation occurrences ordered by reachability in \(\mathscr G_\ell\). A subset
\[
I\subseteq\mathcal O_\ell
\]
is \emph{completed} when it is an order ideal: whenever \(o\in I\), every predecessor operation of \(o\) is also in \(I\). Let
\[
\mathfrak F_\ell
:=
J(\mathcal O_\ell)
\]
denote the finite poset of order ideals, ordered by inclusion. By the standard order-ideal construction, \(\mathfrak F_\ell\) is a finite distributive lattice with
\[
I\wedge J=I\cap J,
\qquad
I\vee J=I\cup J.
\]

For \(I\in\mathfrak F_\ell\), define the
\emph{live frontier} \(\partial I\) using the
designated external input and output interfaces
of the represented block graph.

A wire belongs to \(\partial I\) when its value
is available after execution of the completed
operations \(I\) and at least one of the following
holds:
\begin{enumerate}
    \item it is required by an operation outside \(I\);
    \item it belongs to the designated external output
    interface of the block.
\end{enumerate}
Availability includes externally supplied source
wires and wires produced by operations in \(I\).
A wire whose producer has not executed is not
available at the frontier.

Shared live values are retained once, and jointly
live values are represented by the corresponding
product object. The source and sink wires are
therefore treated as genuine external interfaces,
rather than requiring an internal producer or
consumer.

In particular, the initial frontier retains the
externally supplied block input, while the terminal
frontier retains the designated block output,
even though no internal operations remain unfinished. Write
\[
X_{\ell,I}
=
\prod_{e\in\partial I}X_e
\]
for its frontier carrier. The represented forward execution of the completed region gives a prefix
\[
Q_{\ell,I}:
X_\ell\longrightarrow X_{\ell,I},
\]
and completion of the unexecuted region gives a suffix
\[
S_{\ell,I}:
X_{\ell,I}\longrightarrow X_\ell.
\]
By construction,
\begin{equation}
\label{eq:frontier-factorization}
\boxed{
S_{\ell,I}Q_{\ell,I}=B_\ell.
}
\end{equation}

The endpoint conventions give
\[
\begin{aligned}
X_{\ell,\varnothing}
&\cong X_\ell,
&
X_{\ell,\mathcal O_\ell}
&\cong X_\ell,\\
Q_{\ell,\varnothing}
&=\operatorname{id}_{X_\ell},
&
S_{\ell,\varnothing}
&=B_\ell,\\
Q_{\ell,\mathcal O_\ell}
&=B_\ell,
&
S_{\ell,\mathcal O_\ell}
&=\operatorname{id}_{X_\ell},
\end{aligned}
\]
after the canonical identifications of the
external carrier types.
Hence the initial frontier represents the complete
unexecuted block, and the terminal frontier
represents its completed output. Both satisfy
Equation~\eqref{eq:frontier-factorization}.
Thus every live frontier is an object of
\[
\operatorname{Fact}_{\mathcal A}(B_\ell).
\]

\begin{proposition}[Frontier cut functor]
\label{prop:frontier-cut-functor}
For \(I\subseteq J\) in \(\mathfrak F_\ell\), execution of the operations in \(J\setminus I\) induces a represented map
\[
T_{I:J}:X_{\ell,I}\to X_{\ell,J}
\]
satisfying
\[
T_{I:J}Q_{\ell,I}=Q_{\ell,J},
\qquad
S_{\ell,I}=S_{\ell,J}T_{I:J}.
\]
Hence
\[
\boxed{
C_\ell^{\max}:
\mathfrak F_\ell
\longrightarrow
\operatorname{Fact}_{\mathcal A}(B_\ell),
\qquad
I\longmapsto(Q_{\ell,I},S_{\ell,I})
}
\]
is a functor.
\end{proposition}

\begin{proof}
Because \(I\subseteq J\) are order ideals, every operation needed to produce the \(J\)-frontier from the \(I\)-frontier lies in the finite acyclic region \(J\setminus I\). Their represented composite is \(T_{I:J}\). Executing the block prefix through \(I\) and then through \(J\setminus I\) is exactly the prefix through \(J\), giving the first identity. The remaining suffix from \(I\) factors through the later frontier \(J\), giving the second. Identity and composition follow from empty execution and associativity of represented composition.
The endpoint cases use the designated external
interfaces. For \(I=\varnothing\), execution begins
with the externally supplied source carrier. For
\(J=\mathcal O_\ell\), it ends at the designated
output carrier. Thus the same composition identities
hold when either frontier is an endpoint, and the
construction defines a functor on the entire
order-ideal lattice.
\end{proof}

The chain written in Equation~\eqref{eq:maximal-block-path} of Appendix~\ref{app:dimension-labelled-transformer} is therefore one maximal chain through the richer lattice \(\mathfrak F_\ell\). Different legal evaluation orders of parallel query, key, and value computations give different maximal chains but the same frontier lattice and the same factorization functor. The maximally resolved cut system is consequently independent of an arbitrary serialization of the Transformer DAG.

\subsection{Part~1 data attached to every fine frontier}
\label{app:fine-frontier-signature}

Fix deterministic Set realization for the distinction analysis. At each frontier \(I\in\mathfrak F_\ell\), define the local whole-state distinction relation
\begin{equation}
\label{eq:frontier-whole-kernel}
\boxed{
E^{\mathrm{whole}}_{\ell,I}
:=
\mathsf D_{X_\ell}(Q_{\ell,I})
=
\ker R(Q_{\ell,I}).
}
\end{equation}
The corresponding exact semantic quotient is
\[
q_{\ell,I}:
R(X_\ell)\twoheadrightarrow
R(X_\ell)/E^{\mathrm{whole}}_{\ell,I}.
\]
By Theorem~\ref{thm:exact-quotient-classification}, this quotient is canonical up to unique isomorphism over \(R(X_\ell)\).

The receiver side of the same cut is the
represented suffix \(S_{\ell,I}\).
Its realized continuation family is initially
typed on the frontier carrier
\(R(X_{\ell,I})\), whereas the exact quotient
\(q_{\ell,I}\) has codomain
\(R(X_\ell)/E^{\mathrm{whole}}_{\ell,I}\).
We therefore transport the continuation family
to the effective quotient before comparing
attainable branches.

The canonical quotient--image isomorphism is
\[
j_{\ell,I}:
R(X_\ell)/E^{\mathrm{whole}}_{\ell,I}
\xrightarrow{\;\cong\;}
\operatorname{im}R(Q_{\ell,I}).
\]
Let
\[
\iota_{\ell,I}:
\operatorname{im}R(Q_{\ell,I})
\hookrightarrow R(X_{\ell,I})
\]
be the image inclusion.

For a declared downstream output object \(W\),
let
\[
\mathcal G_{\ell,I}
\subseteq
\operatorname{Hom}_{\mathcal S}
(R(X_{\ell,I}),W)
\]
be the realized restricted continuation family
supplied by the represented architecture.

Define the quotient-transported family
\[
\overline{\mathcal G}_{\ell,I}
:=
\left\{
g\circ\iota_{\ell,I}\circ j_{\ell,I}
\;\middle|\;
g\in\mathcal G_{\ell,I}
\right\}.
\]

The attained complete branches are then
\[
\boxed{
\mathfrak B
(q_{\ell,I},\overline{\mathcal G}_{\ell,I}).
}
\]

Indeed, every \(g\in\mathcal G_{\ell,I}\)
satisfies
\[
(g\circ\iota_{\ell,I}\circ j_{\ell,I})
\circ q_{\ell,I}
=
g\circ R(Q_{\ell,I}).
\]
Thus quotient transport preserves exactly the
complete functions attainable through the
original frontier.

Equal whole-state distinction relations at two
frontiers imply equal quotient information, but
do not imply equal continuation availability.
Continuation equivalence additionally requires
the transported continuation families to agree
under the comparison map, as characterized by
Theorem~\ref{thm:continuation-transport}.

For a target
\[
f:R(X_\ell)\to(W,d_W),
\]
define the frontier target variation
\begin{equation}
\label{eq:frontier-target-variation}
\Delta_{\ell,I}(f)
:=
\Delta_{E^{\mathrm{whole}}_{\ell,I}}(f).
\end{equation}
Every \(G\in\overline{\mathcal G}_{\ell,I}\)
obeys the deterministic barrier
\[
\sup_x d_W\bigl(f(x),Gq_{\ell,I}(x)\bigr)
\ge
\frac12\Delta_{\ell,I}(f).
\]

The same frontier also carries parameter-family data. Let \(\Theta_\ell\) be the complete trainable parameter space specified in Appendix~\ref{app:dimension-labelled-transformer}, including all LayerNorm gains and shifts, affine weights and biases, output biases, activation parameters when declared trainable, and all declared parameter-sharing identifications. For a frontier \(I\), let
\[
\Theta_{\ell,I}^-,
\qquad
\Theta_{\ell,I}^+,
\qquad
\Theta_{\ell,I}^{\mathrm{sh}}
\]
be the parameter-coordinate spaces occurring only in the completed region, only in the uncompleted region, and on both sides through an explicit sharing declaration. The global parameter space maps into the compatibility pullback
\begin{equation}
\label{eq:frontier-parameter-pullback}
\Theta_\ell
\longrightarrow
\Theta_{\ell,I}^-
\times_{\Theta_{\ell,I}^{\mathrm{sh}}}
\Theta_{\ell,I}^+.
\end{equation}
When the declared coordinate incidence is exhaustive this map is an isomorphism onto the compatible parameter locus. No assumption of parameter independence across the cut is required.

Fixing a producer-side parameter state \(\vartheta^-\) determines the compatible receiver-parameter fiber
\[
\Theta_{\ell,I}^+(\vartheta^-).
\]
The downstream parameterized receiver family is then
\[
\mathcal Q_{\ell,I}^{\vartheta^-}:
\operatorname{Disc}\bigl(\Theta_{\ell,I}^+(\vartheta^-)\bigr)
\longrightarrow
X_{\ell,I}/\mathcal A.
\]
For an upstream represented context \(A:\Omega\to X_{\ell,I}\), Part~\ref{part:represented-architecture} supplies the family image
\begin{equation}
\label{eq:frontier-family-image}
\mathcal C_A(\mathcal Q_{\ell,I}^{\vartheta^-})
=
\operatorname{im}
\bigl(
\mathsf D_\Omega A^*\mathcal Q_{\ell,I}^{\vartheta^-}
\bigr)
\end{equation}
and the common receiver-family kernel
\begin{equation}
\label{eq:frontier-family-envelope}
E_{A,\ell,I}^{\mathrm{fam}}
=
\bigwedge_{\vartheta^+}
\mathsf D_\Omega
\bigl(
S_{\ell,I,\vartheta^+}A
\bigr).
\end{equation}
These are different summaries: the image retains which receiver distinction relations occur; the meet retains only their common refinement.

Whenever the realized suffix is differentiable at the state under study, the same cut has the receiver pullback
\begin{equation}
\label{eq:frontier-cotangent-pullback}
\mathcal R_{\ell,I}^{\ast}
:=
D R(S_{\ell,I})^*.
\end{equation}
For a downstream loss covector \(\lambda_{\ell+1}\),
\[
\lambda_{\ell,I}
=
\mathcal R_{\ell,I}^{\ast}\lambda_{\ell+1}.
\]
No differentiability is required for the cut, distinction, quotient, target-variation, or parameter-family constructions themselves.

For each legal frontier \(I\), collect
the represented and realized quantities
defined above into the tuple
\begin{equation}
\label{eq:maximal-cut-signature}
\boxed{
\Sigma_{\ell,I}^{\mathrm{cut}}
=
\Bigl(
Q_{\ell,I},
S_{\ell,I},
E^{\mathrm{whole}}_{\ell,I},
q_{\ell,I},
\mathcal G_{\ell,I},
\Delta_{\ell,I},
\mathcal C_{\ell,I},
E_{\ell,I}^{\mathrm{fam}},
\mathcal R_{\ell,I}^{\ast}
\Bigr),
}
\end{equation}
with analytic entries retained only on the domains where their additional metric, probabilistic, or differentiable structure has been declared. Retaining these tuples over all legal
frontiers gives the frontier-indexed
data collection
\begin{equation}
\label{eq:maximal-block-ledger}
\boxed{
\Lambda_\ell^{\max}
:=
\bigl(
\Sigma_{\ell,I}^{\mathrm{cut}}
\bigr)_{I\in\mathfrak F_\ell}.
}
\end{equation}

\subsection{Branch-relative distinctions are distinct from whole-state distinctions}
\label{app:branch-vs-whole-fine}

The product frontier keeps all live state needed by the block. To analyze what a selected branch presentation itself forgets, project the frontier to the declared branch coordinates. If
\[
\pi^{\mathrm{br}}_{\ell,I}:X_{\ell,I}\to X^{\mathrm{br}}_{\ell,I}
\]
is the corresponding product projection, define
\begin{equation}
\label{eq:branch-prefix-kernel}
E^{\mathrm{br}}_{\ell,I}
:=
\ker R\bigl(
\pi^{\mathrm{br}}_{\ell,I}Q_{\ell,I}
\bigr).
\end{equation}
In general
\[
E^{\mathrm{whole}}_{\ell,I}
\subseteq
E^{\mathrm{br}}_{\ell,I},
\]
because forgetting live bypass coordinates can only identify additional predecessor states.

The fine block carries a whole-state information functor and a separately indexed branch-access profile:
\[
I_\ell^{\mathrm{whole}}:\mathfrak F_\ell\to\mathsf{Eq}(R(X_\ell)),
\qquad
I_{\ell,I}^{\mathrm{br}}:=E_{\ell,I}^{\mathrm{br}}
\quad(I\in\mathfrak F_\ell).
\]
The complete frontier maps form a compatible chain of distinction refinements. Branch projections at different frontiers need not commute with the frontier transition maps, so their indexed relations need not be monotone and do not define a functor on the entire frontier lattice. Along a subdiagram on which $\pi_J^{\mathrm{br}}T_{I:J}$ factors through $\pi_I^{\mathrm{br}}$ for each comparable pair, the restricted branch profile is functorial. This difference records precisely how branch accessibility can vary while the residual stream continues to carry information.

\subsection{Exact fine-cut analysis of LayerNorm}
\label{app:exact-layernorm-cut-analysis}

For one token, let
\[
P_0
=
I_d-\frac{1}{d}\mathbf 1\mathbf 1^\top
\]
be the centering projection and define
\[
\psi_\varepsilon(c)
=
\frac{c}{\sqrt{\|c\|^2/d+\varepsilon}},
\qquad
c\in\mathbf 1^\perp,
\qquad
\varepsilon>0.
\]
The branch LayerNorm map is
\[
N(x)
=
\gamma\odot\psi_\varepsilon(P_0x)+\beta.
\]

\begin{lemma}[Injectivity of stabilized radial normalization]
\label{lem:layernorm-radial-injective}
The map \(\psi_\varepsilon:\mathbf 1^\perp\to\mathbf 1^\perp\) is injective.
\end{lemma}

\begin{proof}
If \(\psi_\varepsilon(c)=0\), then \(c=0\). For nonzero \(c\), the normalized vector has the same direction as \(c\), while its norm is
\[
r\longmapsto
\frac{r}{\sqrt{r^2/d+\varepsilon}},
\qquad r\ge0,
\]
which is strictly increasing because \(\varepsilon>0\). Equality of normalized vectors therefore gives equality of direction and norm, hence equality of the centered vectors.
\end{proof}

Let
\[
J_\gamma
=
\{j:\gamma_j\ne0\}
\]
and let \(\Pi_{J_\gamma}\) be coordinate projection onto that support.

\begin{proposition}[Exact LayerNorm distinction relation]
\label{prop:exact-layernorm-kernel}
For fixed \((\gamma,\beta,\varepsilon)\),
\begin{equation}
\label{eq:layernorm-kernel-general}
\boxed{
N(x)=N(x')
\iff
\Pi_{J_\gamma}
\psi_\varepsilon(P_0x)
=
\Pi_{J_\gamma}
\psi_\varepsilon(P_0x').
}
\end{equation}
The shift \(\beta\) does not affect this fixed-parameter distinction relation. If every coordinate of \(\gamma\) is nonzero, then
\begin{equation}
\label{eq:layernorm-full-support-kernel}
\boxed{
N(x)=N(x')
\iff
x'-x\in\operatorname{span}\{\mathbf1\}.
}
\end{equation}
For a token set \(S\), the full LayerNorm relation is the product, equivalently the intersection, of the tokenwise relations.
\end{proposition}

The intermediate LayerNorm cuts from Appendix~\ref{app:layernorm-cut-factorization} expose further represented structure even where their final branch distinction relation agrees. Mean extraction, centering, variance formation, inverse-scale formation, and gain/shift are therefore separate cut roles despite the final exact quotient described above.

\subsection{Exact fine-cut analysis of query, key, and value production}
\label{app:exact-qkv-cut-analysis}

Fix one block \(\ell\) and parameter state. For one token difference \(\delta x=x-x'\), the affine maps satisfy
\[
Q_{\ell,a}(x)=Q_{\ell,a}(x')
\iff
W^Q_{\ell,a}\delta x=0,
\]
with identical formulas for \(K\) and \(V\). Their affine biases cancel in the equality test at this cut.

For the joint head family, define the linear map
\[
\mathcal W_\ell^{QKV}
:
\R^d
\longrightarrow
\prod_{a=1}^{h_\ell}
\bigl(
\R^{q_\ell}\times\R^{q_\ell}\times\R^{v_\ell}
\bigr)
\]
by
\[
\mathcal W_\ell^{QKV}\delta x
=
\bigl(
W^Q_{\ell,a}\delta x,
W^K_{\ell,a}\delta x,
W^V_{\ell,a}\delta x
\bigr)_a.
\]
Then the exact tokenwise joint relation is
\begin{equation}
\label{eq:qkv-joint-kernel}
\boxed{
\ker\mathcal W_\ell^{QKV}
=
\bigcap_{a=1}^{h_\ell}
\left(
\ker W^Q_{\ell,a}
\cap
\ker W^K_{\ell,a}
\cap
\ker W^V_{\ell,a}
\right).
}
\end{equation}
Across tokens the joint relation is the product of these tokenwise relations. Thus the product-to-meet theorem of Part~\ref{part:represented-architecture} becomes literal linear-algebraic intersection at the Q/K/V frontier.

The affine biases are distinction-neutral at their own affine output cuts, but they are not dynamically or globally irrelevant. Query/key biases enter subsequent score values; value biases enter source-resolved values; hidden feed-forward biases alter which region of an arbitrary nonlinearity is visited. Fixed-map kernel neutrality at an affine cut therefore does not imply neutrality after later nonlinear composition or across a parameter family.

\subsection{Exact fine-cut analysis of scores, softmax, and weighted values}
\label{app:exact-score-softmax-analysis}

For one head and one receiver token, write the admissible source set as \(\mathcal M_{\ell,a,i}\). The score contraction maps \((q_i,(k_j)_j)\) to
\[
\left(
\frac{q_i^\top k_j}{\tau_{\ell,a}}
\right)_{j\in\mathcal M_{\ell,a,i}}.
\]
Since \(\tau_{\ell,a}>0\) is fixed at the cut, it does not alter the score distinction relation. Two query/key states are score-equivalent exactly when all admissible pairings agree:
\begin{equation}
\label{eq:score-kernel-exact}
\boxed{
q_i^\top k_j
=
q_i'^\top k_j'
\quad
\forall j\in\mathcal M_{\ell,a,i}.
}
\end{equation}
When the value state is carried, score-frontier equivalence additionally requires equality of the displayed value coordinates.

The parameter-space transformation
\[
W^Q\mapsto AW^Q,
\quad
b^Q\mapsto Ab^Q,
\quad
W^K\mapsto A^{-\top}W^K,
\quad
b^K\mapsto A^{-\top}b^K,
\qquad
A\in GL(q_\ell),
\]
is an exact represented re-presentation symmetry of the score realization. It is a parameter-space gauge freedom, not the state-space kernel pair in Equation~\eqref{eq:score-kernel-exact}. These two equivalence notions remain separate.

\begin{lemma}[Exact softmax quotient]
\label{lem:exact-softmax-quotient}
For \(s,s'\in\R^m\),
\[
\operatorname{softmax}(s)
=
\operatorname{softmax}(s')
\]
if and only if
\[
\boxed{
s'=s+c\mathbf1
}
\]
for some \(c\in\R\).
\end{lemma}

\begin{proof}
Equality of softmax coordinates implies equality of all probability ratios. Hence
\[
e^{s_j-s_k}=e^{s_j'-s_k'}
\]
for every \(j,k\), so \(s_j'-s_j\) is independent of \(j\). The converse is the usual additive invariance of softmax.
\end{proof}

Therefore each admissible receiver/head logit vector is quotiented by the one-dimensional subspace \(\operatorname{span}\{\mathbf1\}\). With masking represented by restricting the softmax domain, no formal infinite coordinates are required.

The next contraction
\[
(\alpha_{ij},v_j)_j
\longmapsto
(c_{ij})_j,
\qquad
c_{ij}=\alpha_{ij}v_j,
\]
has the exact relation
\begin{equation}
\label{eq:weighted-value-kernel}
\boxed{
(\alpha,v)\sim(\alpha',v')
\iff
\alpha_{ij}v_j
=
\alpha'_{ij}v_j'
\quad
\text{for every admissible }i,j.
}
\end{equation}
No division by \(\alpha\), nonzero-value assumption, or rank assumption is required.

\subsection{Exact fine-cut analysis of source and head aggregation}
\label{app:exact-attention-aggregation-analysis}

Source aggregation
\[
(c_{aij})_j
\longmapsto
z_{ai}=\sum_jc_{aij}
\]
has translation space
\begin{equation}
\label{eq:source-sum-kernel}
\boxed{
\left\{
(\delta c_{aij})_j:
\sum_j\delta c_{aij}=0
\right\}.
}
\end{equation}
Thus two source-resolved states are equivalent at the head state exactly when their source contributions differ by a zero-sum family.

Headwise output projection gives
\[
p_{ai}=O_{\ell,a}z_{ai},
\]
so
\begin{equation}
\label{eq:output-projection-kernel}
\boxed{
z_{ai}\sim z'_{ai}
\iff
z_{ai}-z'_{ai}\in\ker O_{\ell,a}.
}
\end{equation}
The joint head relation is the product of these relations.

Head aggregation with fixed output bias is
\[
t_i=b^O_\ell+\sum_ap_{ai}.
\]
The bias cancels in the fixed-map equality test, and
\begin{equation}
\label{eq:head-sum-kernel}
\boxed{
(p_{ai})_a\sim(p'_{ai})_a
\iff
\sum_a(p_{ai}-p'_{ai})=0.
}
\end{equation}
Thus source identity and head identity are forgotten at different represented cuts.

The first residual addition is
\[
(H_\ell,T_\ell)
\longmapsto
U_\ell=H_\ell+T_\ell.
\]
Its exact pair-state relation is
\begin{equation}
\label{eq:first-residual-pair-kernel}
\boxed{
(H,T)\sim(H',T')
\iff
H+T=H'+T'.
}
\end{equation}
Unlike the preceding branch contractions, this operation removes the bypass/branch decomposition itself and can therefore enlarge the whole-state distinction relation.

\subsection{Exact fine-cut analysis of the arbitrary-activation feed-forward branch}
\label{app:exact-ffn-cut-analysis}

Set \(z_{\ell,0}=\widehat U_\ell\). For \(1\le r\le n_\ell\), define
\[
G_{\ell,r}(z)
=
W_{\ell,r}z+b_{\ell,r},
\qquad
Z_{\ell,r}(z)
=
\phi_{\ell,r}(G_{\ell,r}(z)).
\]
For the first stage \(W_{\ell,1}\in\R^{w\times d}\); for later hidden stages \(W_{\ell,r}\in\R^{w\times w}\). At the affine preactivation cut,
\begin{equation}
\label{eq:ffn-affine-kernel}
\boxed{
G_{\ell,r}(z)=G_{\ell,r}(z')
\iff
z-z'\in\ker W_{\ell,r}.
}
\end{equation}
At the activation cut the exact local relation is simply the kernel pair of the declared activation:
\begin{equation}
\label{eq:arbitrary-activation-kernel}
\boxed{
G_{\ell,r}(z)
\sim_{\phi_{\ell,r}}
G_{\ell,r}(z')
\iff
\phi_{\ell,r}(G_{\ell,r}(z))
=
\phi_{\ell,r}(G_{\ell,r}(z')).
}
\end{equation}
No injectivity, smoothness, coordinatewise action, or monotonicity is required.

Let \(F_{\ell,r}^{\mathrm{pre}}\) and \(F_{\ell,r}^{\mathrm{act}}\) denote the branch prefixes from \(\widehat U_\ell\) to the \(r\)-th preactivation and activation. Their distinction relations obey the exact monotone chain
\[
\ker F_{\ell,1}^{\mathrm{pre}}
\subseteq
\ker F_{\ell,1}^{\mathrm{act}}
\subseteq
\cdots
\subseteq
\ker F_{\ell,n_\ell}^{\mathrm{pre}}
\subseteq
\ker F_{\ell,n_\ell}^{\mathrm{act}}
\subseteq
\ker F_\ell,
\]
where \(F_\ell\) is the final affine feed-forward branch output. Hidden biases do not change the kernel of an individual affine map, but they can change the later activation kernel because they translate the preactivation before the arbitrary map \(\phi_{\ell,r}\) is applied.

The final affine output
\[
F_\ell(z)
=
W_{\ell,n_\ell+1}z+b_{\ell,n_\ell+1}
\]
has fixed-map equality relation determined by \(\ker W_{\ell,n_\ell+1}\). The final output bias cancels in that equality test. The second residual addition
\[
(U_\ell,F_\ell)
\longmapsto
H_{\ell+1}=U_\ell+F_\ell
\]
has the exact pair-state relation
\begin{equation}
\label{eq:second-residual-pair-kernel}
\boxed{
(U,F)\sim(U',F')
\iff
U+F=U'+F'.
}
\end{equation}

\subsection{Whole-state distinction plateaus inside one block}
\label{app:whole-state-plateaus}

Define the two residual-stage maps
\[
R_{\ell,1}:X_\ell\to X_\ell,
\qquad
R_{\ell,1}(H)=H+T_\ell(H),
\]
and
\[
R_{\ell,2}:X_\ell\to X_\ell,
\qquad
R_{\ell,2}(U)=U+F_\ell(U).
\]
Then
\[
B_\ell=R_{\ell,2}R_{\ell,1}.
\]

Partition the frontier lattice into the frontiers strictly before the first residual addition, the interval after the first residual addition and strictly before the second, and the terminal frontier after the second residual addition.

The one-block whole-state plateau is stated as Theorem~\ref{thm:one-block-whole-state-plateaus} in the main text.

\begin{proof}[Proof of Theorem~\ref{thm:one-block-whole-state-plateaus}]
Every frontier before the first merge contains the original residual state \(H_\ell\) as a displayed product factor, so its prefix has the form
\[
H\longmapsto(H,\eta(H))
\]
for some represented branch state \(\eta\); this map is injective. After the first residual merge, every frontier before the second contains \(U_\ell=R_{\ell,1}(H)\) as a displayed factor and has the form
\[
H\longmapsto
\bigl(R_{\ell,1}(H),\zeta(R_{\ell,1}(H))\bigr).
\]
Its kernel is therefore exactly \(\ker R(R_{\ell,1})\). The terminal prefix is \(B_\ell\) itself.
\end{proof}

The exact semantic quotient chain is consequently
\begin{equation}
\label{eq:maximal-one-block-quotient-chain}
\boxed{
X_\ell
\twoheadrightarrow
X_\ell/\ker R(R_{\ell,1})
\twoheadrightarrow
X_\ell/\ker R(B_\ell),
}
\end{equation}
with every fine carried frontier inside either plateau semantically isomorphic at the quotient grain while remaining distinct in \(\mathcal A\).

For any target \(f\), the whole-state target variation therefore has only two possibly nonzero increments:
\begin{equation}
\label{eq:one-block-target-telescope}
\boxed{
\Delta_{\ker B_\ell}(f)
-
\Delta_{\Delta}(f)
=
\Bigl[
\Delta_{\ker R_{\ell,1}}(f)-\Delta_{\Delta}(f)
\Bigr]
+
\Bigl[
\Delta_{\ker B_\ell}(f)-\Delta_{\ker R_{\ell,1}}(f)
\Bigr].
}
\end{equation}
Branch-relative target variation can increase at every fine contraction listed above.

\subsection{The exact parameter family survives cut contraction}
\label{app:parameter-family-survives-contraction}

Let \(\Theta_\ell\) be the complete block parameter space. Pointwise total composition of the fully resolved represented block gives a parameterized coarse block family
\begin{equation}
\label{eq:coarse-block-parameter-family}
\boxed{
\mathcal B_\ell:
\operatorname{Disc}(\Theta_\ell)
\longrightarrow
X_\ell/\mathcal A,
\qquad
\theta_\ell\longmapsto B_{\ell,\theta_\ell}:X_\ell\to X_\ell.
}
\end{equation}
Composition removes internal carriers from the exposed process but does not remove the parameter index that generated the composite arrow. Thus the exact fine parameter coordinates---all Q/K/V weights and biases, output blocks and bias, LayerNorm gains and shifts, feed-forward weights and biases, and activation parameters---remain coordinates of the coarse block family.

The dimension-labelled architecture coordinate \(\kappa_\ell\), the exact block count \(P_\ell\), the scale--shape coordinates of Appendix~\ref{app:scale-shape-coordinates}, and the architecture dependency matroid of Appendix~\ref{app:architecture-matroid} therefore survive as matched data of the block family. They are not reconstructible from the scalar block count alone.

\subsection{Matched effective block and the fine-to-coarse ledger}
\label{app:matched-effective-block}

After all internal frontiers have been analyzed, contract the represented block path to its single composite arrow. The exposed coarse cut system is then only the initial and terminal factorization objects
\[
(\operatorname{id}_{X_\ell},B_\ell),
\qquad
(B_\ell,\operatorname{id}_{X_\ell})
\]
of \(\operatorname{Fact}_{\mathcal A}(B_\ell)\). The fine carriers no longer appear as dynamical interfaces of the coarse process.

Before performing that contraction, retain the quantities already derived from them in the \emph{matching ledger}
\begin{equation}
\label{eq:block-matching-ledger}
\boxed{
\Lambda_\ell
=
\Bigl(
\Lambda_\ell^{\max},
\Theta_\ell,
\kappa_\ell,
P_\ell,
\rho_\ell,
M_{\mathrm{arch},\ell},
P_{\mathrm{inn},\ell}
\Bigr),
}
\end{equation}
where \(\Lambda_\ell^{\max}\) is the frontier ledger in Equation~\eqref{eq:maximal-block-ledger}, \(\rho_\ell\) denotes the exact dimension-ratio coordinates, \(M_{\mathrm{arch},\ell}\) is the architecture-coordinate matroid, and \(P_{\mathrm{inn},\ell}\) is retained when a same-function refinement has supplied the cut-subspace polymatroid of Appendix~\ref{app:innovation-matroid}.

\begin{definition}[Matched effective Transformer block]
\label{def:matched-effective-block}
The maximally matched coarse block is
\begin{equation}
\label{eq:matched-effective-block}
\boxed{
\mathbb B_\ell^{\mathrm{eff}}
:=
\bigl(
X_\ell,
\mathcal B_\ell,
\Lambda_\ell
\bigr).
}
\end{equation}
The boundary-to-boundary family
\(\mathcal B_\ell\) specifies the
coarse represented block, while
\(\Lambda_\ell\) retains the
fine-cut data derived before the
internal interfaces were contracted.

These data include the selected
producer and continuation maps,
kernel relations, restricted
receiver families, and, where
declared, the differential and
architectural information associated
with those cuts.

The construction retains the
maximal collection specified above.
It does not assert that every
scientific observable requires
every retained entry.
\end{definition}

For a selected observable
\(\mathcal O\), one may seek a
smaller retained-data object
\[
\Lambda_\ell
\twoheadrightarrow
\Lambda_\ell^{(\mathcal O)}.
\]
This reduction is exact for
\(\mathcal O\) when the observable
factors through the reduced data.
For example, a complete-state
kernel calculation need not retain
the gradient covectors, while a
local gradient calculation requires
the corresponding forward features
and backward covectors.

\begin{proposition}[
Criterion for observable-exact block contraction
]
\label{thm:controlled-block-contraction}

Let \(\mathcal O\) be a scientific
observable of the maximally resolved
block system \(\mathscr G_\ell\).

If there exists a function
\(\overline{\mathcal O}\) such that
\[
\mathcal O(\mathscr G_\ell)
=
\overline{\mathcal O}
\bigl(
\mathbb B_\ell^{\mathrm{eff}}
\bigr),
\]
then the matched effective block
retains sufficient information
to calculate \(\mathcal O\).

More generally, if the observable
factors through
\(\Lambda_\ell^{(\mathcal O)}\),
then this smaller retained-data
object is sufficient for the
same calculation.
\end{proposition}

This is the block-level instance of controlled forgetting: internal carriers are removed only after their effect has been transferred to the retained block variables. Beginning directly from \(B_\ell\) would not supply \(\Lambda_\ell\), because composition and nonfaithful realization do not reconstruct the contracted factorization.

\begin{proposition}[
Fine structure is not determined by the
realized block map and parameter count
]
\label{prop:fine-structure-not-recoverable}

The realized input--output map of a block
at a specified parameter state, together
with its scalar parameter count, does not
determine its maximally retained
fine-cut data.

In particular, these two quantities
do not determine the internal value
carrier dimension, feed-forward hidden
carrier dimension, or the complete
family of represented internal cuts.
\end{proposition}

\begin{proof}
Consider two untied PreNorm block
architectures with
\[
d=2,\qquad h=q=n=1,
\]
and no trainable activation parameters.
Choose
\[
\kappa_1:\ (v,w)=(1,2),
\qquad
\kappa_2:\ (v,w)=(2,1).
\]

By Equation~\eqref{eq:exact-block-total-count},
both architectures have scalar parameter
count
\[
P_{\ell,1}=P_{\ell,2}=33.
\]

Choose parameter states in which the
attention and feed-forward output
projections, together with their
output biases, vanish.

Let \(B_{\ell,1}\) and \(B_{\ell,2}\)
denote the corresponding represented
block composites. Since both residual
branches then contribute zero,
\[
R(B_{\ell,1})
=
R(B_{\ell,2})
=
\operatorname{id}_{R(X_\ell)}.
\]

Nevertheless, the value carriers have
dimensions \(1\) and \(2\), respectively,
while the feed-forward hidden carriers
have dimensions \(2\) and \(1\).
The represented internal cuts and
their retained data therefore differ.

Thus identical realized block maps
and scalar parameter counts do not
determine the maximally retained
fine-cut data.
\end{proof}

\subsection{Retained frontier data and learning gradients}
\label{app:ledger-to-trained-refinement}

The block-level input--output map
does not reconstruct the internal
features and continuations needed
for local gradient calculations.

The frontier data retain the
producer maps \(Q_{\ell,I}\),
the receiver maps \(S_{\ell,I}\),
and the relevant parameter
coordinates. At a differentiable
cut, the producer determines
the features entering a local
operation, while the receiver
determines the pullback of the
downstream loss covector.
These two quantities determine
the local parameter gradient.

When differentiability is available at a retained innovation site, a local parameter perturbation with producer feature \(\phi_{\ell,I}\) and output-side parameter \(W\) has first-order gradient
\begin{equation}
\label{eq:ledger-local-gradient}
\boxed{
\widehat G_{\ell,I}
=
\lambda_{\ell,I}
\phi_{\ell,I}^{\top},
\qquad
\lambda_{\ell,I}
=
D R(S_{\ell,I})^*\lambda_{\ell+1}.
}
\end{equation}
The complete block map alone
does not retain the separate
producer feature
\(\phi_{\ell,I}\)
and downstream covector
\(\lambda_{\ell,I}\)
appearing in this local gradient.
The retained frontier data
make both factors available
at the selected cut.

\subsection{Block segments and the general insertion theorem}
\label{app:block-segment-insertion}

For integers \(a\le b\), define the represented block segment
\begin{equation}
\label{eq:block-segment-composite}
\mathcal B_{a:b}
:=
B_bB_{b-1}\cdots B_a
:
X_a\to X_{b+1}.
\end{equation}
For a block \(j\in\{a,\ldots,b\}\), define the preceding segment
\[
A_{a:j}
:=
B_{j-1}\cdots B_a,
\qquad
A_{a:a}=\operatorname{id}_{X_a},
\]
and the following segment
\[
S_{j:b}
:=
B_b\cdots B_{j+1},
\qquad
S_{b:b}=\operatorname{id}_{X_{b+1}}.
\]

\begin{theorem}[Multi-block cut insertion]
\label{thm:multiblock-cut-insertion}
For every internal frontier factorization
\[
B_j
=
S_{j,I}Q_{j,I},
\]
the corresponding cut of the segment \(\mathcal B_{a:b}\) is
\begin{equation}
\label{eq:multiblock-cut-frame}
\boxed{
C_{j,I}^{[a,b]}
=
\left(
Q_{j,I}A_{a:j},
\;
S_{j:b}S_{j,I}
\right)
\in
\operatorname{Fact}_{\mathcal A}(\mathcal B_{a:b}).
}
\end{equation}
For fixed \(j\), this assignment is a functor
\[
J_j^{[a,b]}:
\operatorname{Fact}_{\mathcal A}(B_j)
\longrightarrow
\operatorname{Fact}_{\mathcal A}(\mathcal B_{a:b}).
\]
\end{theorem}

\begin{proof}
Associativity gives
\[
(S_{j:b}S_{j,I})(Q_{j,I}A_{a:j})
=
S_{j:b}B_jA_{a:j}
=
\mathcal B_{a:b}.
\]
A factorization morphism \(T\) satisfying \(TQ=Q'\) and \(S=S'T\) remains a factorization morphism after precomposition by \(A_{a:j}\) and postcomposition by \(S_{j:b}\). Identity and composition are preserved.
\end{proof}

This theorem is the exact multi-block form of cut relativity. The local internal arrow has not changed. Its producer prefix and receiver suffix have.

\subsection{Producer pullback and receiver extension are the two multi-block actions}
\label{app:producer-receiver-multiblock-actions}

Let
\[
E_{j,I}^{\mathrm{loc}}
:=
\ker R(Q_{j,I})
\]
be the distinction relation of the local block prefix at frontier \(I\). Naturality of the Part~1 distinction relation gives
\begin{equation}
\label{eq:multiblock-distinction-pullback}
\boxed{
E_{j,I}^{[a,b]}
=
R(A_{a:j})^\sharp
E_{j,I}^{\mathrm{loc}}.
}
\end{equation}
Preceding blocks pull the selected
frontier's equivalence relation
back to the input of the complete
segment. Following blocks do not
change which segment-input states
the producer prefix has already
identified; they change the
available receiver continuation.

The actual segment receiver after that same frontier is
\begin{equation}
\label{eq:multiblock-receiver}
\boxed{
S_{j,I}^{[a,b]}
=
S_{j:b}S_{j,I}.
}
\end{equation}
Whenever differentiability is available, the receiver covector therefore obeys
\begin{equation}
\label{eq:multiblock-covector}
\boxed{
\lambda_{j,I}^{[a,b]}
=
D R(S_{j,I})^*
D R(B_{j+1})^*
\cdots
D R(B_b)^*
\lambda_{b+1}.
}
\end{equation}
The two sides of cut relativity are therefore mathematically different operations:
\[
\boxed{
\begin{array}{ccl}
\text{upstream blocks}
&:&
\text{pull back the distinction relation and change the producer state},\\[2mm]
\text{downstream blocks}
&:&
\text{postcompose the receiver and pull back the loss covector}.
\end{array}
}
\]

At a differentiable output-gated site, let \(\phi_{j,I}\) be its producer feature. Inside the segment,
\begin{equation}
\label{eq:multiblock-gradient-relativity}
\boxed{
\widehat G_{j,I}^{[a,b]}
=
\lambda_{j,I}^{[a,b]}
\phi_{j,I}\bigl(A_{a:j}(H_a)\bigr)^\top.
}
\end{equation}
The same local parameterized computation can therefore have a different first-order trained role because its producer state, receiver covector, or both have changed.

\subsection{Contextual receiver families under preceding blocks}
\label{app:multiblock-contextual-families}

Fix a frontier \(I\) inside block \(j\) and its compatible downstream parameter family \(\mathcal Q_{j,I}\). The preceding block segment \(A_{a:j}\) is an upstream context in the producer fiber. Its contextual family profile is
\begin{equation}
\label{eq:multiblock-family-profile}
\boxed{
\mathcal C_{A_{a:j}}(\mathcal Q_{j,I})
=
\left\{
R(A_{a:j})^\sharp E_\theta:
E_\theta\in
\mathcal C_{\operatorname{id}}(\mathcal Q_{j,I})
\right\}.
}
\end{equation}
The corresponding common kernel is
\[
E_{A_{a:j},\mathcal Q_{j,I}}
=
R(A_{a:j})^\sharp
E_{\operatorname{id},\mathcal Q_{j,I}}
\]
whenever the relevant pullback preserves the declared meet, as it does for intersection in Set.

Two consequences from Part~\ref{part:represented-architecture} apply immediately.
\begin{enumerate}
    \item If \(R(A_{a:j})\) is surjective, pullback is order-reflecting, so distinct ambient distinction relations in the receiver family cannot merge merely because of that context.
    \item If \(R(A_{a:j})\) is injective, every segment-input distinction survives, but pullback need not be injective on ambient equivalence relations. Distinct receiver relations can therefore become equal on the reachable image.
\end{enumerate}
Hence composition with an information-preserving preceding block can change the effective internal architecture profile of a later block without any upstream information loss. The injective-context phenomenon of Theorem~\ref{thm:injective-attention-nonconstant} is therefore not restricted to a toy attention witness; it is the generic contextual mechanism by which earlier Transformer blocks can alter the effective semantics of later internal receiver families.

\subsection{Downstream composition can change role without changing information}
\label{app:downstream-role-without-information-loss}

Suppose two receiver presentations of the same frontier are related by a semantic isomorphism
\[
T:Z\xrightarrow{\cong}Z',
\qquad
Q'=TQ.
\]
Their quotient information is the same. After a downstream block segment is attached, architectural equivalence still requires the actual continuation family to transport under \(T\). By Theorem~\ref{thm:continuation-transport},
\[
\mathfrak B(Q,\mathcal G)
=
\mathfrak B(Q',\mathcal G')
\iff
\mathcal G=T^*\mathcal G'.
\]
Thus a downstream Transformer segment can change the computational role of a semantically recoverable internal state simply because the represented suffix is not closed under the required re-presentation. This is the receiver-side counterpart of the producer-context effect above.

\subsection{Two-block cut relativity}
\label{app:two-block-cut-relativity}

Consider
\[
X_\ell
\xrightarrow{B_\ell}
X_{\ell+1}
\xrightarrow{B_{\ell+1}}
X_{\ell+2}.
\]
For a frontier \(I\) inside the first block,
\begin{equation}
\label{eq:two-block-first-cut}
\boxed{
C_{\ell,I}^{[\ell,\ell+1]}
=
\bigl(
Q_{\ell,I},
B_{\ell+1}S_{\ell,I}
\bigr).
}
\end{equation}
Relative to the isolated block,
the producer prefix is unchanged
and the receiver continuation is
extended by the second block.
The equivalence relation induced
on the two-block segment input
is therefore unchanged at this
cut, while the downstream
loss covector becomes
\[
\lambda_{\ell,I}^{(2)}
=
D R(S_{\ell,I})^*
D R(B_{\ell+1})^*
\lambda_{\ell+2}.
\]

For a frontier \(J\) inside the second block,
\begin{equation}
\label{eq:two-block-second-cut}
\boxed{
C_{\ell+1,J}^{[\ell,\ell+1]}
=
\bigl(
Q_{\ell+1,J}B_\ell,
S_{\ell+1,J}
\bigr).
}
\end{equation}
Relative to the isolated second block, the receiver suffix is unchanged and the producer context is changed by \(B_\ell\). Its segment-input distinction relation is
\[
\boxed{
E_{\ell+1,J}^{(2)}
=
R(B_\ell)^\sharp
E_{\ell+1,J}^{\mathrm{loc}}.
}
\]

Thus a two-block segment already separates the two directions of DLE:
\[
\boxed{
\begin{array}{c|c|c}
\text{cut location}
&\text{producer relative to isolated block}
&\text{receiver relative to isolated block}\\
\hline
\text{block }\ell
&\text{unchanged}
&\text{extended}\\
\text{block }\ell+1
&\text{extended}
&\text{unchanged}
\end{array}}
\]

\subsection{Exact two-block whole-state information chain}
\label{app:two-block-information-chain}

Write
\[
R_{j,1}:H_j\mapsto U_j,
\qquad
R_{j,2}:U_j\mapsto H_{j+1},
\qquad
B_j=R_{j,2}R_{j,1}.
\]
For the two-block segment beginning at \(H_\ell\), define
\begin{align*}
E_0&:=\Delta_{X_\ell},\\
E_1&:=\ker R(R_{\ell,1}),\\
E_2&:=\ker R(B_\ell),\\
E_3&:=\ker R(R_{\ell+1,1}B_\ell),\\
E_4&:=\ker R(B_{\ell+1}B_\ell).
\end{align*}
Then
\begin{equation}
\label{eq:two-block-kernel-chain}
\boxed{
E_0
\subseteq
E_1
\subseteq
E_2
\subseteq
E_3
\subseteq
E_4.
}
\end{equation}
Every fine carried frontier of the two-block computation has exactly one of these five whole-state distinction relations. There can therefore be at most four strict whole-state information-loss events: the four residual additions.

For a target \(f\), the total target-variation increase telescopes exactly as
\begin{equation}
\label{eq:two-block-target-telescope}
\boxed{
\Delta_{E_4}(f)-\Delta_{E_0}(f)
=
\sum_{r=0}^{3}
\bigl[
\Delta_{E_{r+1}}(f)-\Delta_{E_r}(f)
\bigr].
}
\end{equation}
Each summand is nonnegative.
Fine branch-relative target variation
may change between these events
even while the complete-state
kernel relation remains constant
along a residual plateau.

\subsection{Three-block cut relativity and simultaneous two-sided change}
\label{app:three-block-cut-relativity}

Consider
\[
X_\ell
\xrightarrow{B_\ell}
X_{\ell+1}
\xrightarrow{B_{\ell+1}}
X_{\ell+2}
\xrightarrow{B_{\ell+2}}
X_{\ell+3}.
\]
For a cut \(I\) inside the first block,
\begin{equation}
\label{eq:three-block-first-cut}
\boxed{
C_{\ell,I}^{(3)}
=
\bigl(
Q_{\ell,I},
B_{\ell+2}B_{\ell+1}S_{\ell,I}
\bigr).
}
\end{equation}
For a cut \(J\) inside the middle block,
\begin{equation}
\label{eq:three-block-middle-cut}
\boxed{
C_{\ell+1,J}^{(3)}
=
\bigl(
Q_{\ell+1,J}B_\ell,
B_{\ell+2}S_{\ell+1,J}
\bigr).
}
\end{equation}
For a cut \(K\) inside the final block,
\begin{equation}
\label{eq:three-block-final-cut}
\boxed{
C_{\ell+2,K}^{(3)}
=
\bigl(
Q_{\ell+2,K}B_{\ell+1}B_\ell,
S_{\ell+2,K}
\bigr).
}
\end{equation}

Relative to each block analyzed in isolation, the first block acquires only additional receiver context, the last acquires only additional producer context, and the middle block acquires both. Thus three consecutive blocks are the shortest block segment in which every internal cut of one block can simultaneously have a changed producer history and a changed receiver continuation while its local represented computation is held fixed.

At a differentiable output-gated cut \(J\) inside the middle block,
\begin{equation}
\label{eq:three-block-middle-gradient}
\boxed{
\widehat G_{\ell+1,J}^{(3)}
=
\left[
D R(S_{\ell+1,J})^*
D R(B_{\ell+2})^*
\lambda_{\ell+3}
\right]
\phi_{\ell+1,J}
\bigl(B_\ell(H_\ell)\bigr)^\top.
}
\end{equation}
The producer feature and the receiver covector are both context-shifted relative to the isolated middle block. This is the direct training-level realization of two-sided cut relativity.

\subsection{Exact three-block whole-state information chain}
\label{app:three-block-information-chain}

Define
\begin{align*}
E_0&:=\Delta,\\
E_1&:=\ker R(R_{\ell,1}),\\
E_2&:=\ker R(B_\ell),\\
E_3&:=\ker R(R_{\ell+1,1}B_\ell),\\
E_4&:=\ker R(B_{\ell+1}B_\ell),\\
E_5&:=\ker R(R_{\ell+2,1}B_{\ell+1}B_\ell),\\
E_6&:=\ker R(B_{\ell+2}B_{\ell+1}B_\ell).
\end{align*}
Then
\begin{equation}
\label{eq:three-block-kernel-chain}
\boxed{
E_0\subseteq E_1\subseteq E_2\subseteq E_3\subseteq E_4\subseteq E_5\subseteq E_6.
}
\end{equation}
Every fine carried frontier in the three-block segment belongs to one of these seven whole-state information plateaus. At most six strict whole-state distinction-growth events occur, one at each residual addition.

For every target \(f\),
\begin{equation}
\label{eq:three-block-target-telescope}
\boxed{
\Delta_{E_6}(f)-\Delta_{E_0}(f)
=
\sum_{r=0}^{5}
\bigl[
\Delta_{E_{r+1}}(f)-\Delta_{E_r}(f)
\bigr].
}
\end{equation}
This separates the exact information-level contribution of each residual merge from all fine branch transformations whose effects are instead recorded in the continuation, geometry, and receiver-weighted parts of the ledger.

\subsection{The arbitrary \texorpdfstring{$N$}{N}-block theorem}
\label{app:n-block-cut-relativity}

For an \(N\)-block segment
\[
\mathcal B_{a:b},
\qquad
N=b-a+1,
\]
Theorem~\ref{thm:multiblock-cut-insertion}
determines the producer and receiver
factorization at every internal
frontier of a finite block segment.
The corresponding kernel pairs,
continuation families, and gradient
pullbacks follow by composition
with the preceding and following
blocks.

The segment cut-relativity result is Theorem~\ref{thm:n-block-cut-relativity} in Section~\ref{sec:transformer-multiblock-context}. The complete information chain below supplies its fine-grain structural consequences.

Define the ordered residual-stage prefixes
\[
\mathcal R_0=\operatorname{id},
\]
then successively compose
\[
R_{a,1},R_{a,2},R_{a+1,1},R_{a+1,2},\ldots,R_{b,1},R_{b,2}.
\]
Let
\[
E_r=\ker R(\mathcal R_r),
\qquad
0\le r\le2N.
\]
Then
\begin{equation}
\label{eq:n-block-kernel-chain}
\boxed{
E_0\subseteq E_1\subseteq\cdots\subseteq E_{2N}.
}
\end{equation}
Every carried fine frontier lies on one of these \(2N+1\) whole-state information plateaus. Hence an \(N\)-block PreNorm segment has at most \(2N\) strict whole-state distinction-growth events, regardless of how many LayerNorm, Q/K/V, score, softmax, source, head, affine, or activation interfaces are retained internally.

The target variation obeys the exact telescoping law
\begin{equation}
\label{eq:n-block-target-telescope}
\boxed{
\Delta_{E_{2N}}(f)-\Delta_{E_0}(f)
=
\sum_{r=0}^{2N-1}
\left[
\Delta_{E_{r+1}}(f)-\Delta_{E_r}(f)
\right].
}
\end{equation}
Thus the complete-state information cost of an arbitrarily fine Transformer segment localizes to residual merges, while fine internal interfaces remain necessary to determine presentation, continuation availability, geometric deformation, parameter dependence, innovation structure, and training-response information.

\subsection{Composition of matched effective blocks}
\label{app:matched-block-composition}

The coarse block objects can themselves be composed without returning to an unstructured scalar description. Let
\[
\mathbb B_i^{\mathrm{eff}}
=(X_i,\mathcal B_i,\Lambda_i)
\]
and
\[
\mathbb B_{i+1}^{\mathrm{eff}}
=(X_{i+1},\mathcal B_{i+1},\Lambda_{i+1})
\]
be adjacent matched blocks with compatible residual carriers.

Let $\Theta_{\mathrm{sh}}$ be the declared space of coordinates shared by the two blocks, and let $r_i:\Theta_i\to\Theta_{\mathrm{sh}}$ and $r_{i+1}:\Theta_{i+1}\to\Theta_{\mathrm{sh}}$ read out their respective shared coordinates. Their joint parameter space consists of compatible assignments:
\begin{equation}
\label{eq:composite-parameter-space}
\boxed{\Theta_{i:i+1}
=\Theta_i\times_{\Theta_{\mathrm{sh}}}\Theta_{i+1}
=\{(\theta_i,\theta_{i+1}):r_i(\theta_i)=r_{i+1}(\theta_{i+1})\}.}
\end{equation}
In the absence of sharing, $\Theta_{\mathrm{sh}}$ is a one-point space and the joint space is $\Theta_i\times\Theta_{i+1}$. For independent shared-coordinate constraints of dimension $p_{\mathrm{sh}}$ on free parameter spaces of dimensions $p_i,p_{i+1}$, its dimension is $p_i+p_{i+1}-p_{\mathrm{sh}}$. A globally shared parameter is updated once: its gradient sums the contributions from all occurrences. For three or more blocks, the joint space consists of assignments agreeing on every declared overlap; its construction is independent of how adjacent blocks are grouped.

Define
\begin{equation}
\label{eq:effective-block-product}
\boxed{
\mathbb B_{i+1}^{\mathrm{eff}}
\odot
\mathbb B_i^{\mathrm{eff}}
:=
\left(
X_i,
\mathcal B_{i+1}\mathcal B_i,
\Lambda_{i:i+1}
\right),
}
\end{equation}
where \(\Lambda_{i:i+1}\) is obtained from the two ledgers by the exact transformation rules already derived:
\begin{enumerate}
    \item For every retained frontier data tuple in the earlier block, postcompose the receiver suffix by \(B_{i+1}\). The corresponding receiver cotangent pullback acquires the factor \(D R(B_{i+1})^*\).
    \item For every retained frontier data tuple in the later block, precompose the producer prefix by \(B_i\). Pull its predecessor equivalence relation back by \(R(B_i)^\sharp\).
    \item the boundary cut \((B_i,B_{i+1})\) is added explicitly;
    \item architecture scale--shape coordinates and dependency constraints are combined with the declared parameter-sharing identifications;
    \item innovation data, when retained, are transported through the corresponding producer and receiver maps before a selected predictive quotient is formed.
\end{enumerate}

\begin{theorem}[Associativity of matched block composition]
\label{thm:matched-block-associativity}
Whenever three adjacent matched blocks have compatible residual carriers and parameter-sharing declarations,
\[
\boxed{
(\mathbb B_3^{\mathrm{eff}}\odot\mathbb B_2^{\mathrm{eff}})
\odot\mathbb B_1^{\mathrm{eff}}
=
\mathbb B_3^{\mathrm{eff}}\odot
(\mathbb B_2^{\mathrm{eff}}\odot\mathbb B_1^{\mathrm{eff}})
}
\]
up to the canonical identification of the combined matching ledger.
\end{theorem}

\begin{proof}
The coarse represented arrows associate because composition in \(\mathcal A\) is associative. Producer-side distinction transport associates because
\[
R(BA)^\sharp
=
R(A)^\sharp R(B)^\sharp.
\]
Receiver-side continuation transport and cotangent pullback associate because ordinary composition and the chain rule are associative. The fiber-product parameter assignments agree under either bracketing because both specify the same tuples of local parameters satisfying every declared shared-coordinate equality. Therefore every component of the matched ledger agrees under the canonical rebracketing.
\end{proof}

This theorem means that one may coarse-grain a long Transformer segment block by block without making the result depend on whether blocks are grouped from the left, from the right, or hierarchically.

\subsection{Interpretation of the matched cut construction for training}
\label{app:matched-training-interpretation}

Parameter count is a boundary-level coordinate; it does not determine the typed internal operators or their gradients. The architecture-coordinate matroid records dependencies among dimension/count variables, while the subspace polymatroid of Appendix~\ref{app:innovation-matroid} records overlap among first-order predictor variations exposed at different cuts. Receiver weighting can change which variations affect a selected loss without changing their algebraic span. These are different reductions.

The block contraction in this appendix is exact when the retained ledger contains the interfaces needed by the chosen continuation. It should not be identified with the predictive quotient of the complete learning process: that quotient is generated by the training kernel and the selected observable in Section~\ref{sec:derived-predictive}. In particular a maximal forward cut ledger need not retain optimizer memory, and a minimal future-loss state need not retain every microscopic source/head cut. Matching is a procedure for carrying required effects through a declared contraction; its sufficiency is specific to the quantity being followed.

\section{Proofs for represented interfaces and contextual families}\label{app:categorical-proof-details}

\subsection*{Two-sided arrow organization}
\begin{proof}[Proof of Two-sided arrow organization]
An object of $\int\mathsf{Rec}$ is a pair $(X,F)$ with $F:X\to Y$, hence an object of $\mathcal A^\to$. A morphism over $A:X\to X'$ is a coslice morphism
\[
F\to A^*F'=F'A,
\]
equivalently a map $B:Y\to Y'$ satisfying
\[
BF=F'A.
\]
Thus the total morphisms are exactly commuting squares. The chosen cartesian lift of $A:\Omega\to X$ at $F:X\to Y$ is
\[
\begin{array}{ccc}
\Omega & \xrightarrow{FA} & Y\\
\downarrow A && \downarrow \Id_Y\\
X&\xrightarrow{F}&Y.
\end{array}
\]
Strict functoriality follows from associativity.

Dually, an object of $\int\mathsf{Prod}$ is again an arrow $F:X\to Y$. A morphism over $B:Y\to Y'$ is a slice morphism from $BF:X\to Y'$ to $F':X'\to Y'$, hence a map $A:X\to X'$ with
\[
F'A=BF.
\]
These are the same commuting squares. The displayed postcomposition squares give a split opcartesian cleavage for $\operatorname{cod}$.
\end{proof}

\subsection*{Natural distinction relation}
\begin{proof}[Proof of Natural distinction relation]
The receiver fibration reindexes $F$ to $FA$. Functoriality of realization gives
\[
R(FA)=R(F)R(A).
\]
Kernel pairs are stable under pullback: the kernel pair of the composite $R(F)R(A)$ is the inverse image along $R(A)\times R(A)$ of the kernel pair of $R(F)$. This is exactly the displayed equality. Naturality on receiver-fiber morphisms follows from
\[
F'=hF
\quad\Longrightarrow\quad
FA=(hF)A
\]
and monotonicity of kernel-pair inclusion under postcomposition.
\end{proof}

\subsection*{Squared-loss decomposition}
\begin{proof}[Proof of Squared-loss decomposition]
For every $g\in\mathcal G$,
\[
Y-g(Z_H)
=
(Y-m_Q)+(m_Q-g(Z_H)).
\]
The second term is $\sigma(Z_H)$-measurable and
\[
\mathbb E[Y-m_Q\mid\sigma(Z_H)]=0.
\]
The cross term therefore vanishes in expectation. Expanding the squared norm and taking the infimum gives the result.
\end{proof}

\subsection*{Approximate contextual barrier}
\begin{proof}[Proof of Approximate contextual barrier]
Take $\omega,\omega'$ in one $P_0$-fiber. Lipschitz continuity gives
\[
d\bigl(
G_\theta Q_\theta A(\omega),
G_\theta Q_\theta A(\omega')
\bigr)
\le\eta.
\]
The triangle inequality bounds the target separation by the two endpoint errors plus $\eta$. At least one endpoint error is therefore at least half of
\[
d(f(\omega),f(\omega'))-\eta.
\]
Take the positive part and the supremum over $E_{P_0}$.
\end{proof}

\subsection*{Architecture semantics transformation}
\begin{proof}[Proof of Architecture semantics transformation]
Fix $d$ and an aligned receiver $j$. Since
\[
A'=A\kappa,
\]
the reindexed family at $A'$ is obtained by further precomposition of the family at $A$ through $\kappa$. The naturality equation
\[
\mathsf D_{\Omega'}(Q A\kappa)
=
R(\kappa)^\sharp
\mathsf D_\Omega(QA)
\]
holds for every parameter-state arrow $Q$ by Theorem~\ref{thm:shadow-naturality}. Taking images over the parameter index gives
\[
\mathcal C_{A'}(\mathcal Q_{d,j})
=
R(\kappa)^\sharp
\mathcal C_A(\mathcal Q_{d,j}),
\]
with pullback applied elementwise. This is exactly the $j$th coordinate of the naturality equation for $\Sigma$. Apply the argument to every aligned receiver.
\end{proof}

\section{Causal optimizers and closed-loop Transformer training}
\label{app:causal-optimizer-training}

Training is modeled as a closed-loop
process between the represented
Transformer and an optimizer acting
through a declared observation
interface. The Transformer and
data law generate observations;
the optimizer converts observation
histories into parameter actions;
the implementation realizes those
actions as parameter updates.

Histories inducing the same
conditional laws for all future
optimizer actions define the
optimizer's behavioral-state
classes. Composing their action
laws with the parameter-update
implementation gives the
closed-loop training kernel.
The predictive information
required for a selected task
or contextual observable is
then derived from that kernel
in Section~\ref{sec:derived-predictive}.

\subsection{The declared training-observation interface}
\label{app:declared-training-interface-E}

Fix a depth \(L\) and an admissible dimension-labelled Transformer architecture coordinate \(c_L\). The coordinate records the residual width, typed heads, feed-forward branches, normalization and bias coordinates, and parameter-sharing identifications specified in Appendix~\ref{app:dimension-labelled-transformer}. Write \(\mathcal C_L\) for the collection of such architecture coordinates at depth \(L\), so \(c_L\in\mathcal C_L\).
Let
\[
\Theta_{L,c_L}
\]
be the exact parameter space
specified by the declared
architecture coordinate
\((L,c_L)\), including its
parameter-sharing identifications
and implementation conventions.

The optimizer acts on this
parameter space through a
declared observation and action
interface. Its state and update
law are defined independently
of which internal Transformer
frontiers are subsequently
retained for a particular
scientific observable.

Let
\[
\mathsf{Obs}_{L,c_L}
\]
be the measurable space of optimizer observations and let
\[
\mathsf{Act}_{L,c_L}(\theta)
\]
be the measurable action space at \(\theta\in\Theta_{L,c_L}\).  For ordinary additive optimization one may take
\[
\mathsf{Act}_{L,c_L}(\theta)=T_\theta\Theta_{L,c_L},
\]
but the definition below does not require Euclidean addition.  A realized action is turned into the implemented next parameter state by a declared update-realization map
\begin{equation}
\label{eq:implemented-action-realization-E}
\boxed{
\mathfrak R_{L,c_L}
:
(\theta,u)
\longmapsto
\theta'
\in\Theta_{L,c_L}.}
\end{equation}
This map contains the declared arithmetic and parameter-update implementation; on a real-valued Euclidean slice it reduces to \(\theta'=\theta+u\).

The represented Transformer, data/target law, automatic-differentiation convention, implementation, and declared protocol variables induce an observation kernel under a fixed fresh-sampling convention: conditional on the current $\theta$ and time $t$, the observation randomness is sampled independently of unrecorded past state. If a sampler or environment carries persistent variables that affect future observations, append those variables to the complete training state and include their kernel in the composition below; the displayed $\Theta\times\Omega_{\mathbf O}$ realization then applies only after that state augmentation (or when no such variables exist). Under the fresh-sampling convention the observation kernel is
\begin{equation}
\label{eq:architecture-observation-kernel-E}
\boxed{
\Gamma_{L,c_L,t}
:
\Theta_{L,c_L}
\rightsquigarrow
\mathsf{Obs}_{L,c_L}.}
\end{equation}
The optimizer is permitted to condition only on this interface.  The interface can be first-order, higher-order, or otherwise enriched, but it is explicit.  Changing the information supplied to the optimizer changes \(\Gamma\); it is not represented by silently enlarging optimizer memory.

For the first-order training regime used in Part~\ref{part:trained-refinement}, the observation contains the implemented stochastic loss/gradient data generated by the exact forward--backward computation.  In particular, at a retained output-gated cut \((j,I)\), the local gradient contribution remains
\begin{equation}
\label{eq:optimizer-interface-cut-gradient-E}
\boxed{
\widehat G_{j,I}^{(L)}
=
\lambda_{j,I}^{(L)}
\phi_{j,I}(h_{j,I}^{(L)})^\top,}
\end{equation}
with \(h_{j,I}^{(L)}\) and \(\lambda_{j,I}^{(L)}\) generated by the complete producer prefix and receiver suffix in Section~\ref{sec:block-cut-insertion}.  Hence \(c_L\) changes the optimizer observation law through the exact PreNorm block maps on both sides of every cut.  The observation kernel is a derived interface of the represented Transformer, not an independent stochastic oracle.

\begin{definition}[Typed optimizer scheme]
\label{def:typed-optimizer-scheme-E}
A \emph{typed optimizer scheme} \(\mathbf O\) assigns to every declared parameter/action type and optimizer-observation interface a causal stochastic optimizer in the sense of Definition~\ref{def:causal-optimizer-E} below.  The assignment is equivariant under mere relabelings of parameter and observation coordinates.  Architectural refinement is not required to commute with the optimizer; its failure to commute is exactly the training/refinement effect measured in Section~\ref{sec:training-refinement-commutator}.
\end{definition}

This definition distinguishes optimizer generality from architecture generality.  SGD, momentum methods, Adam/AdamW, clipping rules, adaptive preconditioners, and schedule-dependent methods correspond to different causal transducers on a declared interface.  A richer optimization algorithm is represented by a richer declared observation interface and/or a different causal response law, not by an unspecified state variable.

\subsection{A general optimizer is a causal transducer}
\label{app:causal-optimizer-E}

For notational simplicity fix one typed interface \((\mathsf{Obs},\mathsf{Act})\).  A pre-action history of length \(t\) is
\[
h_t
=
(o_0,u_0,\ldots,o_{t-1},u_{t-1}),
\]
with \(o_s\in\mathsf{Obs}\) and \(u_s\in\mathsf{Act}\) type-correct at the parameter state reached at time \(s\).  Let \(\mathsf H_t\) denote the measurable space of such histories and \(\mathsf H=\bigsqcup_t\mathsf H_t\).

\begin{definition}[Causal stochastic optimizer]
\label{def:causal-optimizer-E}
A causal stochastic optimizer on the declared interface is a family of stochastic kernels
\begin{equation}
\label{eq:causal-optimizer-kernel-E}
\boxed{
\mathbf O_t
:
\mathsf H_t\times\mathsf{Obs}
\rightsquigarrow
\mathsf{Act},}
\end{equation}
such that the conditional law of the next action depends only on the realized observation/action history up to the present observation.  Equivalently,
\[
u_t
\sim
\mathbf O_t(\,\cdot\mid h_t,o_t)
\]
is nonanticipative with respect to future observations and data.
\end{definition}

A time-dependent optimizer receives the protocol clock or schedule phase as a declared observation, or retains it in its state. Histories are compared for behavioral equivalence at the same declared phase; histories at different phases may be identified only if their phase evolution and all conditional future action laws coincide. In the time-inhomogeneous formulation below, the transition kernels carry the explicit index $t$. The optimizer remains a causal rule from observed history to actions.

The definition above does not postulate a Markov state.  It permits arbitrary finite or infinite memory.  The state used in Part~\ref{part:trained-refinement} can instead be derived from the optimizer's own future behavior.

\subsection{Optimizer causal states are behavioral quotients of histories}
\label{app:optimizer-causal-state-E}

For a history \(h\in\mathsf H\), a positive integer \(k\), and an admissible future observation string
\[
\mathbf o=(o_0^+,\ldots,o_k^+),
\]
the causal kernels of Definition~\ref{def:causal-optimizer-E} determine a joint conditional law of the next \(k+1\) optimizer actions.  Denote this continuation law by
\[
\mathbf O_h^{[k]}(\,d\mathbf u\mid\mathbf o).
\]

\begin{definition}[Optimizer behavioral equivalence]
\label{def:optimizer-behavioral-equivalence-E}
Two admissible optimizer histories \(h,h'\) are behaviorally equivalent,
\[
h\sim_{\mathbf O}h',
\]
when their declared clock phases agree (or generate identical future phase trajectories) and, for every finite \(k\ge0\) and every admissible future observation string \(\mathbf o\),
\begin{equation}
\label{eq:optimizer-history-equivalence-E}
\boxed{
\mathbf O_h^{[k]}(\,\cdot\mid\mathbf o)
=
\mathbf O_{h'}^{[k]}(\,\cdot\mid\mathbf o).}
\end{equation}
The optimizer causal-state space is the quotient
\begin{equation}
\label{eq:optimizer-causal-state-space-E}
\boxed{
\Omega_{\mathbf O}
:=
\mathsf H/\!\sim_{\mathbf O}.}
\end{equation}
\end{definition}

The quotient retains exactly those past distinctions that can change some future optimizer action under some admissible future observation continuation.  A past distinction that can never alter future optimizer behavior is removed.

\begin{theorem}[Canonical causal-state realization of an optimizer]
\label{thm:optimizer-causal-state-realization-E}
Assume the declared optimizer histories and action/observation spaces are standard Borel and admit regular conditional laws. Assume additionally that the behavioral quotient is realized by a measurable surjection $q:\mathsf H\to\Omega_{\mathbf O}$ onto a standard-Borel space, and that the conditional action kernel and history-extension kernel descend through $q$ to measurable kernels (for conditional actions, up to the declared null sets). Then the realized quotient carries an exact causal optimizer state: there exist a next-action kernel
\begin{equation}
\label{eq:optimizer-state-action-kernel-E}
\boxed{
U_{\mathbf O}
:
\Omega_{\mathbf O}\times\mathsf{Obs}
\rightsquigarrow
\mathsf{Act}}
\end{equation}
and an almost-sure state-update kernel
\begin{equation}
\label{eq:optimizer-causal-state-update-E}
\boxed{
M_{\mathbf O}
:
\Omega_{\mathbf O}\times\mathsf{Obs}\times\mathsf{Act}
\rightsquigarrow
\Omega_{\mathbf O}}
\end{equation}
such that, for every admissible history \(h_t\), all future optimizer-action laws obtained from the original causal transducer are reproduced by the state process
\[
\omega_t=[h_t]_{\mathbf O},
\qquad
u_t\sim U_{\mathbf O}(\,\cdot\mid\omega_t,o_t),
\qquad
\omega_{t+1}\sim M_{\mathbf O}(\,\cdot\mid\omega_t,o_t,u_t).
\]
Moreover this state is behaviorally minimal: if another exact state realization \(\zeta(h)\) makes every future action law factor through \(\zeta(h)\), then there is a well-defined map
\[
\pi:\operatorname{im}\zeta\to\Omega_{\mathbf O}
\]
with
\[
\pi(\zeta(h))=[h]_{\mathbf O}.
\]
Thus every exact optimizer state realization refines the causal-state quotient.
\end{theorem}

\begin{proof}
For a causal class \(\omega=[h]_{\mathbf O}\) and present observation \(o\), define \(U_{\mathbf O}(\cdot\mid\omega,o)\) to be the one-step action law of the original optimizer after history \(h\) and observation \(o\).  Equation~\eqref{eq:optimizer-history-equivalence-E} with \(k=0\) makes this independent of the representative.  Condition on the action through the declared regular conditional kernel and append $(o,u)$ to the history for almost every action under that kernel. Behavioral equivalence and the assumed measurable descent identify the updated class independently of the representative up to the declared conditional null sets, giving $M_{\mathbf O}$.  Iterating these two kernels reproduces the finite continuation laws by construction.

For minimality, suppose \(\zeta(h)=\zeta(h')\).  Exact state sufficiency implies that every future action law under every admissible future observation continuation is the same after \(h\) and \(h'\).  Hence \(h\sim_{\mathbf O}h'\), so \([h]_{\mathbf O}=[h']_{\mathbf O}\).  Therefore \(\pi(\zeta(h))=[h]_{\mathbf O}\) is well defined.
\end{proof}

The behavioral quotient in Equation~\eqref{eq:optimizer-causal-state-space-E} identifies histories with identical future optimizer-action laws. A measurable state realization additionally requires the stated quotient and kernel-descent conditions. Any exact realized optimizer state refines this behavioral equivalence.  Familiar moment variables are useful only insofar as they realize this state.

\paragraph{Examples.}
A memoryless SGD rule has a trivial optimizer causal state once the externally supplied schedule is placed in the observation interface.  Momentum admits velocity as an exact finite-dimensional realization.  Adam admits the first- and second-moment variables, together with any internally retained schedule state, as an exact realization on its standard interface.  An optimizer with genuinely irreducible long memory may have an infinite-dimensional causal-state quotient.  The formalism does not require finite optimizer memory.

\subsection{Differentiable optimizer realizations for tangent analysis}
\label{app:differentiable-optimizer-realization}

The canonical causal state is a behavioral quotient and need not carry a differentiable structure. Tangent calculations therefore use an exact differentiable realization rather than differentiating the quotient itself. A \emph{differentiable optimizer realization} is a finite- or Banach-dimensional state space \(Z_{\mathbf O}^{\mathrm{diff}}\) with smooth state/action update maps on a declared regular stratum, together with an exact realization map from histories, such that all optimizer action laws are reproduced. By Theorem~\ref{thm:optimizer-causal-state-realization-E}, every such exact realization has a canonical factor map
\[
\boxed{
\pi_{\mathbf O}:Z_{\mathbf O}^{\mathrm{diff}}\longrightarrow\Omega_{\mathbf O}.
}
\]
The behavioral quotient answers which past distinctions matter for future optimizer actions; the differentiable realization supplies coordinates in which derivatives of the training update are defined. They are not the same object.

SGD, momentum, and the standard real-valued Adam recursion provide familiar differentiable realizations on strata where their update maps are smooth. Clipping boundaries, quantization boundaries, discrete schedule switches, and other singular implementation events are either treated piecewise on smooth strata or excluded from the resolved tangent regime. These regular strata realize the causal optimizer and the differentiable training-response calculations used below.

\subsection{The optimizer response functional is induced by the causal state}
\label{app:optimizer-response-from-causal-state-E}

A useful additional optimizer-response functional is
\[
\mathscr Q_{\mathrm{Opt}}(G,\omega)
=
-\mathbb E\langle G,U_{\mathrm{Opt}}(G,\omega,\xi)\rangle.
\]
The causal-state construction makes its domain precise.  Suppose the declared observation interface contains a gradient covector \(G\), normalized protocol variables \(s\), and exogenous stochastic variables generated by \(\Gamma\).  For optimizer causal state \(\omega\), define
\begin{equation}
\label{eq:causal-optimizer-response-functional-E}
\boxed{
\mathscr Q_{\mathbf O}(G,\omega;s)
:=-
\int
\langle G,u\rangle
\,U_{\mathbf O}(du\mid\omega,o(G,s)).}
\end{equation}
This is a derived observable of the causal optimizer state and the declared training observation.

Common-scale response degree is likewise a property of the causal transducer rather than of an informal list of moment variables.  Suppose positive gradient rescaling induces measurable actions
\[
S_a^{\mathrm{Obs}}:\mathsf{Obs}\to\mathsf{Obs}
\]
and, by rescaling complete past observations, an induced action
\[
S_a^{\Omega}:\Omega_{\mathbf O}\to\Omega_{\mathbf O}.
\]
The optimizer has response degree \(\nu\) on a homogeneous causal-state domain when the action law obeys
\begin{equation}
\label{eq:causal-optimizer-homogeneity-E}
\boxed{
U_{\mathbf O}
(\,\cdot\mid
S_a^{\Omega}\omega,
S_a^{\mathrm{Obs}}o)
=
(D_{a^\nu})_\#
U_{\mathbf O}(\,\cdot\mid\omega,o),}
\end{equation}
where \(D_{a^\nu}u=a^\nu u\) in the declared action coordinates.  Equation~\eqref{eq:causal-optimizer-homogeneity-E} defines the corresponding optimizer homogeneity regime directly on the behaviorally derived causal state.

\subsection{Architecture and optimizer induce the closed-loop training kernel}
\label{app:derived-closed-loop-training-E}

Return to the Transformer fiber \((L,c_L)\).  The optimizer causal state for its typed interface is denoted
\[
\Omega_{\mathbf O;L,c_L}.
\]
The complete closed-loop training state is now \emph{derived} as
\begin{equation}
\label{eq:derived-complete-training-state-E}
\boxed{
\mathcal Z^{\mathbf O}_{L,c_L}
=
\Theta_{L,c_L}\times\Omega_{\mathbf O;L,c_L}.}
\end{equation}

Given \((\theta,\omega)\), one training step consists of three typed operations:
\begin{enumerate}
    \item the represented Transformer and data law generate an observation
    \[
    o\sim\Gamma_{L,c_L,t}(\,\cdot\mid\theta);
    \]
    \item the causal optimizer generates an action
    \[
    u\sim U_{\mathbf O}(\,\cdot\mid\omega,o);
    \]
    \item implementation realizes the new parameter state and optimizer causal state,
    \[
    \theta'=\mathfrak R_{L,c_L}(\theta,u),
    \qquad
    \omega'\sim M_{\mathbf O}(\,\cdot\mid\omega,o,u).
    \]
\end{enumerate}
Therefore the one-step learning kernel used abstractly in Part~\ref{part:trained-refinement} is the composite
\begin{equation}
\label{eq:derived-training-kernel-E}
\boxed{
\begin{aligned}
K^{\mathbf O}_{L,c_L,t}
\bigl((\theta,\omega),d\theta'\,d\omega'\bigr)
={}&
\int_{\mathsf{Obs}_{L,c_L}}
\Gamma_{L,c_L,t}(do\mid\theta)
\\
&\times
\int_{\mathsf{Act}_{L,c_L}(\theta)}
U_{\mathbf O}(du\mid\omega,o)
\\
&\times
\delta_{\mathfrak R_{L,c_L}(\theta,u)}(d\theta')
M_{\mathbf O}(d\omega'\mid\omega,o,u).
\end{aligned}}
\end{equation}

\begin{theorem}[Closed-loop determination of the learning experiment]
\label{thm:closed-loop-determination-E}
Fix \(L\), \(c_L\), the data/target law, initialization law, declared observation interface \(\Gamma_{L,c_L,t}\), implementation map \(\mathfrak R_{L,c_L}\), and typed causal optimizer scheme \(\mathbf O\).  Then the complete training law
\begin{equation}
\label{eq:derived-training-law-E}
\boxed{
\mu^{\mathbf O}_{L,c_L,T}
=
\mu^0_{L,c_L}
K^{\mathbf O}_{L,c_L,0}
\cdots
K^{\mathbf O}_{L,c_L,T-1}}
\end{equation}
is determined by those data.  In particular, the map
\[
c_L
\longmapsto
\{\mu^{\mathbf O}_{L,c_L,T}\}_{T\ge0}
\]
is induced by the exact Transformer coordinate and optimizer interaction.  No independent training kernel or unspecified optimizer state is required.
\end{theorem}

\begin{proof}
The architecture coordinate determines the represented Transformer, parameter space, matched forward/backward computation, and hence the observation kernel.  The optimizer scheme determines its canonical causal-state realization.  Equation~\eqref{eq:derived-training-kernel-E} then determines every one-step closed-loop kernel, and composition with the declared initialization law gives Equation~\eqref{eq:derived-training-law-E}.
\end{proof}

The resulting closed-loop law specifies the evolution of task responses and contextual-semantic probes through the corresponding pushforwards.

\subsection{The task-approximation quotient}
\label{app:task-approximation-quotient-E}

The task-predictor quotient retains the predictor as an observable of training without requiring parameter-state identity. It is distinct from the contextual-semantic observable, which records the computational roles of selected internal cuts.

Let
\[
r_{L,c_L}:
\Theta_{L,c_L}\to\mathcal H
\]
be the realized predictor.  Let \(\mathbb P_X\) be the declared task-input law and write
\begin{equation}
\label{eq:task-predictor-equivalence-E}
\boxed{
f\sim_{\mathrm{app}}g
\iff
f(X)=g(X)
\quad\mathbb P_X\text{-almost surely},}
\end{equation}
with equality interpreted as equality of the induced predictive law for stochastic predictors.  More generally, the right side can be replaced by equality under the complete declared family of task observables; the choice must be fixed before training is analyzed.

Define the task-approximation space
\begin{equation}
\label{eq:task-approximation-space-E}
\boxed{
\mathsf A_{L,c_L}
:=
r_{L,c_L}(\Theta_{L,c_L})/\!\sim_{\mathrm{app}}}
\end{equation}
and the task projection
\begin{equation}
\label{eq:task-approximation-map-E}
\boxed{
q^{\mathrm{app}}_{L,c_L}(\theta,\omega)
:=
[r_{L,c_L}(\theta)]_{\mathrm{app}}.}
\end{equation}
The optimizer causal state is intentionally absent from this quotient.

Let
\[
\mathcal R_{L,c_L}:\mathsf A_{L,c_L}\to\R
\]
be the evaluation-risk functional induced by the declared target law and loss.  If \(a=[r(\theta)]_{\mathrm{app}}\), then \(\mathcal R_{L,c_L}(a)\) is the population evaluation loss of that predictor class.

\subsection{The optimizer comparison state is also derived behaviorally}
\label{app:optimizer-state-lift-E}

Part~\ref{part:trained-refinement} uses a same-function comparison lift between adjacent model scales.  The parameter part of that lift is supplied by the identity-fiber refinement.  The optimizer part should not be defined by an optimizer-specific instruction such as ``copy the moments.''  It is induced from the causal-state quotient whenever the declared coarse history has a type-correct fine history embedding.

Fix a declared comparison datum \(c=(c_c,c_f)\in\mathcal C_L\times\mathcal C_{mL}\). The notation \(D_Lc=c_c\) and \(D_{mL}c=c_f\) refers only to the two component architecture coordinates of this datum; the maps are projections of an explicitly selected pair, not depth-RG couplings. A type-compatible same-function architecture comparison is
\[
\mathcal T_{L,D_Lc}
\longrightarrow
\mathcal T_{mL,D_{mL}c}.
\]
Let
\[
\mathfrak J^{\mathrm{hist}}_{m,L,c}
:
\mathsf H_{L,D_Lc}
\longrightarrow
\mathsf H_{mL,D_{mL}c}
\]
be the declared history embedding obtained by applying the parameter comparison lift to inherited coordinates and the declared neutral/identity-fiber convention to inserted coordinates and observations.

\begin{definition}[Behaviorally compatible optimizer refinement]
\label{def:behaviorally-compatible-optimizer-refinement-E}
The optimizer scheme is behaviorally compatible with the comparison history embedding when
\begin{equation}
\label{eq:optimizer-history-lift-congruence-E}
\boxed{
h\sim_{\mathbf O,L}h'
\quad\Longrightarrow\quad
\mathfrak J^{\mathrm{hist}}_{m,L,c}(h)
\sim_{\mathbf O,mL}
\mathfrak J^{\mathrm{hist}}_{m,L,c}(h').}
\end{equation}
\end{definition}

\begin{proposition}[Canonical optimizer-state comparison lift]
\label{prop:canonical-optimizer-state-lift-E}
Under Equation~\eqref{eq:optimizer-history-lift-congruence-E}, the comparison history embedding descends uniquely to a map
\begin{equation}
\label{eq:canonical-optimizer-state-lift-E}
\boxed{
J^{\Omega}_{m,L,c}
:
\Omega_{\mathbf O;L,D_Lc}
\longrightarrow
\Omega_{\mathbf O;mL,D_{mL}c},
\qquad
[h]\longmapsto
[\mathfrak J^{\mathrm{hist}}_{m,L,c}(h)].}
\end{equation}
If the history embeddings compose coherently, then the optimizer-state lifts compose coherently.
\end{proposition}

\begin{proof}
Equation~\eqref{eq:optimizer-history-lift-congruence-E} makes the quotient map independent of the chosen representative.  Coherence follows by applying the quotient to the coherent composition of history embeddings.
\end{proof}

The complete comparison lift in Section~\ref{sec:refinement-innovations} is therefore
\begin{equation}
\label{eq:complete-derived-comparison-lift-E}
\boxed{
J_{m,L,c}
=
J^{\Theta}_{m,L,c}
\times
J^{\Omega}_{m,L,c},}
\end{equation}
where \(J^{\Theta}\) is the architecture-level same-function parameter lift.  The optimizer is still allowed to respond differently after refinement; Equation~\eqref{eq:training-refinement-commutator} measures exactly that failure of commutation.  Behavioral compatibility is only the condition required to compare the two optimizer states type-correctly.

\subsection{Comparison scope}
The closed-loop kernel determines the complete training law for each declared architecture and optimizer. The behaviorally compatible history lift supplies a type-correct comparison of two such experiments, without asserting that their training kernels commute. Long-time convergence and terminal prediction are separate dynamical properties to be derived from a specified kernel.

\section{Architectural innovation geometry and optimizer comparison}\label{app:architecture-innovation-details}
\subsection{Effective innovations are the fine tangent directions not inherited from the coarse model}

At a comparison state, let
\[
V_{L,c}
=
T_\theta\Theta_{L,c}/\ker Dr_{L,c}
\]
be the function-changing tangent space. The differentiated parameter lift induces an injection
\[
I_{m,L,c}:V_{L,c}\hookrightarrow V_{mL,c}.
\]
The effective innovation space is
\[
\boxed{
N_{m,L,c}
=
V_{mL,c}/I_{m,L,c}V_{L,c}.
}
\]
It contains precisely the first-order predictor variations made available by the fine represented model modulo those already available to the coarse model.

On a finite reachable support \(u_1,\ldots,u_M\), an inserted feed-forward output perturbation has the form
\[
v(u_r)=\Delta W_{n+1}z_n(u_r)+\Delta b_{n+1}.
\]
For ReLU, width \(w\ge M-1\) permits an explicit hidden-feature construction whose augmented evaluation matrix has full row rank. Every vector-valued residual assignment on the support is then realized exactly by an output-side perturbation. Full row rank is an open condition, so the constructed point has a nonempty regular interpolation neighborhood. This is the local fact used below: overparameterization creates a regular same-function sector in which newly exposed output coordinates can realize arbitrary finite-support residual assignments. The full-row-rank condition holds on the constructed regular interpolation neighborhood and supplies the local tangent directions used below.

\subsection{Global realizability of local tangent perturbations}
\label{sec:innovation-gluing}

Locally admissible tangent perturbations
need not be realizable by a single
globally compatible Transformer
parameter variation. Shared residual
carriers, tied parameters, and
overlapping represented interfaces
impose compatibility conditions
between the local changes.

Let \(N_j\) denote the local
innovation space at retained fine
cut \(j\), and collect these classes
into
\[
N^{\mathrm{loc}}_{m,L,c}=\bigoplus_j N_j.
\]
The exact Transformer wiring imposes linearized compatibility constraints of three kinds: residual-state agreement across shared carriers, equality constraints induced by tied parameter families, and overlap/interface agreement where neighboring local perturbations describe the same represented cut variable. When these linearized constraints descend to the local innovation quotients (in particular, when they annihilate the inherited tangent representatives), package them into the compatibility operator
\[
C_{m,L,c}:
N^{\mathrm{loc}}_{m,L,c}
\longrightarrow
E^{\mathrm{res}}_{m,L,c}
\oplus
E^{\mathrm{share}}_{m,L,c}
\oplus
E^{\mathrm{overlap}}_{m,L,c}.
\]
These are necessary compatibility conditions. The actual global parameter tangent space determines which compatible local classes can be realized simultaneously. Let
\[
\rho_{m,L,c}:T_{\theta_f}\Theta_{mL,c}\longrightarrow
\bigoplus_jN_j
\]
restrict a single global parameter tangent to its selected local innovation classes. For a descended compatibility operator the global restriction satisfies $C_{m,L,c}\rho_{m,L,c}=0$. If the constraints do not descend to these quotient classes, formulate them on representative local tangents instead; $\operatorname{im}\rho$ remains the exact globally realizable space. In the descended case,
\begin{equation}\label{eq:gluing-necessary-exact}
\boxed{\mathcal I^{\mathrm{real}}_{m,L,c}:=\operatorname{im}\rho_{m,L,c}
\ \subseteq\ \ker C_{m,L,c}.}
\end{equation}
The image is the exact globally realizable innovation space. Equality with $\ker C$ requires an exact local-to-global lifting argument for the selected cover and its parameter-sharing equations. The compatibility kernel alone is not substituted for that image. In a free, unshared inserted output-gate sector, each output perturbation is already a global parameter tangent and hence belongs to $\mathcal I^{\mathrm{real}}$.

\subsection{Receiver continuation determines which innovations matter}
\label{sec:receiver-effective-innovations}

New function space is not automatically useful function space. The downstream receiver decides whether an innovation changes the objective. For a gluable innovation \(v\) and the receiver covector \(\lambda\) induced by the downstream loss, the first-order task work is
\[
\mathsf w(\lambda,v)=\langle\lambda,v\rangle.
\]
The same predictor variation can be useful to one receiver and invisible to another.

On finite support, full-row-rank interpolation in a globally free output-gate sector realizes the residual assignment
\[
v(u_r)=-\lambda_r^\sharp
\]
by one admissible global parameter tangent. Its receiver functional is
\begin{equation}\label{eq:receiver-aligned-descent}
\boxed{\Lambda[v]=-\sum_r\mu_r\|\lambda_r\|^2.}
\end{equation}
Every nonzero receiver signal in this regular sector therefore supplies a strict fine-model descent direction. To conclude that its predictor variation is \emph{new relative to the coarse model}, assume the coarse predictor is first-order stationary: $\Lambda$ annihilates the inherited tangent image $I_{m,L,c}V_{L,c}$. Then $\Lambda[v]<0$ implies $v\notin I_{m,L,c}V_{L,c}$, so $[v]\ne0$ in $N_{m,L,c}$. Under this stationarity
assumption, \(\Lambda\)
descends to the quotient
\(N_{m,L,c}\).
A class on which the
descended functional is
strictly negative is nonzero,
is not inherited from the
coarse tangent image, and
gives a strict first-order
decrease in the selected
objective.

\begin{proposition}[Finite attainable improvement along a regular fine-model path]
\label{prop:finite-attainable-chord}
Let $\gamma:[0,a]\to\Theta_{mL,c}$ be a twice differentiable path through the lifted parameter state, lying in one admissible regular sector and respecting all parameter-sharing constraints. Write $L(s)=\mathcal L(r_{mL,c}(\gamma(s)))$. If $-L'(0)=g>0$ and $L''(s)\le M$ for $0\le s\le a$, with $M>0$, then
\begin{equation}\label{eq:finite-gain-chord-bound}
L(0)-L(s)\ge gs-\tfrac12Ms^2\qquad(0\le s\le a).
\end{equation}
If $g/M\le a$, the best attainable loss decrease in the fine represented class from the lifted state is at least $g^2/(2M)>0$.
\end{proposition}
\begin{proof}
Taylor's integral formula gives $L(s)=L(0)+sL'(0)+\int_0^s(s-r)L''(r)\,dr\le L(0)-gs+Ms^2/2$. Take $s=g/M$ when it lies in the declared path interval.
\end{proof}

This bound concerns attainable improvement along an actual fine-model parameter path. The optimizer's selection of that path is a different statement governed by its transition kernel.

The innovation quotient compares fine predictor directions against those inherited from the coarse architecture. Receiver covectors test their local effect on the task; the optimizer kernel determines which parameter updates are actually taken.

\subsection{Training and refinement do not generally commute}
\label{sec:training-refinement-commutator}

Let
\[
K_{L,t}^{\mathbf O},
\qquad
K_{mL,t}^{\mathbf O}
\]
be the derived closed-loop training kernels at the two adjacent depths. Let $J_t:Z_L\rightsquigarrow Z_{mL}$ be a measurable coarse-to-fine comparison kernel on complete training states, with the correct parameter marginal. We use \emph{forward} Markov-kernel composition: $\mu_{t+1}=\mu_tK_t$ and $K_{c,t}J_{t+1}$ first updates coarse state, then lifts it. The one-step signed-kernel discrepancy is
\begin{equation}\label{eq:training-refinement-commutator}
\boxed{\mathfrak C_{m,L,c;t}
=K_{L,t}^{\mathbf O}J_{t+1}-J_tK_{mL,t}^{\mathbf O}
:\ Z_L\rightsquigarrow\mathsf M_{\rm signed}(Z_{mL}).}
\end{equation}
For a time-independent lift, $\mathfrak C_t=K_{L,t}^{\mathbf O}J-JK_{mL,t}^{\mathbf O}$. Zero discrepancy is precisely the one-step intertwining relation. A parameter lift alone does not produce $J_t$ on optimizer memory; this comparison requires its own measurable state lift.

With $K^{a:b}=K_a\cdots K_{b-1}$ and identity kernels for empty products, the exact nonhomogeneous Duhamel identity is
\begin{equation}\label{eq:typed-duhamel}
\boxed{
K_L^{0:T}J_T-J_0K_{mL}^{0:T}
=\sum_{t=0}^{T-1}K_L^{0:t}\,\mathfrak C_{m,L,c;t}\,K_{mL}^{t+1:T}.}
\end{equation}
Indeed, introduce and subtract the successive operators $K_L^{0:t}J_tK_{mL}^{t:T}$. Adjacent terms telescope. For $T=1$ the formula is the definition of $\mathfrak C_0$, and for $T=2$ it reads $K_{L,0}\mathfrak C_1+\mathfrak C_0K_{mL,1}$, confirming the composition order. The same identity holds for deterministic comparison maps after identifying each map with its Dirac kernel.

The same comparison has an exact finite-update work identity. If \(u_c\) is the coarse action, \(u_f\) the fine action from the lifted state, and \(Iu_c\) the inherited-coordinate copy, define
\[
c_{m,L,c;t}=u_f-Iu_c.
\]
Then
\[
\boxed{
\mathscr W_{m,L,c;t}
=
-\int_0^1
D\ell_{mL}
\big|_{\iota\theta+Iu_c+s c_{m,L,c;t}}
[c_{m,L,c;t}]\,ds.
}
\]
No infinitesimal-step approximation is used. This work identity is conditional on a differentiable loss along the displayed chord; it is not a positivity claim.

\section{Exact semantic quotients beyond Set}\label{sec:exact-quotients}

The distinction relation is useful because, under a stronger semantic assumption, it classifies the quotient part of a realized receiver exactly. This is the point at which the familiar Set-level kernel results enter. They are not separate foundations; they are consequences of the quotient/equivalence-relation correspondence.

Assume in this subsection that $\mathcal S$ is Barr-exact. The deterministic realization
\[
\mathcal S=\mathbf{Set}
\]
is the primary example.

For $S\in\mathcal S$, let
\[
\mathsf{Quot}(S)
\]
denote the category whose objects are regular epimorphisms
\[
q:S\twoheadrightarrow Q
\]
and whose morphisms are commuting triangles over $S$.

\begin{theorem}[Exact quotient classification]\label{thm:exact-quotient-classification}
For every $S$ in an exact semantic category, the kernel-pair construction gives an equivalence
\[
\boxed{
\operatorname{KP}:
\mathsf{Quot}(S)
\simeq
\mathsf{Eq}(S),
}
\]
up to the ordinary choice of quotient representatives.

Consequently:
\begin{enumerate}[label=(\roman*)]
\item a map $r:S\to T$ factors through a regular quotient $q:S\twoheadrightarrow Q$ exactly when
\[
\operatorname{KP}(q)
\subseteq
\operatorname{KP}(r);
\]
\item two regular quotients of $S$ with the same kernel pair are uniquely isomorphic over $S$;
\item every internal equivalence relation on $S$ is the kernel pair of its effective quotient.
\end{enumerate}
\end{theorem}

\begin{proof}
Exactness says that every internal equivalence relation is effective: it is the kernel pair of a coequalizer, and that coequalizer is a regular epimorphism. Conversely, every regular epimorphism is the coequalizer of its kernel pair. The universal property of the coequalizer gives factorization in (i). If two regular epimorphisms have the same kernel pair, each factors uniquely through the other; the two induced maps are inverse by uniqueness, proving (ii). Statement (iii) is the effectiveness clause itself.
\end{proof}

This theorem unifies the set-level fiber-factorization, unique-conversion, and canonical-quotient statements. They answer one question: once a represented receiver has been reduced to an effective equivalence relation, which quotient presentation exactly retains those distinctions?

An arbitrary realized arrow need not itself be a regular epimorphism. Factor it as
\[
R(X)
\xrightarrow{q_F}
\operatorname{im}R(F)
\hookrightarrow
R(Y),
\]
where $q_F$ is regular epi. Then
\[
\boxed{
\operatorname{KP}(R(F))
=
\operatorname{KP}(q_F).
}
\]
Thus the distinction relation canonically classifies the quotient part of the realized receiver without repeatedly corestricting notation to the realized image.

\paragraph{Set specialization.}
In $\mathbf{Set}$, Theorem~\ref{thm:exact-quotient-classification} says that a function is classified at the distinction grain by its fiber partition. If
\[
q:X\twoheadrightarrow Z,
\qquad
q':X\twoheadrightarrow Z'
\]
are surjective, then
\[
\ker q=\ker q'
\]
if and only if there is a unique bijection
\[
T:Z\xrightarrow{\cong}Z'
\]
with
\[
q'=Tq.
\]
Likewise, a function $f:X\to W$ factors through $q$ exactly when it is constant on every $q$-fiber.

When the paper works explicitly in deterministic Set semantics, we will often suppress the realization symbol and write $F(x)$, $\ker F$, or $\operatorname{im}F$ for the corresponding constructions on $R(F)$. The represented arrow in $\mathcal A$ remains the architectural object; the suppressed notation only avoids carrying $R$ through every elementary calculation.

\subsection{Factorization equivalence and semantic quotient equivalence are different}

Fix a represented branch
\[
B:X\to Z
\]
and two objects of its factorization category,
\[
X\xrightarrow{Q}Y\xrightarrow{G}Z,
\qquad
X\xrightarrow{Q'}Y'\xrightarrow{G'}Z.
\]
An isomorphism
\[
T:(Q,G)\xrightarrow{\cong}(Q',G')
\]
in $\mathsf{Fact}_{\mathcal A}(B)$ is a represented passive re-presentation:
\[
TQ=Q',
\qquad
G=G'T.
\]

Equality of distinction relations is weaker. It gives, after exact semantic quotienting, a unique semantic isomorphism
\[
\bar T:
\operatorname{im}R(Q)
\xrightarrow{\cong}
\operatorname{im}R(Q')
\]
over $R(X)$. Nothing in exactness requires $\bar T$ to be the realization of an available architectural arrow $T:Y\to Y'$, still less an isomorphism in $\mathsf{Fact}_{\mathcal A}(B)$.

The paper therefore distinguishes three claims:
\[
\boxed{
\begin{array}{c}
\text{same distinction relation}\\
\Downarrow\\
\text{unique semantic quotient conversion}\\
\not\Rightarrow\\
\text{represented passive re-presentation}.
\end{array}
}
\]
The missing implication is not a defect. It is where presentation-sensitive continuation enters.

\subsection{How context pullback acts on equivalence relations in Set}

The naturality theorem states
that upstream composition pulls
receiver equivalence relations
back to predecessor states.

In \(\mathbf{Set}\), injective
and surjective producer maps
have different effects on this
pullback. These differences
explain the contextual
local/attention receiver
construction in
Part~\ref{part:represented-architecture}.

Let
\[
A:\Omega\to X
\]
and define
\[
A^\sharp:
\mathsf{Eq}(X)\to\mathsf{Eq}(\Omega),
\qquad
E\longmapsto
(A\times A)^{-1}E.
\]

For the later neural witness we need only two elementary Set facts about this pullback map. If $A$ is surjective, then $A^\sharp$ is order-reflecting: whenever
\[
A^\sharp E\subseteq A^\sharp E',
\]
surjectivity lets every pair in $E$ be lifted to a pair in $\Omega$, so $E\subseteq E'$. Distinct ambient relations therefore cannot merge under a surjective context.

If $A$ is injective, then $A^\sharp$ is surjective onto $\mathsf{Eq}(\Omega)$: transport the classes of any relation on $\Omega$ to $A(\Omega)$ and leave points outside the image as singleton classes. But $A^\sharp$ need not be injective, because two ambient relations may differ only outside $A(\Omega)$. Thus an injective context can preserve every predecessor state while making different ambient receiver relations agree on the reachable image. This is the Set-level mechanism used by the local-versus-attention witness.

Exact quotient classification determines which distinctions are identified by a map; receiver-limited continuation determines which surviving distinctions remain computationally usable.

\subsection{Grothendieck totalization of the effective architecture presheaf}

Every Set-valued presheaf has a Grothendieck construction. Applying it to the derived image presheaf gives
\[
\boxed{
p_D:
\int\mathfrak A_D
\longrightarrow
\mathsf{Ctx}_X.
}
\]
This is a discrete fibration.

An object of the total category is a pair
\[
(A,a),
\qquad
a\in\mathfrak A_D(A),
\]
where \(A\) is a represented
producer context and \(a\) is
a context-dependent equivalence
class of nominal receiver
architectures.

A context morphism
\(\kappa:A'\to A\)
transports the class through
the presheaf restriction map
\[
\mathfrak A_D(\kappa):
\mathfrak A_D(A)\to
\mathfrak A_D(A').
\]

We call
\[
\boxed{
\mathsf{Arch}_D
:=
\int\mathfrak A_D
}
\]
the \emph{effective architecture space at the declared semantic grain}. The terminology is now downstream of the more primitive map
\[
\Sigma:\underline D\to\mathsf{Sem}.
\]
Nothing additional is postulated when the fibration is formed.

\subsection{A finite observable partition-refinement example}\label{sec:finite-partition-example}
\begin{example}[A distinction invisible at one step but visible after two]\label{ex:two-step-distinction}
Take states $x,y,u,v,A,B$ with $\Phi(A)=1$ and $\Phi=0$ elsewhere. Let
\[x\to u,\quad y\to v,\quad u\to A,\quad v\to B,\quad A\to A,\quad B\to B.\]
Initially $x,y$ are in the same current-risk block. They are still together after the first refinement, because both next states have current risk $0$. The next refinement separates them: the second future risk is $1$ from $x$ and $0$ from $y$. Direct finite partition refinement gives the stable partition
\[\{x\},\quad\{y,v,B\},\quad\{u\},\quad\{A\}.\]
This finite-state example isolates the partition-refinement mechanism; the kernel for a specified Transformer training experiment is constructed in Section~\ref{sec:training-closed-loop}.
\end{example}

\section{Finite predictive quotients, marginal closure, and symmetries}\label{app:predictive-reduction-details}
\subsection{Future observations versus a state-Markov quotient}
There are two similar but distinct notions of predictive sufficiency. One can equate two \emph{complete states} if they induce the same law of the entire future sequence $(\Phi(Z_t),\Phi(Z_{t+1}),\ldots)$. Alternatively, one can require that the resulting equivalence itself support a Markov transition for \emph{every} complete-state initial law. The second is stronger: equality of future observable \emph{path} laws need not imply equality of distributions over the next state's predictive equivalence classes.

\begin{example}[Future-path equivalence need not be a strong lumping]\label{ex:future-path-not-lump}
Consider a homogeneous finite Markov chain with states $a,b,c,d,e,A,B$. The current observable is $0$ at $a,b,c,d,e,A$ and $1$ at $B$. Transitions are
\[
a\to c,\quad b\to\tfrac12d+\tfrac12e,\quad c\to\tfrac12A+\tfrac12B,
\quad d\to A,\quad e\to B,\quad A\to A,\quad B\to B.
\]
Starting from $a$ and from $b$, the entire observable path has exactly the same distribution: two initial zeros, followed by a constant zero or one, each with probability $1/2$. Thus $a$ and $b$ are future-path equivalent. Yet $a$ moves certainly to $c$, whose future observation law is a nontrivial mixture, while $b$ moves to $d$ or $e$, each of which has a deterministic future observation law. The transition probabilities from $a,b$ into the future-path-equivalence classes differ. Hence that partition is not strongly lumpable.
\end{example}

The usual \emph{history-based} predictive state, defined by the conditional future-output law given an actual observed history, can be updated by Bayesian conditioning. It is not in general the same object as assigning to each hidden complete state the unconditional future-observation law and declaring those hidden states Markov-equivalent. The hierarchy in Theorem~\ref{thm:forced-minimal} supplies the stronger complete-state quotient when that is the intended goal.

\subsection{A constructive exact algorithm on finite implemented state spaces}\label{sec:finite-predictive-algorithm}
A finite-precision parameter implementation together with finite optimizer memory can yield a finite complete state space $Z$. When it does, the strongly lumpable observable quotient is constructible by partition refinement. Finite parameter precision by itself does not bound the state of an optimizer with unbounded memory.

\begin{definition}[Partition-refinement sequence]\label{def:partition-refine}
Assume for this subsection that $K$ is time homogeneous on a finite state space $Z$, and the set $\Phi(Z)$ is finite. Let $\mathcal P_0$ partition $Z$ into the fibers of $\Phi$. Given $\mathcal P_n$, define $z\sim_{n+1}z'$ precisely when (i) $z,z'$ belong to the same $\mathcal P_n$ block and (ii) for every block $B\in\mathcal P_n$,
\begin{equation}\label{eq:partition-signature}
K(z,B)=K(z',B).
\end{equation}
Let $\mathcal P_{n+1}$ be the resulting partition.
\end{definition}

\begin{theorem}[Finite termination and coarsest exact Markov factor]\label{thm:finite-bisimulation}
The partitions refine monotonically and stabilize after at most $|Z|-|\mathcal P_0|$ strict refinement rounds. The stable partition $\mathcal P_*$ is the coarsest partition refining current-observable equivalence on which $K$ induces an exact Markov chain. Its block-label process reproduces all finite-time joint laws of the retained current-observable observations, for any initial state distribution on $Z$.
\end{theorem}
\begin{proof}
Every strict refinement increases the number of blocks, which is at most $|Z|$. At stabilization, the probability of entering any block depends only on the current block by \eqref{eq:partition-signature}; this defines the quotient transition matrix. Induction on path length shows equality of the finite-dimensional observed path laws. If $\mathcal R$ is any strongly lumpable partition refining $\mathcal P_0$, then by induction every $\mathcal P_n$ block is a union of $\mathcal R$ blocks: $\mathcal R$-equivalent states give identical transition probabilities to $\mathcal R$-block unions and hence to $\mathcal P_n$ blocks. Thus $\mathcal R$ refines $\mathcal P_*$. No coarser admissible Markov quotient exists.
\end{proof}

The partition-refinement algorithm uses the standard
finite-state criterion for exact Markov lumpability
\citep{kemeny1960finite,buchholz1994lumpability}.
Bisimulation methods likewise compare transition
structure, while quantitative extensions measure
departures from exact equivalence
\citep{ferns2004bisimulation}. In this application,
the initial partition is defined by the declared
computational observable and the transition probabilities
come from the implemented training kernel. Refinement
then retains precisely the additional complete-state
distinctions needed for an exact Markov quotient.
Changing either the observable or the training kernel
can change that quotient.

An explicit six-state, two-step partition-refinement example is retained in Appendix~\ref{sec:finite-partition-example}.

\paragraph{Canonical reachability.} For a finite state space and the declared initialization law $\mu_0^S$, let $Z_{\rm reach}^S=\bigcup_{t\ge0}\operatorname{supp}(\mu_0^SK^t)$. It is closed under transitions of positive probability. Running the same algorithm on $Z_{\rm reach}^S$ produces the smallest exact Markov quotient required for \emph{this} canonical experiment, potentially smaller than a quotient of every mathematically possible parameter/optimizer state. With time-dependent kernels, the algorithm must retain time or use backward time-indexed partitions; a single homogeneous transition matrix is not supplied by an arbitrary training schedule.

\subsection{A second reduction: exact linear closure of observable laws}\label{sec:linear-observation-closure}
A Markov quotient is stronger than necessary if our task is only to propagate the \emph{marginal distributions} of selected contextual-semantic or task observables. There is another exact, canonically generated reduction that does not require the reduced coordinates to form a Markov chain.

\begin{definition}[Backward observable-function space]\label{def:observation-space}
For a finite, homogeneous complete-state kernel $K$ let
\begin{equation}\label{eq:observation-space}
\mathcal V=\operatorname{span}\left\{K^n\mathbf 1_{\{\Phi\in B\}}:\ n\ge0,\ B\subseteq\Phi(Z)\right\}\subseteq\mathbb R^Z.
\end{equation}
The constant function belongs to $\mathcal V$. For a time-dependent kernel, replace $K^n$ by $K_tK_{t+1}\cdots K_{t+n-1}$ and write $\mathcal V_t$.
\end{definition}

\begin{theorem}[Minimal linear closure of the full observable marginal law]\label{thm:linear-closure}
For a finite homogeneous $K$, $\mathcal V$ is the smallest $K$-invariant linear subspace containing the indicators of all current observable values. Let $b_1,\ldots,b_r$ be a basis, with $r\le |Z|$. Write
\[Kb_j=\sum_{i=1}^r A_{ji}b_i,\qquad m_{t,i}=\mu_t(b_i).\]
Then
\begin{equation}\label{eq:linear-closed-evo}
\boxed{m_{t+1,j}=\sum_{i=1}^r A_{ji}m_{t,i}.}
\end{equation}
For every measurable observable event $B\subseteq E$, $\zeta_t(B)$ is a fixed linear functional of $m_t$. Hence the \emph{entire} one-time risk-response law is recovered from an exact finite-dimensional linear evolution, regardless of whether the current-observable quotient is Markov or any useful state-space lumping exists.
\end{theorem}
\begin{proof}
By construction, $K\mathcal V\subseteq\mathcal V$ and every current observable indicator belongs to $\mathcal V$. Any other invariant subspace containing these indicators contains all their $K$-iterates, and hence $\mathcal V$. From $\mu_{t+1}=\mu_tK$,
\[m_{t+1,j}=\mu_t(Kb_j)=\sum_iA_{ji}\mu_t(b_i).\]
Finally $\zeta_t(B)=\mu_t(\mathbf1_{\{\Phi\in B\}})$ and the indicator belongs to $\mathcal V$.
\end{proof}

\begin{corollary}[Exact finite algorithm; no guessed dimension]\label{cor:linear-algorithm}
The ascending subspaces
\[\mathcal V^{(0)}=\operatorname{span}\{\mathbf1_{\{\Phi=y\}}:y\in\Phi(Z)\},\quad
\mathcal V^{(n+1)}=\mathcal V^{(n)}+K\mathcal V^{(n)}\]
stabilize after at most $|Z|-\dim\mathcal V^{(0)}$ strict growth steps; their stable value is $\mathcal V$. The closure dimension $r$ is determined by the invariant function space generated by the actual kernel.
\end{corollary}

A closed linear recursion does not automatically define a stochastic transition matrix on $r$ abstract states: the basis may consist of signed functions, and the matrix $A$ need not have nonnegative entries. It guarantees equality of the observable \emph{marginals}, not equality of complete observable \emph{path laws}. The strong partition quotient in Theorem~\ref{thm:finite-bisimulation} supplies the latter when available. This difference matters because one-time distributions of receiver distinctions and task responses differ from joint trajectories of changing computational roles.

For a finite time-inhomogeneous $K_t$, define $\mathcal V_t$ using forward kernel products. Then $K_t\mathcal V_{t+1}\subseteq\mathcal V_t$, and bases at successive times give the exact nonautonomous linear system $m_{t+1}=A_tm_t$. Infinite optimizer memory or a continuous complete-state space can produce an infinite-dimensional observable-function space.

\subsection{Exact reductions from Transformer re-labeling symmetries}\label{sec:predictive-symmetry}
The specified Transformer has representational re-labelings, including permutations of equally typed hidden FFN units and, where head masks and temperature declarations permit, attention-head re-indexings with the corresponding output-block permutations. A relabeling preserves the realized predictor when all coupled coordinates are transported consistently. ABF's typed optimizer definition requires equivariance under \emph{mere coordinate relabelings}; its implementation and observation interface must be checked on the same action.

\begin{proposition}[Verifiable symmetry reduction]\label{prop:symmetry-lumping}
Let a group $G$ act measurably on the complete training state $Z$. Suppose the following two identities hold for the \emph{actual} declared experiment:
\begin{equation}\label{eq:actual-symmetry}
\Phi(gz)=\Phi(z),\qquad K_t(gz,gA)=K_t(z,A)
\quad(g\in G,\ z\in Z,\ A\in\mathcal B(Z)).
\end{equation}
Then every function in the forced predictive-information hierarchy is $G$-invariant; equivalently, $\mathcal H_t^{(\infty)}\subseteq\mathcal B(Z)^G$. No future retained observable can distinguish states in the same $G$-orbit. If the orbit space is measurably realizable, the orbit map is a valid exact Markov quotient.
\end{proposition}
\begin{proof}
The selected observable is $G$-invariant. If $f$ is invariant, equivariance gives
\[(K_tf)(gz)=\int f(z')K_t(gz,dz')=\int f(gw)K_t(z,dw)=(K_tf)(z).\]
Induction through \eqref{eq:forced-hierarchy} proves the first assertion. For the orbit quotient, the equivariance identity makes the transition probability into a union of orbits constant along each orbit, which is exactly strong lumpability.
\end{proof}

For an FFN-unit permutation, Appendix A's affine/activation composition preserves the predictor. The second identity follows when the observation, optimizer-action, causal-memory, and implementation kernels transform equivariantly. Masks, frozen parameters, and coordinate-dependent optimizer or numerical rules can break this symmetry. The two-weight example already demonstrates the failure for nonorthogonal parameter rescalings that preserve the predictor but change gradient descent.

\bibliographystyle{plainnat}
\bibliography{references_clean}

@inproceedings{arora2018acceleration,
  title = {On the Optimization of Deep Networks: Implicit Acceleration by Overparameterization},
  author = {Arora, Sanjeev and Cohen, Nadav and Hazan, Elad},
  booktitle = {Proceedings of the 35th International Conference on Machine Learning},
  series = {Proceedings of Machine Learning Research},
  volume = {80},
  pages = {244--253},
  year = {2018},
  url = {https://proceedings.mlr.press/v80/arora18a.html}
}

@article{barnett2015transducer,
  title = {Computational Mechanics of Input--Output Processes: Structured Transformations and the $\epsilon$-Transducer},
  author = {Barnett, Nix and Crutchfield, James P.},
  journal = {Journal of Statistical Physics},
  volume = {161},
  pages = {404--451},
  year = {2015},
  doi = {10.1007/s10955-015-1327-5},
  url = {https://arxiv.org/abs/1412.2690}
}

@inproceedings{blackwell1951comparison,
  title={Comparison of Experiments},
  author={Blackwell, David},
  booktitle={Proceedings of the Second Berkeley Symposium on Mathematical Statistics and Probability},
  pages={93--102},
  year={1951},
  publisher={University of California Press}
}

@article{brunton2016koopman,
  title = {Koopman Invariant Subspaces and Finite Linear Representations of Nonlinear Dynamical Systems for Control},
  author = {Brunton, Steven L. and Brunton, Bingni W. and Proctor, Joshua L. and Kutz, J. Nathan},
  journal = {PLOS ONE},
  volume = {11},
  number = {2},
  pages = {e0150171},
  year = {2016},
  doi = {10.1371/journal.pone.0150171}
}

@article{buchholz1994lumpability,
  title = {Exact and Ordinary Lumpability in Finite Markov Chains},
  author = {Buchholz, Peter},
  journal = {Journal of Applied Probability},
  volume = {31},
  number = {1},
  pages = {59--75},
  year = {1994},
  doi = {10.2307/3215235}
}

@inproceedings{chen2024softmaxdynamics,
  title = {Training Dynamics of Multi-Head Softmax Attention for In-Context Learning: Emergence, Convergence, and Optimality},
  author = {Chen, Siyu and Sheen, Heejune and Wang, Tianhao and Yang, Zhuoran},
  booktitle = {Proceedings of the 37th Annual Conference on Learning Theory},
  year = {2024},
  url = {https://arxiv.org/abs/2402.19442}
}

@article{dewhurst2018individuation,
  title={Individuation without Representation},
  author={Dewhurst, Joe},
  journal={The British Journal for the Philosophy of Science},
  volume={69},
  number={1},
  pages={103--116},
  year={2018},
  doi={10.1093/bjps/axw018}
}

@inproceedings{dinh2017sharp,
  title = {Sharp Minima Can Generalize For Deep Nets},
  author = {Dinh, Laurent and Pascanu, Razvan and Bengio, Samy and Bengio, Yoshua},
  booktitle = {Proceedings of the 34th International Conference on Machine Learning},
  series = {Proceedings of Machine Learning Research},
  volume = {70},
  pages = {1019--1028},
  year = {2017},
  url = {https://proceedings.mlr.press/v70/dinh17b.html}
}

@article{elhage2021framework,
  title={A Mathematical Framework for Transformer Circuits},
  author={Elhage, Nelson and Nanda, Neel and Olsson, Catherine and Henighan, Tom and Joseph, Nicholas and Mann, Ben and Askell, Amanda and Bai, Yuntao and Chen, Anna and Conerly, Tom and DasSarma, Nova and Drain, Dawn and Ganguli, Deep and Hatfield-Dodds, Zac and Hernandez, Danny and Jones, Andy and Kernion, Jackson and Lovitt, Liane and Ndousse, Kamal and Amodei, Dario and Brown, Tom and Clark, Jack and Kaplan, Jared and McCandlish, Sam and Olah, Chris},
  journal={Transformer Circuits Thread},
  year={2021},
  url={https://transformer-circuits.pub/2021/framework/index.html}
}

@inproceedings{ericsson2024einspace,
  title={einspace: Searching for Neural Architectures from Fundamental Operations},
  author={Ericsson, Linus and Espinosa, Miguel and Yang, Chenhongyi and Antoniou, Antreas and Storkey, Amos and Cohen, Shay B. and McDonagh, Steven and Crowley, Elliot J.},
  booktitle={Advances in Neural Information Processing Systems},
  volume={37},
  year={2024},
  doi={10.52202/079017-0061}
}

@inproceedings{ferns2004bisimulation,
  title = {Metrics for Finite Markov Decision Processes},
  author = {Ferns, Norm and Panangaden, Prakash and Precup, Doina},
  booktitle = {Proceedings of the 20th Conference on Uncertainty in Artificial Intelligence},
  pages = {162--169},
  year = {2004},
  url = {https://mlanthology.org/uai/2004/ferns2004uai-metrics/}
}

@article{fresco2021longarm,
  title={Long-Arm Functional Individuation of Computation},
  author={Fresco, Nir},
  journal={Synthese},
  volume={199},
  number={5--6},
  pages={13993--14016},
  year={2021},
  doi={10.1007/s11229-021-03407-x}
}

@article{garcia2026nofreeswap,
  title={No Free Swap: Protocol-Dependent Layer Redundancy in Transformers},
  author={Garcia, Gabriel},
  journal={arXiv preprint arXiv:2605.16234},
  year={2026},
  doi={10.48550/arXiv.2605.16234},
  url={https://arxiv.org/abs/2605.16234}
}

@inproceedings{gavranovic2024categorical,
  title={Position: Categorical Deep Learning is an Algebraic Theory of All Architectures},
  author={Gavranovi{\'c}, Bruno and Lessard, Paul and Dudzik, Andrew Joseph and von Glehn, Tamara and Madeira Ara{\'u}jo, Jo{\~a}o Guilherme and Veli{\v c}kovi{\'c}, Petar},
  booktitle={Proceedings of the 41st International Conference on Machine Learning},
  pages={15209--15241},
  year={2024},
  volume={235},
  series={Proceedings of Machine Learning Research},
  publisher={PMLR}
}

@article{geiger2025causalabstraction,
  title = {Causal Abstraction: A Theoretical Foundation for Mechanistic Interpretability},
  author = {Geiger, Atticus and Ibeling, Duligur and Zur, Amir and Chaudhary, Maheep and Chauhan, Sonakshi and Huang, Jing and Arora, Aryaman and Wu, Zhengxuan and Goodman, Noah and Potts, Christopher and Icard, Thomas},
  journal = {Journal of Machine Learning Research},
  volume = {26},
  number = {83},
  pages = {1--64},
  year = {2025},
  url = {https://jmlr.org/beta/papers/v26/23-0058.html}
}

@inproceedings{geva2021ffn,
  title={Transformer Feed-Forward Layers Are Key-Value Memories},
  author={Geva, Mor and Schuster, Roei and Berant, Jonathan and Levy, Omer},
  booktitle={Proceedings of the 2021 Conference on Empirical Methods in Natural Language Processing},
  pages={5484--5495},
  year={2021},
  publisher={Association for Computational Linguistics},
  doi={10.18653/v1/2021.emnlp-main.446},
  url={https://aclanthology.org/2021.emnlp-main.446/}
}

@article{harbecke2019role,
  title={The Role of the Environment in Computational Explanations},
  author={Harbecke, Jens and Shagrir, Oron},
  journal={European Journal for Philosophy of Science},
  volume={9},
  number={3},
  pages={37},
  year={2019},
  doi={10.1007/s13194-019-0263-7}
}

@inproceedings{he2025incontext,
  title = {In-Context Linear Regression Demystified: Training Dynamics and Mechanistic Interpretability of Multi-Head Softmax Attention},
  author = {He, Jianliang and Pan, Xintian and Chen, Siyu and Yang, Zhuoran},
  booktitle = {Proceedings of the 42nd International Conference on Machine Learning},
  series = {Proceedings of Machine Learning Research},
  volume = {267},
  pages = {22686--22742},
  year = {2025},
  url = {https://proceedings.mlr.press/v267/he25q.html}
}

@inproceedings{hewitt2021conditional,
  title={Conditional Probing: Measuring Usable Information Beyond a Baseline},
  author={Hewitt, John and Ethayarajh, Kawin and Liang, Percy and Manning, Christopher D.},
  booktitle={Proceedings of the 2021 Conference on Empirical Methods in Natural Language Processing},
  pages={1626--1639},
  year={2021},
  publisher={Association for Computational Linguistics},
  doi={10.18653/v1/2021.emnlp-main.122},
  url={https://aclanthology.org/2021.emnlp-main.122/}
}

@inproceedings{huang2024incontext,
  title = {In-context Convergence of Transformers},
  author = {Huang, Yu and Cheng, Yuan and Liang, Yingbin},
  booktitle = {Proceedings of the 41st International Conference on Machine Learning},
  series = {Proceedings of Machine Learning Research},
  volume = {235},
  pages = {19660--19722},
  year = {2024},
  url = {https://proceedings.mlr.press/v235/huang24d.html}
}

@inproceedings{jacot2018ntk,
  title = {Neural Tangent Kernel: Convergence and Generalization in Neural Networks},
  author = {Jacot, Arthur and Gabriel, Franck and Hongler, Cl{\'e}ment},
  booktitle = {Advances in Neural Information Processing Systems},
  volume = {31},
  year = {2018},
  url = {https://arxiv.org/abs/1806.07572}
}

@book{kemeny1960finite,
  title = {Finite Markov Chains},
  author = {Kemeny, John G. and Snell, J. Laurie},
  publisher = {D. Van Nostrand Company},
  address = {Princeton, NJ},
  year = {1960}
}

@article{klein2026isomorphism,
  title={Computational Individuation: Isomorphism, Not Indeterminacy},
  author={Klein, Colin},
  journal={Analysis},
  volume={86},
  number={1},
  pages={60--70},
  year={2026},
  doi={10.1093/analys/anaf003}
}

@inproceedings{kornblith2019similarity,
  title={Similarity of Neural Network Representations Revisited},
  author={Kornblith, Simon and Norouzi, Mohammad and Lee, Honglak and Hinton, Geoffrey},
  booktitle={Proceedings of the 36th International Conference on Machine Learning},
  series={Proceedings of Machine Learning Research},
  volume={97},
  pages={3519--3529},
  year={2019},
  publisher={PMLR}
}

@inproceedings{littman2001predictive,
  title = {Predictive Representations of State},
  author = {Littman, Michael L. and Sutton, Richard S. and Singh, Satinder},
  booktitle = {Advances in Neural Information Processing Systems},
  year = {2001},
  url = {https://proceedings.neurips.cc/paper/2001/file/1e4d36177d71bbb3558e43af9577d70e-Paper.pdf}
}

@inproceedings{mainali2026exact,
  title = {Exact Learning Dynamics of In-Context Learning in Linear Transformers and Its Application to Non-Linear Transformers},
  author = {Mainali, Nischal and Teixeira, Lucas},
  booktitle = {Proceedings of the 4th (2025) and 3rd (2024) NeurIPS Workshops on Symmetry and Geometry in Neural Representations},
  series = {Proceedings of Machine Learning Research},
  publisher = {PMLR},
  volume = {282},
  pages = {333--368},
  year = {2026},
  url = {https://proceedings.mlr.press/v282/mainali26a.html}
}

@misc{richmond2024externalism,
  title={Computational Externalism},
  author={Richmond, Andrew},
  year={2024},
  howpublished={PhilSci-Archive preprint},
  url={https://philsci-archive.pitt.edu/id/eprint/24113}
}

@inproceedings{roeder2021linear,
  title={On Linear Identifiability of Learned Representations},
  author={Roeder, Geoffrey and Metz, Luke and Kingma, Durk},
  booktitle={Proceedings of the 38th International Conference on Machine Learning},
  series={Proceedings of Machine Learning Research},
  volume={139},
  pages={9030--9039},
  year={2021},
  publisher={PMLR}
}

@inproceedings{saxe2014exact,
  title = {Exact Solutions to the Nonlinear Dynamics of Learning in Deep Linear Neural Networks},
  author = {Saxe, Andrew M. and McClelland, James L. and Ganguli, Surya},
  booktitle = {International Conference on Learning Representations},
  year = {2014},
  url = {https://arxiv.org/abs/1312.6120}
}

@article{shagrir2001content,
  title={Content, Computation and Externalism},
  author={Shagrir, Oron},
  journal={Mind},
  volume={110},
  number={438},
  pages={369--400},
  year={2001},
  doi={10.1093/mind/110.438.369}
}

@article{shalizi2001computational,
  title = {Computational Mechanics: Pattern and Prediction, Structure and Simplicity},
  author = {Shalizi, Cosma Rohilla and Crutchfield, James P.},
  journal = {Journal of Statistical Physics},
  volume = {104},
  pages = {819--881},
  year = {2001},
  doi = {10.1023/A:1010388907793},
  url = {https://arxiv.org/abs/cond-mat/9907176}
}

@article{sprevak2010computation,
  title={Computation, Individuation, and the Received View on Representation},
  author={Sprevak, Mark},
  journal={Studies in History and Philosophy of Science Part A},
  volume={41},
  number={3},
  pages={260--270},
  year={2010},
  doi={10.1016/j.shpsa.2010.07.008}
}

@inproceedings{sutter2025nonlinear,
  title = {The Non-Linear Representation Dilemma: Is Causal Abstraction Enough for Mechanistic Interpretability?},
  author = {Sutter, Denis and Minder, Julian and Hofmann, Thomas and Pimentel, Tiago},
  booktitle = {Advances in Neural Information Processing Systems},
  year = {2025},
  url = {https://arxiv.org/abs/2507.08802}
}

@article{vasudeva2025attention,
  title = {Implicit Bias and Fast Convergence Rates for Self-Attention},
  author = {Vasudeva, Bhavya and Deora, Puneesh and Thrampoulidis, Christos},
  journal = {Transactions on Machine Learning Research},
  year = {2025},
  url = {https://mlanthology.org/tmlr/2025/vasudeva2025tmlr-implicit/}
}

@inproceedings{vaswani2017attention,
  title={Attention Is All You Need},
  author={Vaswani, Ashish and Shazeer, Noam and Parmar, Niki and Uszkoreit, Jakob and Jones, Llion and Gomez, Aidan N. and Kaiser, Lukasz and Polosukhin, Illia},
  booktitle={Advances in Neural Information Processing Systems},
  volume={30},
  year={2017},
  url={https://proceedings.neurips.cc/paper_files/paper/2017/hash/3f5ee243547dee91fbd053c1c4a845aa-Abstract.html}
}

@inproceedings{wei2016networkmorphism,
  title={Network Morphism},
  author={Wei, Tao and Wang, Changhu and Rui, Yong and Chen, Chang Wen},
  booktitle={Proceedings of the 33rd International Conference on Machine Learning},
  pages={564--572},
  year={2016},
  volume={48},
  series={Proceedings of Machine Learning Research},
  publisher={PMLR}
}

@inproceedings{white2020encodings,
  title={A Study on Encodings for Neural Architecture Search},
  author={White, Colin and Neiswanger, Willie and Nolen, Sam and Savani, Yash},
  booktitle={Advances in Neural Information Processing Systems},
  volume={33},
  year={2020}
}

@inproceedings{xiong2020layernorm,
  title = {On Layer Normalization in the Transformer Architecture},
  author = {Xiong, Ruibin and Yang, Yunchang and He, Di and Zheng, Kai and Zheng, Shuxin and Xing, Chen and Zhang, Huishuai and Lan, Yanyan and Wang, Liwei and Liu, Tieyan},
  booktitle = {Proceedings of the 37th International Conference on Machine Learning},
  series = {Proceedings of Machine Learning Research},
  volume = {119},
  pages = {10524--10533},
  year = {2020},
  publisher = {PMLR},
  url = {https://proceedings.mlr.press/v119/xiong20b.html}
}

@inproceedings{xu2020usable,
  title={A Theory of Usable Information Under Computational Constraints},
  author={Xu, Yilun and Zhao, Shengjia and Song, Jiaming and Stewart, Russell and Ermon, Stefano},
  booktitle={International Conference on Learning Representations},
  year={2020},
  url={https://openreview.net/forum?id=r1eBeyHFDH}
}

@inproceedings{ying2019nasbench,
  title={{NAS}-Bench-101: Towards Reproducible Neural Architecture Search},
  author={Ying, Chris and Klein, Aaron and Christiansen, Eric and Real, Esteban and Murphy, Kevin and Hutter, Frank},
  booktitle={Proceedings of the 36th International Conference on Machine Learning},
  pages={7105--7114},
  year={2019},
  volume={97},
  series={Proceedings of Machine Learning Research},
  publisher={PMLR}
}

\end{document}